\documentclass{article}

\usepackage[utf8]{inputenc}
\usepackage{microtype}
\usepackage[margin=1in]{geometry}
\usepackage{setspace}
\usepackage{ulem}
\usepackage[dvipsnames]{xcolor}
\definecolor{myRed}{RGB}{232, 33, 50}
\definecolor{myOrange}{RGB}{255, 135, 58}
\definecolor{myCyan}{RGB}{72, 159, 181}
\definecolor{myBlue}{RGB}{47, 22, 224}
\definecolor{myBlack}{RGB}{28, 28, 32}
\definecolor{myDarkGrey}{RGB}{82, 80, 93}
\definecolor{myGrey}{RGB}{157, 155, 164}
\definecolor{myLightGrey}{RGB}{190, 188, 200}
\definecolor{myWhite}{RGB}{237, 231, 227}

\usepackage[hyphens]{url}
\usepackage{float, amsmath, amssymb, amsfonts, amsthm, verbatim, tikz-cd, graphicx, multicol, mathtools, braket, stmaryrd}
\usepackage{subcaption}
\usepackage{booktabs}
\usepackage{wrapfig}
\usepackage{quiver}
\usepackage{bm}
\usepackage[above, below]{placeins}  

\usepackage{titletoc}
\usepackage{shorttoc}
\usepackage{xpatch}

\newtoggle{onlyshorttoc}
\makeatletter
\patchcmd{\@startshorttoc}{\bgroup}{\bgroup\toggletrue{onlyshorttoc}}{}{}
\makeatother

\usepackage[ruled, vlined]{algorithm2e}
\usepackage{textcomp}
\usepackage{tabularx}
\usepackage{mathtools, bbm} 
\usepackage{color, soul}
\usepackage{titlesec}
\def\BibTeX{{\rm B\kern-.05em{\sc i\kern-.025em b}\kern-.08em
    T\kern-.1667em\lower.7ex\hbox{E}\kern-.125emX}}
\usepackage[hidelinks]{hyperref}
\hypersetup{
  colorlinks = true,
  urlcolor = myCyan,
  linkcolor = myRed,
  citecolor = myBlue
}
\usepackage{yfonts}

\usepackage[round]{natbib}
\newcommand{\sent}[1]{\text{`}\textit{\color{myDarkGrey}{#1}}\text{'}}
\newcommand{\sentq}[1]{\sent{#1}}
\newcommand{\word}[1]{\textit{\color{myDarkGrey}{#1}}}
\newcommand{\wordq}[1]{\word{\emph{#1}}}
\newcommand{\wbox}[1]{\texttt{#1}}

\newtheorem{definition}{Definition}
\newtheorem*{definition*}{Definition}
\AtBeginDocument{}%
\AtBeginDocument{}%
\AtBeginDocument{}%
\AtBeginDocument{}%

\DeclareCaptionLabelFormat{adja-page}{\hrulefill\\#1 #2 \emph{(previous page)}}

\titleformat*{\section}{\Large\bfseries}
\titleformat*{\subsection}{\large\bfseries}
\titleformat*{\subsubsection}{\bfseries}

\allowdisplaybreaks

\usepackage{multicol}
\usepackage{wrapfig}
\usepackage{booktabs}

\DeclareCaptionLabelFormat{adja-page}{\hrulefill\\#1 #2 \emph{(previous page)}}

\begin{document}

\title{Towards a Comparative Framework for Compositional AI Models.}

\author{
Tiffany Duneau \\\\
Quantinuum, University of Oxford}

\maketitle

\begin{abstract}
The DisCoCirc framework for natural language processing allows the construction of compositional models of text, by combining units for individual words together according to the grammatical structure of the text. 
The compositional nature of a model can give rise to two things: \textit{compositional generalisation} - the ability of a model to generalise outside its training distribution by learning compositional rules underpinning the entire data distribution - and \textit{compositional interpretability} - making sense of how the model works by inspecting its modular components in isolation, as well as the processes through which these components are combined. We present these notions in a framework-agnostic way using the language of category theory, and adapt a series of tests for compositional generalisation to this setting.

Applying this to the DisCoCirc framework, we consider how well a selection of models can learn to compositionally generalise. We compare both quantum circuit based models, as well as classical neural networks, on a dataset derived from one of the bAbI tasks, extended to test a series of aspects of compositionality.
Both architectures score within $5\%$ of one another on the productivity and substitutivity tasks, but differ by at least $10\%$ for the systematicity task, and exhibit different trends on the overgeneralisation tasks. Overall, we find the neural models are more prone to overfitting the \textit{Train} data.
Additionally, we demonstrate how to interpret a compositional model on one of the trained models. By considering how the model components interact with one another, we explain how the model behaves.
\end{abstract}

\startcontents[overview]

\begin{spacing}{1}
\begin{footnotesize}
\printcontents[overview]{l}{1}{%
\subsection*{Overview}%
\setcounter{tocdepth}{2}%
}
\end{footnotesize}
\end{spacing}

\clearpage

\section{Introduction}
\label{chap:intro}
In recent years, the stage for Natural Language Processing (\emph{NLP}) and Artificial Intelligence (\emph{AI}) more broadly has seen a huge boom following the advent of a new paradigm for constructing models. Large Language Models (\emph{LLMs}) such as GPT-4, \citep{openai_gpt-4_2024} and Llama \citep{touvron_llama_2023} have taken the world by storm and enjoy widespread applications as chat-bots, virtual assistants, code writers and more.

The success of these models lies in their remarkable ability to generate human-like text in response to a wide variety of questions and tasks \citep{bubeck_sparks_2023}. By contrast, our capacity to understand why such models give certain answers, and the content of these answers on certain types of question are not always quite as remarkable \citep{arkoudas_gpt-4_2023}.
A major concern when designing models for certain high-stakes tasks such as advising a patient on their diagnosis or treatment options or making financial decisions, is for humans to be able to understand or verify decisions made by the models \citep{rudin_stop_2019, vilone_notions_2021, ferdaus_towards_2024}. If things go wrong it is important to understand where and why so that we can learn from mistakes and improve our models.

Certain tasks can be solved via the application and chaining together of rules that manipulate some simple base elements. This sort of reasoning is known as compositional (more on this in \autoref{chap:compositionality}), and crucially, is a type of reasoning that LLMs (and Transformers more generally) have historically struggled with \citep{mirzadeh_gsm-symbolic_2024, dziri_faith_2023}. 
Meanwhile, there is significant literature that explores such reasoning in the symbolic AI community, with great success. Theorem provers and assistants \citep{barendregt_chapter_2001}, SAT solvers \citep{alouneh_comprehensive_2019}, and knowledge graphs \citep{yahya_natural_2012} are some of the major, successful outputs of this line of research. Their inherent reliance on logical formulae acts as a double-edged sword however, as while ensuring correctness and making such methods easy to interpret, it also makes these techniques hard to apply in realistic situations where the appropriate logical formulae may not be obvious or easy to generate.

The Distributional Compositional (\emph{DisCo-}) frameworks \citep{coecke_mathematical_2010, coecke_mathematics_2020} provide a way to build models whose structure is guided by formal rules (namely grammar) - the compositional part, whilst the atomic meanings these rules deal with and combine are learnable - the distributional part. 
This dependence on structure can also make them effective at solving tasks requiring compositional reasoning, provided the same rules are used to solve the task as to build the model. Moreover, the model's compositional structure can also provide a path towards interpretability \citep{tull_towards_2024}.
While in previous work, the DisCo models have been both theorised \citep{Zeng_2016, coecke2020foundations, laakkonen_quantum_2024} and implemented \citep{lorenz_qnlp_2021, duneau_scalable_2024} as quantum models,
in this work, we additionally explore an alternative `classical' implementation using neural networks, and make some preliminary comparisons to the more standard quantum architecture.

\subsection{Outline}
This work is structured as follows:
We start with an exploration of compositionality in \autoref{chap:compositionality}, through the example of DisCoCirc.
The key notions of \textit{compositional generalisation}, and \textit{compositional interpretability} are introduced, as well as methods for quantifying how compositional a model is in a practical setting. 
Additionally, work towards presenting compositionality as applied to language and machine learning models more generally, within a categorical framework, is provided in \autoref{app:compositionality}.

The rest of the work is concerned with testing and measuring the concepts laid out in \autoref{chap:compositionality}, for the specific case of DisCoCirc models. 
We provide a brief overview of the datasets and training methodologies used in \autoref{part:methodology}, including the specific implementations of the DisCoCirc framework as a quantum circuit model, and as a classical neural network.

\autoref{sec:compgen_babi} details the results obtained when investigating compositional generalisation. We evaluate a series of models, via a study over four aspects of compositionality.
In \autoref{sec:interp-babi}, we select a trained model to be decomposed and interpreted, showing how its compositional structure leads to compositional interpretability. 
Finally \autoref{chap:outro} summarises the results and indicates a list of directions for future work.

\newcommand{\sentences}[1][L]{\Sigma_{#1}}
\newcommand{\sentencescat}[1][L]{\Sigma_\mathcal{#1}}
\newcommand{\grammarsig}[1][G]{\mathcal{#1}}
\newcommand{\syntaxcat}[1][{\grammarsig}]{S_{#1}}
\newcommand{\meaningcat}[1][L]{\mathcal{M}_{#1}}
\newcommand{\validdiags}[1][L]{\mathsf{D}_{\grammarsig}(#1)}

\newcommand{\interpx}[2][L]{\mathfrak{#2}_{#1}}
\newcommand{\interp}[1][L]{\interpx[#1]{I}}
\newcommand{\labels}[1][]{{\nabla_{#1}}}
\newcommand{\corruptedlabels}[0]{\blacktriangledown}
\newcommand{\oracle}[1][]{g_{#1}}

\newcommand{\model}[1][M]{\mathfrak{#1}}
\newcommand{\modelimp}[1][M]{\mathfrak{#1}_\mathsf{imp}}
\newcommand{\modelfunc}[1][M]{\MakeLowercase{#1}}

\section{Compositionality}
\label{chap:compositionality}
Informally, we can consider a process to be compositional if it attends to only two ingredients in order to accomplish its goal: (i) basic atoms, and (ii) a strategy for combining these atoms.
Formalising what compositionality means for a given goal is then intrinsically tied to a specification of what kind of things are atoms, what kind of rules are available when combining them, and what kind of operations are allowed on these inputs as part of the processing.
For the particular process of assigning meaning to sentences of a natural language (say, English), there has been much debate surrounding whether it is compositional; if so, how to formalise such a thing, and even whether the notion of compositionality has any empirical content at all \citep{partee_compositionality_1984, pelletier_principle_1994, szabo_case_2012}.

Meanwhile, for other goals, category theory has proved an ideal tool for formalising and reasoning about such a notion. Indeed, the presentation of a monoidal category echoes the above dichotomy: it is inherently a story about atoms (objects and morphisms), and what it means to put them together. More generally, we can consider categories with extra structure as a way to formalise what the rules in question are. As for the allowed operations: the specification of a structure-preserving functor between categories seems an ideal way to capture the sort of compositional-friendly operations we might want our process to consist of.

Here, we shall approach the problem from a slightly different angle: rather than trying to find a compositional account of meaning in general, we will be interested in learning approximations to this process - natural language processing models - that are compositional in a useful way, for a restricted fragment of language that avoids the pitfalls set out in \citet{partee_compositionality_1984} (for example, idioms, context-dependent meanings, and composite nouns) and may therefore uncontrovertially be taken as compositional.
We will consider compositionality to be a
property of a process, taken \textit{relative to a choice of rules} (the syntax) which may or may not apply to the following:
\begin{itemize}
    \item [(A)] The meaning of a natural language, such as English.
    \item [(B)] The internal structure of a model (\textit{syntactic}).
    \item [(C)] The function a model implements (\textit{semantic}).
\end{itemize}
We distinguish situations (B) and (C), introducing the respective notions of \textit{syntactic} and \textit{semantic} compositionality to describe them.
Ultimately, our aim is to ensure that the syntax relative to which a particular model and natural language task are compositional, are in fact one and the same: this is what makes the compositionality of the model useful.
\autoref{sec:tests_for_comp} lays out five aspects of compositionality that can be used in two ways: first as a way to ensure that the compositional rules are appropriate, and second as a way to test empirically whether a model is semantically compositional.
In \autoref{sec:compgen_babi}, we use these tests to compare a series of models trained on the same task.

\subsection{Definitions by example}
\label{sec:comp-defs-ex}
For steps towards a more formal treatment, see \autoref{app:compositionality}; here we will introduce the relevant concepts via the example of DisCoCirc, as applied to the question answering task explored in \autoref{part:methodology}. Futher examples are also provided in \autoref{app:grammar-examples}.

\subsubsection{Syntax}
Starting at the highest level, our grammar involves two main class of rule. The first concerns the internal parsing of a sentence, decomposing it into a subject, verb, object, and so on, with the appropriate dependencies between each type. For the task we will analyse here, the required rules are quite simple, as there are only two types of sentences to produce.

The second class of rules are concerned with \emph{coreference} - the process of linking different occurrences of a word (or even different words) as references to the same entity. In the present case, this is as simple as linking one occurrence of \word{Alice} to a later occurrence of the same word, however more sophisticated systems might allow \word{she} and \word{Bob's sister} to refer to Alice too.

\begin{figure}[H]
    \centering
    \includegraphics[width=0.82\linewidth]{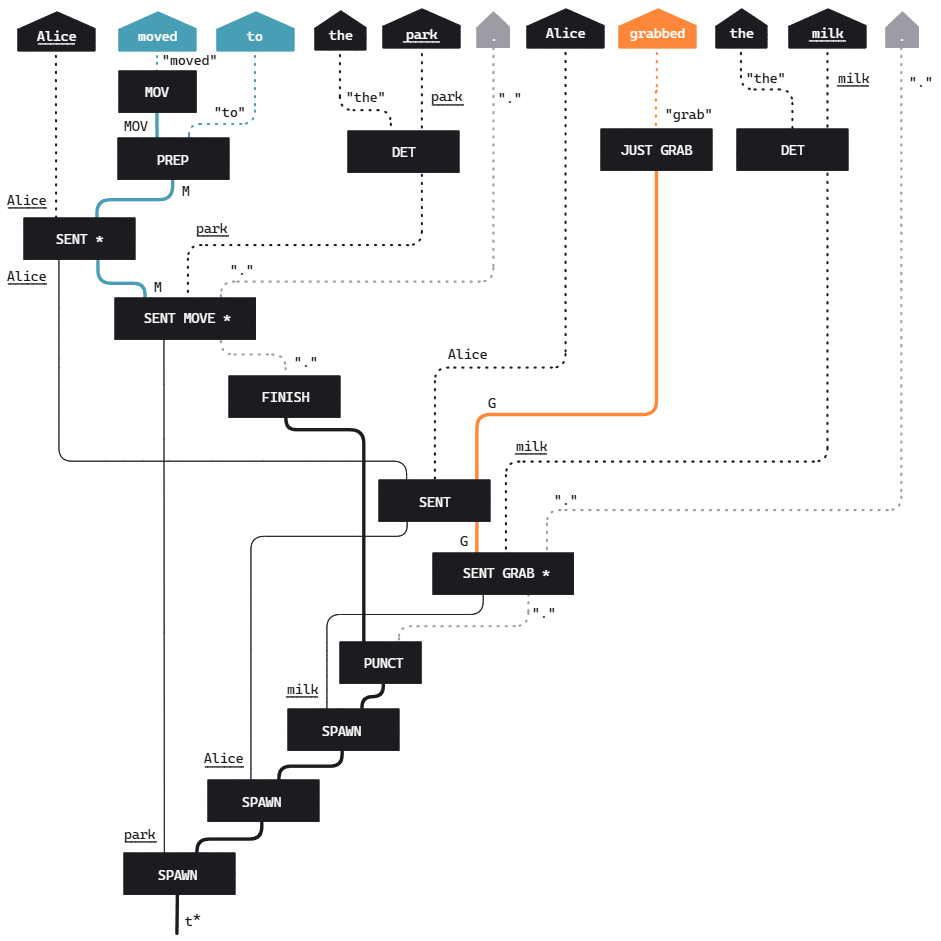}
    \caption{An example syntax derivation for a bAbI~6 story. Each box corresponds to a rule, where the states are base rules that introduce the atoms. The rules are laid out explicitly in \autoref{fig:compositionality/babi6_grammar_rules} of \autoref{app:grammar-examples}. The final type $t_*$ is drawn as a thick black wire. For the nouns, we distinguish regular and underlined wires for co-reference purposes. The dashed wires distinguish types that are within a particular sentence from those that track the entities present in the entire story.}
    \label{fig:compositionality/discocric-parsing-ex}
\end{figure}

We present the rules for bAbI~6 in \autoref{fig:compositionality/babi6_grammar_rules} of \autoref{app:grammar-examples} as generators in a monoidal category, and in an example text derivation in \autoref{fig:compositionality/discocric-parsing-ex}.

\paragraph{}
Though this already hints that sentences are operations applied to some actors or entities, the key insight required for DisCoCirc, is that sentences, or indeed entire stories, act like processes that modify a set of \emph{discourse referents} \citep{karttunen_discourse_1969} that occur in the text. Taking this view allows us to assign some more detailed structure to the wires. Rather than just being labelled as types, we can label these with diagram fragments of some `wire category', that the rule morphisms we defined act on.

\begin{figure}[h]
    \centering
    \includegraphics[width=0.9\linewidth]{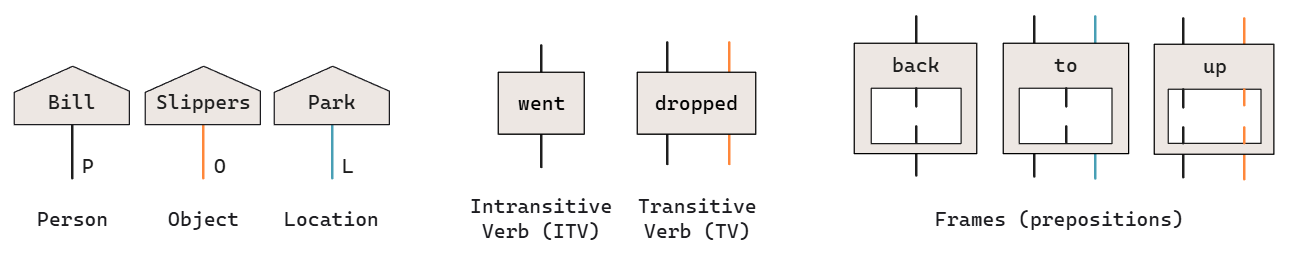}
    \caption{
    Generator shapes for the syntactic `wire' category of DisCoCirc models.
    There are three base objects, $\mathsf{P}$, $\mathsf{O}$ and $\mathsf{L}$, which stand for people, objects and locations respectively, as well as the identity $\mathsf{I}$.
    The full vocabulary list for bAbI task 6 is given by generator shape in \autoref{tab:babi6_vocabulary}. The boxes are drawn with pale insides to distinguish them from the rules.
    }
    \label{fig:discocric-generators}
\end{figure}

The generating morphism types for this `wire category' are given in \autoref{fig:discocric-generators}, diagrams (or morphisms) of the wire category are then any type-checking compositions of the base generators, along the sequential, parallel, and inside directions. Notice that certain boxes are drawn as higher order \emph{frames}, or boxes-with-holes. These represent morphisms that act on other morphisms, where the source types are drawn inside the box with a gap of the intended shape, and the target type is on the outside.
Formally, these morphisms live at a higher level (see \autoref{app:grammar-examples/discocirc}); in the interest of drawing these diagrams more clearly we shall represent this higher level of composition going `inside' the page, as the rules on frames ensure that there is no way to cross from the inside of one hole to another.
Just as the monoidal product requires the existence of all parallel paired morphisms obtained by placing boxes side by side, the `inside' composition requires the existence of morphisms \textit{without} holes that arise once frames have been appropriately filled in.

\begin{figure}[h]
    \centering
    \includegraphics[width = 0.45\linewidth]{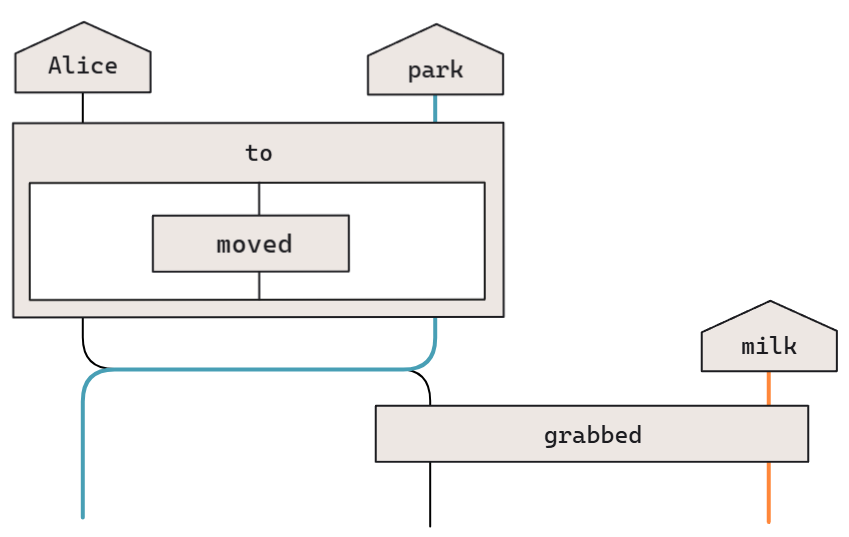}
    \caption{
    Final state obtained in the DisCoCirc wire category when simplifying the parsing diagram in \autoref{fig:compositionality/discocric-parsing-ex}, for the story: \sent{Alice moved to the park. Alice grabbed the milk.}. Notice the the hole in the \textsf{to} frame has been filled in by the box \textsf{moved}.
    }
    \label{fig:compositionality/discocric-parsing-ex-final-diag}
\end{figure}

Having established what the inner structure of the wires looks like, we can now return to the monoidal category that contains our rules. DisCoCirc is a special theory in that, once the wires have been expanded into diagrams, the rule morphisms all become internal wirings, that simply dictate how the base components of the wire category should be plugged together. This is illustrated informally in \autoref{fig:compositionality/app/discocirc_parsing_1} and \autoref{fig:compositionality/app/discocirc_parsing_2} of \autoref{app:grammar-examples}. 
We can hence view the entire derivation as a state which outputs a particular diagram of the wire category. \autoref{fig:compositionality/discocric-parsing-ex-final-diag} displays the resulting state for the example parsing in \autoref{fig:compositionality/discocric-parsing-ex}. Though the final state is one from the wire category alone, the rules are nevertheless important to define which diagrams are stories - not all ways of combining the generators of the wire category will arise from a valid parsing.

\paragraph{}
We are now ready to fix some terminology.
We write $\grammarsig$ to refer to the (monoidal) signature that contains the rules, as well as all the boxes and wires that make up the wire category.
Valid ways of plugging these morphisms together result in diagrams. The category which contains all these diagrams as morphisms, and whose objects are combinations of the base types under a monoidal product, is called the syntax category, $\syntaxcat$. 
For any given story, there is a diagram in $\syntaxcat$ that corresponds to a correct way of parsing it. Typically, we will require that if there are multiple valid derivations, these are formally equivalent (possibly requiring the addition of some extra structure, such as a symmetry). Let $\sentences$ refer to the set of all valid stories in a language $L$, and $\sentencescat$ denote the derived category whose morphisms are indexed by the stories, between dummy source and target types. Let $s_0, s_1 \in \sentences$ be stories, $\sentencescat$ then looks as follows:
$$
\begin{tikzcd}
S_0^s \arrow[r, "s_0"] & S_0^t & S_1^s \arrow[r, "s_1"] & S_1^t & \cdots
\end{tikzcd}
$$
The \emph{syntax}, $e: \sentencescat \rightarrow \syntaxcat$ assigns each story to its parsing. We write $\validdiags$ to denote the diagrams in $\syntaxcat$ that correspond to stories, as selected by $e$. Note that $\syntaxcat$ will also contain some diagrams that correspond to only fragments of stories, like the diagrams for \sent{Alice}, or \sent{to the park}.

\FloatBarrier
\subsubsection{Semantics}
Informally, the semantics of a language is a functor that assigns a meaning to each sentence or story in our language. We call this an \emph{interpretation} and write $\interp: \sentencescat \rightarrow \meaningcat$, for some meaning category $\meaningcat$.

Though this category can in principle be anything we like, a particularly useful class of categories is those that allow us to express computations of some kind. In this way, $\interp$ can map sentences to ways of computing their answers. Considering how an interpretation might factor through the syntax of a language then allows us to capture compositionality: the meaning assigned to a sentence will by definition be a function of the meaning assigned to its individual parts, and the rules that combine these parts together.

\subsubsection{Compositionality}
To make this more concrete, we highlight two kinds of compositionality a model can exhibit. The difference boils down to whether or not we can explicitly break the analysis down via a syntax:

\paragraph{}
\emph{Syntactic compositionality} requires the model to factor through a syntax $e: \sentencescat \rightarrow \syntaxcat$, such that all analysis can be conducted on this decomposition, rather than the model functor directly. 
This means that analysing the decomposition via $\syntaxcat$ \textit{just is} analysing the model: $\model = e;g$, where $g: \syntaxcat \rightarrow \meaningcat$.
Importantly, this implies access to meanings of more than just sentences, since $g$ must also be able to assign values to any diagram of $\syntaxcat$, not just the valid diagrams.

\paragraph{}
\emph{Semantic compositionality}, in contrast, is for situations where we do not have explicit access to a $g$ that factors the model $\model$, or where we wish to analyse a model without explicitly relying on this decomposition, but use merely the fact that it exists. This aims to capture the intuition that while the model behaves in a compositional manner, we might not be able to investigate how it behaves when presented with atoms that do not themselves feature in $\sentences$.
Establishing an equivalence between functions is a hard problem in general; such that finding an appropriate $g$ explicitly may be very difficult. Nevertheless, (as we shall see in the next section) we can find alternative ways to establish that a pair of models are semantically compositional relative to the same structures, without always needing an explicit $g$ to compare to.

\paragraph{}
Pictorially, the distinction lies in the path we use to analyse the model, along the available language data, where the solid arrows commute:
$$
    \begin{tikzcd}
        & \sentencescat \arrow[ld, "{e}"] \arrow[rd, "{\model}"'] \arrow[ld] \arrow[ld, "\text{Syntax}"', dotted, bend right=49] \arrow[rd, "\text{Semantics}", dotted, bend left=49] &             \\
        \syntaxcat \arrow[rr, "{g}"'] &                 & \mathcal{M}
\end{tikzcd}
$$

Weakening to considering only the semantic pathway rather than the decomposition through the syntax is an important distinction that can allow us to analyse models that might not look compositional (such as neural networks), but nevertheless end up implementing a function that exhibits similar properties. In a practical setting, having such compositional properties can be incredibly useful, while finding a syntactically compositional model may be very difficult - semantic compositionality captures this trade-off and characterises the sweet-spot where the model can still benefit from a compositional analysis.

\subsubsection{Useful compositionality}
More explicitly, we will focus here on two properties derived from the above notions of compositionality, that allow us to apply compositionality for practical purposes.

\paragraph{}
A first practical purpose is the context of a natural language processing (NLP) task, $T$. We capture this as some label assignment $\labels[]: \sentencescat[T] \rightarrow \meaningcat[T]$, that sends each sentence to a morphism that outputs the correct label. Further, we assume that $\labels[]$ is \textit{semantically} compositional relative to some syntax $e_{\labels}: \sentencescat[T] \rightarrow \syntaxcat$.

Write $\model|_{\Sigma}$ for the restriction of a model to a certain subset of stories in a language, $\Sigma \subset \sentences$.
$e_{\labels}|_{\Sigma}$ identifies a set of generators of $\syntaxcat$ required to express all the stories of $\Sigma$. We write $\Sigma^*$ to denote the set of stories (still in $\sentences$) that can be obtained using exactly these generators, and call this the \emph{compositional closure} of $\Sigma$ with respect to the syntax $e_{\labels[]}$.

\begin{definition}[Compositional Generalisation, informal]\label{def:compogen_informal}
    A model \emph{compositionally generalises} over $e_{\labels}$ when $\model|_{\Sigma} = \labels[]|_\Sigma$ ensures that $\model|_{\Sigma^*} = \labels|_{\Sigma^*}$, for some choice of $\Sigma \subset \sentences$.
\end{definition}
Provided the initial set $\Sigma$ is suitably chosen so that $\Sigma^* = \sentences$, this captures the requirement that the model $\model$ is \textit{also} semantically compositional with respect to $e_{\labels[]}$.
This presentation lends itself well to practical setups where the model is trained on the set $\Sigma$ such that $\Sigma^* = \sentences$. 
Checking whether the model compositionally generalises establishes whether it has correctly learnt to be semantically compositional.
We explore this task setup in \autoref{sec:compgen_babi}.

\paragraph{}
The second practical purpose concerns interpreting how a model behaves. For this, we introduce a different special interpretation - the \emph{meaning} $\interp^*: \sentences \rightarrow \meaningcat^*$, that interprets a language \textit{for us}. In this case, we assume that this meaning admits a decomposition relative to some syntax $e^*: \sentencescat \rightarrow \syntaxcat$, such that it is syntactically compositional.

\begin{definition}[Compositional Interpretability, informal]\label{def:compinterp_informal}
A model is \emph{compositionally interpretable} when it is syntactically compositional relative to $e^*$.
\end{definition}

In such a case, we can assign meanings to all the components in $\syntaxcat$, and thus all the components the model factors through.
This decomposition can then allow the model's behaviours to be explored at a small scale in order to derive an understanding of the whole. Specifically, we can construct and reason with diagram fragments in $\syntaxcat$ as they mean something to us via $\interp^*$, while also evaluating the model on these fragments and comparing the results in the task meaning category $\meaningcat$. We demonstrate this approach in \autoref{sec:interp-babi}.

\paragraph{}
Both properties are useful individually, however they are even more useful when they coincide: if model exhibits both properties relative to the \textit{same} syntax, such that $e_{\labels[]} = e^*$, the model can generalise beyond the distribution seen in training in a way that is also compatible with its interpretation.

\subsection{Aspects of compositionality}\label{sec:tests_for_comp}
Establishing whether a model is semantically compositional is difficult without providing a syntactically compositional decomposition, however compositional generalisation allows us to use the syntactic compositionality of \textit{one} model to prove that of another. For the sort of tasks where a compositional approach might be prised, we will typically have independent grounds to believe that the task labels $\labels$ are semantically compositional relative to some syntax $e_{\labels}$. A desirable model $\model$ is then one that both agrees with $\labels$ on the training set, and moreover is semantically compositional with respect to the same syntax, as this will guarantee that $\model$ approximates $\labels$.

Following \citet{hupkes_compositionality_2020}, we identify certain properties a syntax may exhibit - aspects of compositionality - which provide a way to select a training set and corresponding compositional closure for measuring compositional generalisation. These tests are called aspects of compositionality, as they are properties we typically require of a syntax in order to consider functions that factor through it compositional. Recall that very few requirements were placed on the kinds of generators allowed to build the syntax, beyond requiring them to be finite. Though this makes the formalism sufficiently expressive to capture any formal grammar, many of these will be degenerate in some way. The properties from which the tests are derived can then be viewed as ways to ensure being compositional relative to that syntax is not vacuous.

In order to define these tests, we will make use of a utility function $\mathsf{acc}_A$ that computes the accuracy of a model over a subset $A$ of $\sentences$ (as judged by $\labels[]$).
This is of concern because we want to build models that are compositional relative to the \textit{same} syntax $e_{\labels[]}$ as $\labels[]$, and that moreover approximate $\labels$.

\paragraph{}
We define the \emph{compositionality factor} of a model, which compares its accuracy on two datasets, $A$ and $B$. From this, we can derive a \emph{compositionality score}, which adjusts the model's accuracy over the dataset $A$ according to how well it generalises to the (generally unseen) dataset $B$. We can view these metrics as measures of compositional generalisation.

\begin{wrapfigure}[15]{r}{0.4\textwidth}
    \centering
    \includegraphics[width=0.85\linewidth]{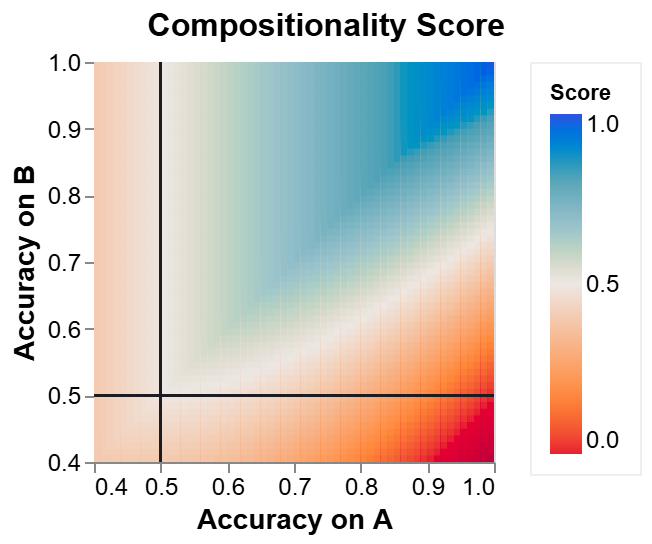}
    \caption{$\textsc{cScore}(A, B)$. Baselines at 0.5 are highlighted with black lines.}
    \label{fig:compositionality/comp-score-fn}
\end{wrapfigure}
We define this for a binary task, with baseline accuracy $0.5$:
\begin{align}
    \label{eq:comp_factor}\tag{\textsc{cf}}
    \textsc{cFact}(A, B) & = \mathsf{max}(0, \mathsf{acc}_\text{A}(\model) - \mathsf{acc}_\text{B}(\model))
    \\
    \label{eq:comp_score}\tag{\textsc{cs}}
    \textsc{cScore}(A, B) & = (1 - 2 \cdot \textsc{cFact}(A, B)) \cdot \mathsf{acc}_\text{A}(\model)
\end{align}

We visualise this function in \autoref{fig:compositionality/comp-score-fn}.
We can also obtain a graded notion of compositionality, $\varepsilon$-compositionality, when a model satisfies:
\begin{equation}
    \label{eq:e_comp}
    \tag{$\varepsilon\textsc{-comp}$}
    \textsc{cFact}(A, B) \leq \varepsilon
\end{equation}

\paragraph{}
In the following sections, we explore various choices that can be made when selecting the datasets $A$ and $B$, in order to target specific aspects of compositionality, formalising the initial adaptation provided in \citet{duneau_scalable_2024}.

\FloatBarrier

\subsubsection{Systematicity}

\begin{figure}[H]
    \centering
    \includegraphics[width=0.6\textwidth]{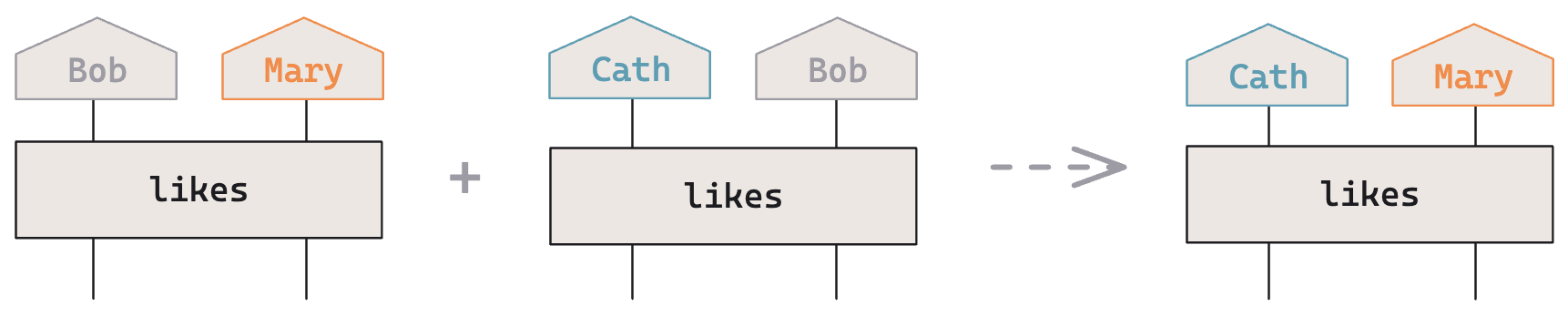}
    \label{fig:compositionality/systematicity}
\end{figure}

This is one of the better known facets of compositionality thanks to \citet{fodor_connectionism_1988} and their arguments for the compositionality of thought via the compositionality of English.
To illustrate with an example, consider the following relatively likely sentence, along with a coarse-grained grammatical type assignment:
\begin{equation}\label{eq:syst_ex_sentence}
    \frac{\word{Mary}}{\mathsf{NP}} \;
    \frac{\word{walked}}{\mathsf{VP}} \;
    \frac{\word{around}}{\mathsf{P}} \;
    \frac{\word{the block}}{\mathsf{NP}} \;
\end{equation}
By replacing each word in \eqref{eq:syst_ex_sentence} with a term of matching grammatical type, we can obtain a new sentence that is much less common:
\begin{equation}\label{eq:syst_ex_sentence2}
    \frac{\word{The Martian sous-chef}}{\mathsf{NP}} \;
    \frac{\word{duelled}}{\mathsf{VP}} \;
    \frac{\word{on}}{\mathsf{P}} \;
    \frac{\word{the telephone box}}{\mathsf{NP}} \;
\end{equation}
Despite being unlikely to have ever encountered \eqref{eq:syst_ex_sentence2} before, we can expect any competent English speaker to nevertheless understand the sentence without issue. The reason being (so the argument goes) the compositionality of English: understanding the structure of the sentence and the individual words is already enough to understand the whole sentence - unlike when we encounter a new word, no new learning needs to occur when we encounter a new sentence composed entirely of old words.

\paragraph{}
A hierarchy of notions of systematicity is introduced in \citet{hadley_systematicity_1994}, depending on how freely terms can be re-combined. In the weak sense, we may only substitute \word{The Martian sous-chef} for \word{Mary} if we have previously encountered sentences in which \word{The Martian sous-chef} occurs as the subject of a verb phrase, whilst in the strong sense it can be substituted anywhere that has the correct grammatical type. To illustrate, we contrast \eqref{eq:syst_ex_sentence2} with \sent{The telephone box duelled on the Martian sous-chef} - while the grammar has been respected, the sentence only makes sense if we personify, and hence change our understanding of, the telephone box.\footnote{See \citet{johnson_systematicity_2004} for a discussion of more problematic versions of this for systematicity in English, and the comments in \autoref{sec:compositionality-caveats}.}

In this work, a syntax is always strongly systematic; however we can consider the strength of a \textit{model's} systematicity.
Applying this distinction to compositional generalisation (reflecting the original usage in \citet{hadley_systematicity_1994}) allows us to characterise how diverse the samples a model is trained on must be in order for it to be able to successfully generalise. The datasets evaluated in this work did not admit such a distinction; we leave further exploration of this concept to future work.

\paragraph{}
We will consider cases where $e_{\labels[]}$ (the syntax relative to which our task is syntactically compositional) is systematic. Take a set of sentences $A \subseteq \sentences$ and their corresponding diagrams $D_A \in \validdiags$. The \emph{systematic closure} $\mathsf{sys}(A)$ of $A$ is the set of diagrams in $\syntaxcat$ that can be (recursively) obtained from those in $D_A$ by swapping sub-diagrams with matching type signatures. Thus:
\begin{definition}[Systematicity]\label{def:systematicity}
    A syntax $e_{\labels[]}$ is \emph{systematic} if there exists a proper subset $A \subset \sentences$ where $A \subset \mathsf{sys}(A)$.
\end{definition}
Notice that this definition is slightly different to the usual presentation, which would require a language to be its own systematic closure: $\mathsf{sys}(\sentences) \subseteq \validdiags$. We avoid this definition as it is true by the definition of a syntax: any type-checking permutation that keeps the outer signature of the diagram fixed will remain a valid sentence. We hence will consider a slightly different property, which requires that the systematic closure be non-trivial.

We can then quantify the systematicity of a \textit{model}, relative to a systematic syntax $e_{\labels[]}$ and a choice of base set $A$ by considering whether the accuracy drops when moving from the base set to its systematic closure (or, as is more typical to machine learning setups, the elements in the systematic closure that do not occur in the base set).
\begin{definition}
    A model $\model$ is \emph{$\varepsilon$-systematic} over $A$ if:
    $\textsc{cFact}(A, \mathsf{sys}(A) \setminus A) \leq \varepsilon$
\end{definition}
Of course, this is easy to satisfy if the model performs badly on the base set. The accuracy of the model on the base set remains an aspect that should be maximised too, which is why in practice we maximise the compositional score $\textsc{cScore}$.

\subsubsection{Productivity}
\begin{figure}[H]
    \centering
    \includegraphics[width=0.7\textwidth]{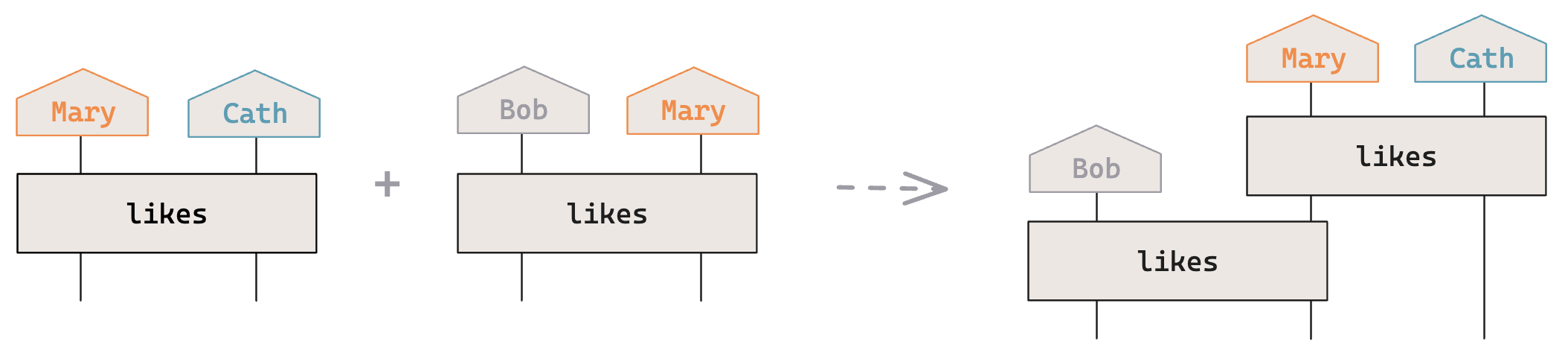}
    \label{fig:compositionality/productivity}
\end{figure}

Productivity of a language is a method through which to create arbitrarily long sentences using only finitely many symbols. It is best viewed as an existence claim about the rules of $L$: namely that there is some combination of rules that can be applied recursively to generate sentences of arbitrary length.
For example, we can arbitrarily augment a simple sentence by adding extra details to the first noun phrase:
\begin{equation}
    \text{\sent{The cat sat on the mat.}}
\end{equation}
\begin{equation}
    \parbox[t]{0.8\linewidth}{\sent{\emph{The} great granddaughter of the Marquis of Lewisham's third niece's old tabby \emph{cat}, who once was the subject of a very long sentence that felt like it would never end (but fortunately will eventually because the both the author and the reader have better things to do), \emph{sat on the mat.}}}
\end{equation}

This leads to a simple definition:
\begin{definition}[Productivity]\label{def:productivity}
    A syntax $e_{\labels[]}$ is \emph{productive} if $|\sentences|$ is not finite.
\end{definition}
This is impractical, however. 
Define the \emph{productive closure} $\mathsf{prod}(A)$ of a certain set $A \subseteq \sentences$ as those sentences that can be obtained by freely applying rules to the atoms found in the sentences of $A$.
This can be obtained by restricting the atoms in the signature $\grammarsig$ that generates $\syntaxcat$ to contain only those occurring in the diagrams $e_{\labels[]}|_A \subset \validdiags$. Call this $\grammarsig'$. The derived syntax category $\syntaxcat[\grammarsig']$ will be a sub-category of $\syntaxcat$. The sentences in $\mathsf{prod}(A)$ are then those whose image under $e_{\labels[]}$ is contained in $\syntaxcat[\grammarsig']$.

Note that this may not be a finite set; thus, we limit the `depth' of the closure by imposing a maximum on the number of rule morphisms that can feature in a derivation - call this $\mathsf{prod}_n(A)$. This in particular distinguishes productivity from systematicity. We can then apply this to a model:
\begin{definition}
    A model $\mathfrak{M}$ is \emph{$\varepsilon$-$n$-productive} over $A$ if:
    $\textsc{cFact}(A, \mathsf{prod}_n(A) \setminus A) \leq \varepsilon$
\end{definition}
In the case of texts, productivity is easy to achieve by concatenating sentences or other texts together, however this can also occur at the sentence level via the construction of complex phrases. In this work, we will focus on concatenating sentences, in order to keep individual sentences gramatically simple.

\subsubsection{Substitutivity}
\begin{figure}[H]
    \centering
    \includegraphics[width=0.7\textwidth]{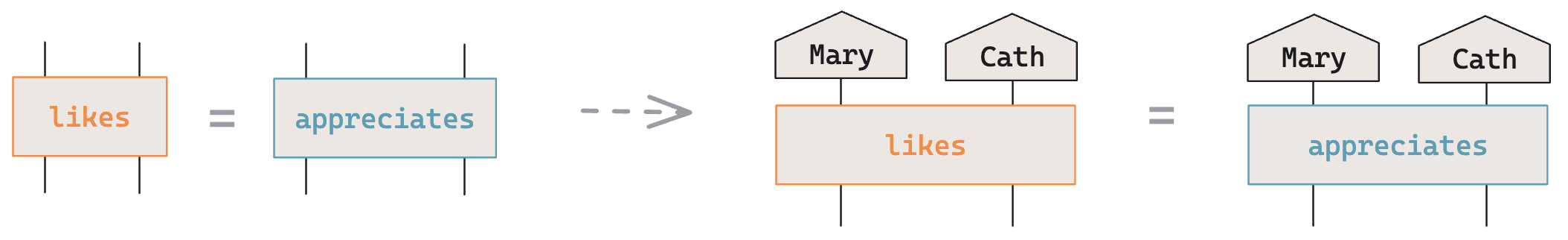}
    \label{fig:compositionality/substitutivity}
\end{figure}

Substitutivity deals with synonyms. The intuition here is that if a model obtains different values for a pair of sentences whose atomic meanings and syntax trees coincide, then the model must be sensitive to something beyond the syntactic structure and atomic meanings, and is therefore not compositional.

The notion seems to presuppose a \textit{syntactically} compositional interpretation in order to determine which atoms are synonyms in the first place, which would make the interpretation substitutive by definition.

In our case, we are interested in \textit{comparing} interpretation functions. We must hence assume the task labels $\labels$ are not merely semantically, but \textit{syntactically} compositional: $\labels = e_{\labels};\oracle$, which we can then use to determine the synonyms. We write $\mathsf{syn}(d)$ for the synonyms of a (possibly composite) diagram fragment $d \in \mathsf{Mor}(\syntaxcat)$:
$$\mathsf{syn}(d) = \{ d' \in \mathsf{Mor}(\syntaxcat)\, |\, \oracle(d) = \oracle(d') \}$$
The \emph{substitutive closure} $\mathsf{sub}(s)$ of a sentence $s \in \sentences$ (and corresponding diagram $D_s$) is the the set of sentences whose diagrams can be obtained by replacing fragments of $D_s$ with a synonym fragment from $\mathsf{syn}(D_s)$. 
In order to account for `nonsense' synonyms (composite terms that aren't ever part of valid sentences but happen to have the same meaning assigned as something else), we further restrict $\mathsf{sub}(s) \subseteq \sentences$.

\begin{definition}[Substitutivity]\label{def:substitutivity}
    A model $\model$ is \emph{substitutive} if for all $s \in \sentences$, 
    $\model(s) = \model(s')$
    where $s' \in \mathsf{sub}(S)$
\end{definition}
For the graded notion, we relax the equivalence a little in order to consider accuracies. Fix a base set $A \subset \sentences$ and define $A' = \{a' \in \mathsf{sub}(a) | a \in A \}$
\begin{definition}
    A model $\mathfrak{M}$ is \emph{$\varepsilon$-substitutive} over $A$ if
    $\textsc{cFact}(A, A' \setminus A) \leq \varepsilon$
\end{definition}

\subsubsection{Overgeneralisation}
\begin{figure}[H]
    \centering
    \includegraphics[width=0.8\textwidth]{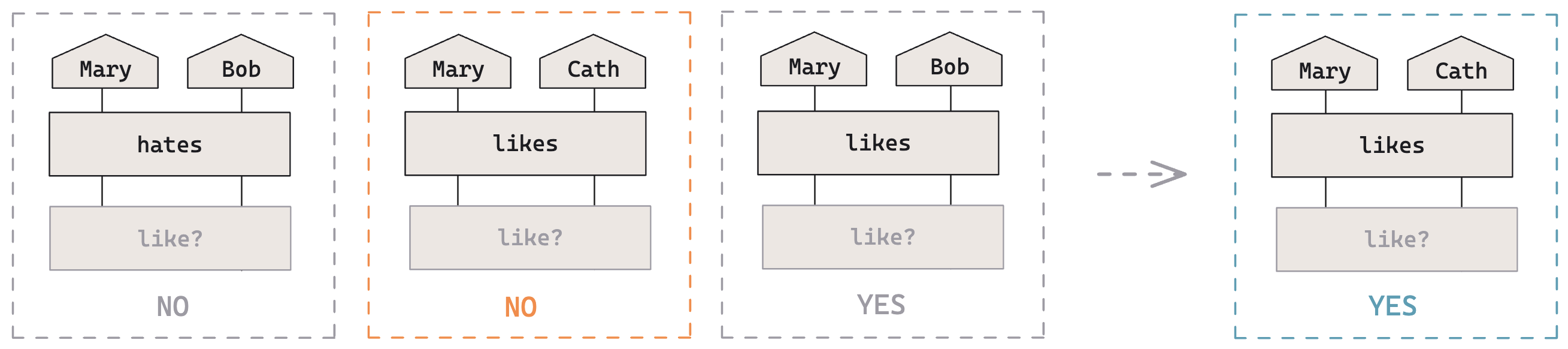}
    \label{fig:compositionality/overgeneralisation}
\end{figure}

This criterion considers a model's ability to memorise exceptions, as opposed to applying a learnt rule to cases where it should not. A compositional model will tend to overgeneralise, whilst a model that is overfitting the data and learning by rote will not. 

Unlike the other tests, overgeneralisation is a property that is visible only when the underlying task is almost, but not \textit{quite} compositional with respect to our chosen syntax $e_{\labels}$. That is, the data involves certain exceptions that do not follow the otherwise compositional pattern. 
Given compositional task labels $\labels$, we can derive corrupted labels $\corruptedlabels$, that give an (incorrect) answer different to $\labels$ on a random choice of inputs $C \subset \sentences$. $\corruptedlabels$ is therefore \textit{not} semantically nor syntactically compositional relative to $e_{\labels}$ (excepting extremely unlikely situations). We restrict training to a subset $A$, where $C \subset A \subset \sentences$.
\begin{definition}[Overgeneralisation]\label{def:overgeneralisation}
    A model $\model$ trained against $\corruptedlabels$ \emph{overgeneralises} if 
    $\mathsf{acc}_C(\model) \geq \mathsf{acc}_{A \setminus C}(\model) $.
\end{definition}
Where $\textsf{acc}$ evaluates the model relative to the true labels $\labels$ rather than the corrupted ones $\corruptedlabels$.
We consider the model to overgeneralise if, relative to $\labels$, it performs just as well on the corrupted instances than on the un-corrupted instances. The graded notion then considers how \textit{much} worse the performance is on the corrupted inputs:
\begin{definition}
    A model $\model$ \emph{$\varepsilon$-overgeneralises} over $C$ if
    $\textsc{cFact}(A \setminus C, C) \leq \varepsilon$
\end{definition}

We can also consider a different notion of overgeneralisation, in which the `exceptions' are introduced in a more structured manner, if we have access to the syntactically compositional decomposition $\labels = e_{\labels};\oracle[]$ of the task labels. 
We can corrupt only the interpretation of selected diagram fragments in $\syntaxcat$ to obtain a corrupted mapping $\oracle[{\corruptedlabels}]$. This will break the functorality of $\oracle[\corruptedlabels]$, and thus $\corruptedlabels = e_{\labels};\oracle[\corruptedlabels]$ will not be syntactically compositional relative to $e_{\labels}$.
Exploring this approach in more detail could provide a route towards handling some non-compositional phenomena while remaining within a mostly compositional framework, and could be particularly promising for the handling of expressions and idioms where particular word sequences (diagram fragments) might be assigned non-compositional meanings when found together but behave compositionally otherwise. While a new atom could always be introduced for such exceptions in order to recover compositionality, this artificially bloats the syntax; doing without could allow for a simpler presentation.
We leave this for future work. 

\subsubsection{Localism}
\begin{figure}[H]
    \centering
    \includegraphics[width=0.7\textwidth]{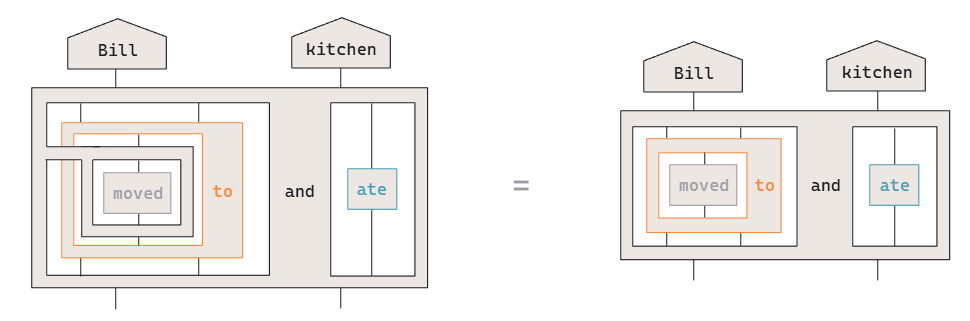}
    \label{fig:compositionality/localism}
\end{figure}

Localism is a property that can be expanded in two ways. The first, much like substitutivity, is tied to having some base interpretation that tells us which terms we might take as equivalent, while the second can be seen as a constraint on the kind of morphisms we allow in $\grammarsig$.
To illustrate the first via an example, a fragment of language often taken as non-local is that of epistemological claims. Consider the following two sentences:
\begin{equation}\label{eq:localism_ex}
    \text{\sent{Mary believes that grass is green.}}
\end{equation}
\begin{equation}\label{eq:localism_ex_2}
    \text{\sent{Mary believes that grass reflects light in the 550 nanometer range.}}
\end{equation}
Although \word{is green} and \word{reflects light in the 550 nanometer range} are synonyms according to their English definitions (in that they identify the same collection of things), it is entirely plausible that Mary doesn't know which wavelengths of light look green. Thus \eqref{eq:localism_ex} might be true while \eqref{eq:localism_ex_2} is false, suggesting that the \textit{whole} sentences should not be synonyms if their meanings are to capture this difference.

However, an interpretation that composes meanings \textit{locally} is not able to distinguish \eqref{eq:localism_ex} and \eqref{eq:localism_ex_2}: both sentences share the `outer' structure \sent{\textsf{N} believes that \textsf{S}}, and the meaning of \textsf{N} (\word{Mary}) and \textsf{S} (\word{grass is green}) are equal.
Put another way, when assigning meaning to a sentence of the form \sent{Mary believes that \textsf{S}}, we need to consider what \textsf{S} means \textit{to Mary} rather than what it means \textit{in English}, which might require some way to `reach in' to the sub-structure of \textsf{S} to make the appropriate adjustments.

\paragraph{}
Formalising this intuition, we consider an interpretation or model to be local if the meaning it assigns to sub-expressions does not depend on the overall expression they are part of. Once again, suppose we have some syntactically compositional labels, such that $e_{\labels}: \sentencescat \rightarrow \syntaxcat$ is our preferred syntax for determining sub-expressions. Localism then becomes a very strong claim:
\begin{definition}\label{def:localism_semantics}
    A model $\model: \sentencescat \rightarrow \meaningcat$ is \emph{local} if it is syntactically compositional relative to $e_{\labels}$.
\end{definition}
Indeed, since localism requires a notion of the interpretation assigned to sub-expressions (or sub-diagrams), $\model$ must be analysed via $\syntaxcat$ for its action on diagram fragments to be well-defined. The functorality of the interpretation then takes care of the rest of the requirements.
Weakening this notion slightly, we may also consider \textit{semantic} locality, in which the model need only be semantically compositional. This notion is more in line with that considered by \citet{hupkes_compositionality_2020}, since it allows for a broader range of model architectures.

\paragraph{}
In this work however, we will focus on functorial models, and so introduce a notion of locality at the syntax level as well. Intuitively, we want to capture the idea that a syntax with a non-local interpretation can be upgraded to a `non-local' syntax with a local interpretation by allowing either the atoms or the rules to be duplicated and indexed to reflect the surrounding grammatical structures.
Thus, we might distinguish \word{light in the 550 nanometer range}, \word{light in the 550 nanometer range}${}_\textit{Mary}$ and \word{green}${}_\textit{Mary}$ if duplicating atoms. Alternatively, duplicating rules might lead to a series of production rules for \sent{\textsf{N} believes that $\mathsf{S}_i$}, where each $\mathsf{S}_i$ has a fixed inner structure.

\begin{definition}[Localism]\label{def:localism_syntax}
    A syntax $e^*: \sentencescat \rightarrow \syntaxcat[\grammarsig^*]$ is \emph{non-local} relative to $e: \sentencescat \rightarrow \syntaxcat$ if we can partition the morphisms and types of $\grammarsig^*$ into equivalence classes, indexed by the items in $\grammarsig$. We write $l: \syntaxcat[\grammarsig^*] \rightarrow \syntaxcat$ for the functor that sends each base morphism in $\grammarsig^*$ to the representative of its equivalence class in $\grammarsig$.
\end{definition}

Suppose we have an initial interpretation $\interp[]$ that is non-local relative to $e$ (and thus cannot factor through $\syntaxcat$). Then we can construct a non-local syntax $e^*$ relative to which $\model$ \textit{is} local:
$$
\begin{tikzcd}
                 & {\sentencescat}\\
    {\syntaxcat} & {\syntaxcat[\grammarsig^*]} & \meaningcat
    \arrow["e"', from=1-2, to=2-1]
    \arrow["{e^*}", from=1-2, to=2-2]
    \arrow["l"', from=2-2, to=2-1]
    \arrow["{\interp[]}", from=1-2, to=2-3]
    \arrow["g"', from=2-2, to=2-3]
\end{tikzcd}
$$
We can also cheat localism somewhat via distributivity in the semantics: rather than duplicating syntax entries, we can add redundancies to the meaning space directly, and allow the outer context (if present) to disambiguate which is required when composing them.

Informally, a graded notion of locality should quantify the redundancy in $\syntaxcat[\grammarsig^*]$. 
We shall be less concerned with this measure in this work, as, unlike the others it is more directly under our control due to its dependence on the syntax. Indeed, the models that we will be investigating are necessarily local, so any amount of non-localism must be added in at the syntax level explicitly. We leave the investigation of non-localism arising in the meaning category to the analysis of final models rather than measuring this specifically.

\subsection{Some caveats and assumptions}
\label{sec:compositionality-caveats}
The task setting at hand - reading comprehension tasks, expressed in English - requires certain assumptions if it is to be solved in a useful way by a compositional model.
To break this down for the general case, we would like both English, and the task of question answering to be compositional relative to some nice rules. In the language used in this paper, this means finding a pair of syntax embeddings $e^*$ and $e_{\labels[]}$ that can witness the compositionality of English meanings ($\interp^*$), and question answering tasks ($\labels[]$), respectively.
This is however, unlikely to be the case.

\paragraph{}
English is, most likely, not quite compositional. There are many well debated phenomena that would suggest otherwise - see e.g. \citet{partee_compositionality_1984} for a summary, and \citet{johnson_systematicity_2004} for a particular focus on systematicity.
For our current purposes, we will assume that we can restrict ourselves to a sub-language of English that \textit{is} compositional in an interesting way, effectively by simplifying and removing all problematic elements.
This does not have to be as damning an assumption as it might initially seem. For one, a lot of English is pretty close to being compositional. Though there are likewise many areas that tend to be less so (poetry, reported speech, ...), things like reading legal documents, logical puzzles, and formal languages likely admit non-vacuous compositional interpretations. Moreover, the benefits of compositional generalisation and interpretability are nothing to frown upon in a setting where training or data collection is expensive - in such cases, perhaps an approximate, but compositional solution would be preferred.

Secondly, we have been intentionally very inclusive about what sorts of rules are permitted as part of a syntax; compositionality is only well defined with respect to a syntax, and as pointed out before we can certainly come up with rules that make being compositional relative to this syntax a vacuous property. When we say that English is not quite compositional, what we actually mean is that the grammar with respect to which English might be made compositional isn't one that satisfies all of our implicit desirability criteria. Though we restrict ourselves to a safe corner for the purpose of this proof-of-concept work, the approach itself has no equivalent constraint.

\paragraph{}
Similarly, question answering in the general case is unlikely to be compositional, for example when asking questions with contextual answers, or indeed when the answers are open-ended, so we will again restrict our attention to a class of compositional tasks.
The tasks we consider are designed to require no background knowledge to solve, so the contextuality issue becomes irrelevant.
For the tasks considered, it does seem to be the case that the only relevant structure that allows us to derive the final answer from the meaning of the context and the meaning of the question, is specifically the co-reference information which tells us which part of the context the question applies to, which is exactly the kind of compositional information available to the model.

\section{Methodology}
\label{part:methodology}

\subsection{Models}
\label{chap:models}
In this work, we study NLP models based on the DisCoCirc framework of \citet{coecke_mathematics_2020}. 
Given a story, we can represent it as a diagram, or \emph{text circuit} by assigning each word to a box of a specific shape, and connecting them together as determined by the grammatical connections between the words as they occur in the story (the rules), as described in \autoref{sec:comp-defs-ex}. 
To evaluate the model, the diagram is interpreted as a computation, whose output we can take to be the model's answer.
Recall that DisCoCirc diagrams contain frames. We will not be concerned with their formal definition here, as we will always expand them out into a series of boxes via the \textsf{Sandwich} expansion described in \citet{laakkonen_quantum_2024}. This expansion is also briefly illustrated in \autoref{fig:frame_expansion} of \autoref{app:grammar-examples}.

For the specific task of Question Answering (\emph{QA}), some extra manipulations are required to derive the answer from the circuit. Given a question, we expand it into a series of \emph{assertions} or statements, each of which correspond to a possible answer to the question. We can then obtain text circuits for these assertions as usual, and compute the overlap between the assertion and story text circuits as a measure of their similarity, in the semantic category of the model. The final answer selected by the model is then the one with the greatest overlap with the story.

\paragraph{}
We consider two classes of model, which differ primarily in the choice of monoidal product selected for the semantic category. Typically, the monoidal product is chosen such that it is not Cartesian - for example the tensor product $\otimes$ over vector spaces. Here, we will also investigate the option of using a Cartesian product $\oplus$, and compare both approaches. We call the first quantum and the second classical after the specific semantic categories we chose to investigate - namely quantum circuits and neural networks.

\subsubsection{Quantum}
\begin{figure}[h]
    \centering
    \begin{subfigure}[b]{0.4\linewidth}
    \centering
    \includegraphics[scale=0.3]{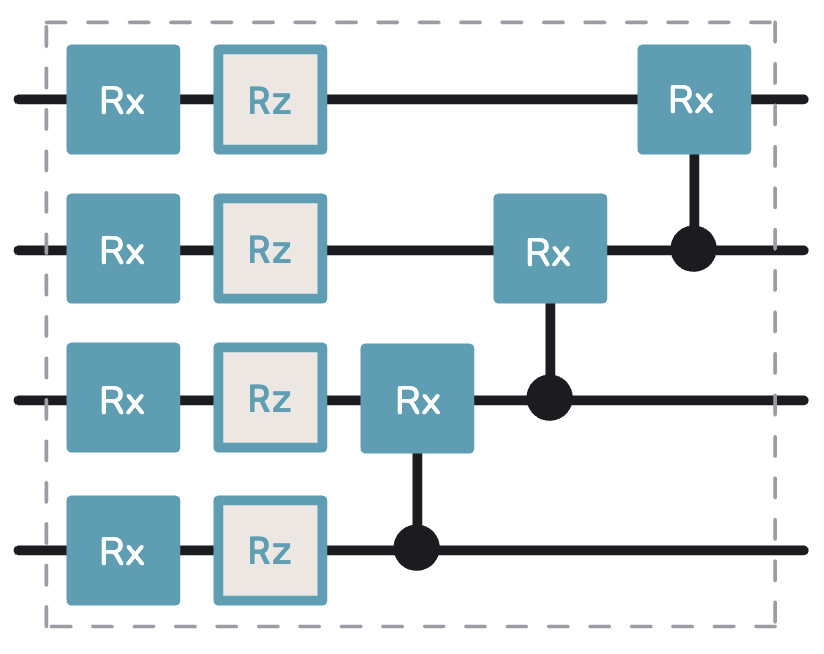}
    \caption{Single layer of the four qubit \textsf{Sim4} Ansatz}
    \label{fig:models/ansatze/sim4}
    \end{subfigure}
    \hspace{2em}
    \begin{subfigure}[b]{0.4\linewidth}
    \centering
    \includegraphics[scale=0.35]{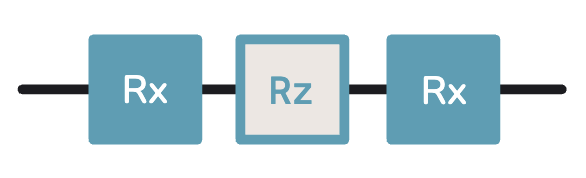}
    \caption{Euler Decomposition}
    \label{fig:models/ansatze/euler}
    \end{subfigure}
    \caption{
    Ansatze used in this work.
    Parametrized gates are highlighted in blue, and the circuit diagrams are read left to right.
    }
\end{figure}
For the non-catesian product, we shall take a \textit{quantum} model with semantic category $CPM(Hilb_2)$, the category of quantum circuits.\footnote{See \citet{coecke_picturing_2017} for details on the CPM construction. In brief, this involves `doubling' diagrams in $Hilb_2$ - a restriction of $FdHilb$, in which all objects decompose as a tensor product of the two-dimensional Hilbert space $\mathcal{H}_2$ - such that they now capture positive operators and density matrices rather than complex vectors and linear maps.} 
Boxes are mapped to a parametrised quantum circuit via an ansatz - in our case \textsf{Sim4} (Ansatz 4 from \citet{Sim_2019}), depicted in \autoref{fig:models/ansatze/sim4}, with three layers, and 1 qubit per noun, following \citet{duneau_scalable_2024}.
Any single-qubit boxes are implemented using the Euler parametrization, which is fully expressive using 3 parameters. We implement all states as unitaries applied to a fixed product state ($\ket{00...0}$).

\subsubsection{Classical}
Our second class of models use the semantic category $\mathcal{NN}$, the category of neural networks, following the idea laid out in \citet{liu_pipeline_2023}. 
We specify the category $\mathcal{NN}$ as the category of arbitrary (differentiable) functions $\mathbb{R}^n \rightarrow \mathbb{R}^m$, with the monoidal product given by the direct sum, $\oplus$. Objects are indexed by the natural numbers $\mathbb{N}$ (including $0$). 
For QA tasks like we investigate here, we require a notion of overlap between texts. In the quantum case, this is implemented via the inner product. We identify the analogue in $\mathcal{NN}$ as
$\mathsf{COMP}_n: \mathbb{R}^n \oplus \mathbb{R}^n \rightarrow \mathbb{R} :: \vec{u} \oplus \vec{v} \Rightarrow \vec{u}^T\vec{v}$

Similarly to the quantum ansatze, we consider certain schemas for the neural network configurations, such that they break down as a series of \emph{linear} functions $\mathbb{R}^n \rightarrow \mathbb{R}^m$ expressed as a matrices $M^{n,m}$, and layers of element-wise non-linear functions $f: \mathbb{R} \rightarrow \mathbb{R}$. A feed-forward neural network (NN) layer is such a matrix followed by a layer of element-wise non-linearities: $(f \oplus \overset{m}{\cdots} \oplus f) \circ M^{n,m}$. We consider the following schema:
\begin{itemize}
    \item $\mathsf{Linear}$: A single NN layer \textit{without} non-linearities.
    \item $\mathsf{Flat}$: A single NN layer.
    \item $\mathsf{Hidden}(d_1, ..., d_n)$: A stack of n + 1 NN layers, with latent dimensions $d_i * noun\_dim$.
\end{itemize}

Note that the expressivity of a neural network strictly increases when moving from $\mathsf{Linear}$ to $\mathsf{Flat}$, and increases (for a fixed hidden dimension) from $\mathsf{Flat}$ to $\mathsf{Hidden}(n)$ to $\mathsf{Hidden}(n, m)$. Since \citet{barron_approximation_1994}, it is known that two hidden layers are sufficient to approximate any function to a desired accuracy. Beyond this threshold increasing the depth can be traded for a (potentially exponential) increase in the width of the hidden dimensions \citep{eldan_power_2016}.
Though less expressive, we consider $\mathsf{Linear}$ models as they provide a close analogue to the quantum scenario, where only the choice of tensor product differs.

The options considered for the classical models are summarised in \autoref{tab:models/neural/functor_choices}. Where feasible, we trained both a \textsf{Linear} and non-linear model for comparison. The remaining hyperparameters were tuned per task and are documented in \autoref{app:training}.

\begin{table}[h]
    \centering
    \begin{tabular}{l r}
        \toprule
        \rule[-0.5em]{0pt}{1.8em}
        \textbf{Hyper parameter} & \textbf{Values Considered}\\
        \midrule
        \rule[-0.5em]{0pt}{1.5em}
         Non-linearity & $\mathsf{ReLu}$, $\mathsf{Mish}$ \\
        \rule[-0.5em]{0pt}{1.5em}
         Configuration & $\mathsf{Linear}$, $\mathsf{Flat}$, $\mathsf{Hidden}(1)$, $\mathsf{Hidden}(2)$, $\mathsf{Hidden}(1,1)$, $\mathsf{Hidden}(2,2)$ \\
        \rule[-0.5em]{0pt}{1.5em}
         Dimension (per noun) & 2, 3, 6, 12, 24, 36, 50\\
        \bottomrule
    \end{tabular}
    \caption{Neural Network functor specifications considered in this work.}
    \label{tab:models/neural/functor_choices}
\end{table}

\subsection{Dataset}
To evaluate our models, we use an extension of an existing synthetic dataset designed to test basic reading comprehension: \textit{bAbI}\citep{bAbI, tamari_dyna-babi_2021}, specfically task 6, which allows for a full study of compositionality. 
The bAbI tasks are intended to cover a range of basic inference and reading comprehension prerequisites. We chose to focus on task 6 in particular, as the format was suited to the DisCoCirc framework, whilst also containing sufficiently interesting structure for testing compositionality. 

\begin{figure}[H]
    \centering
    \includegraphics[width=0.8\textwidth]{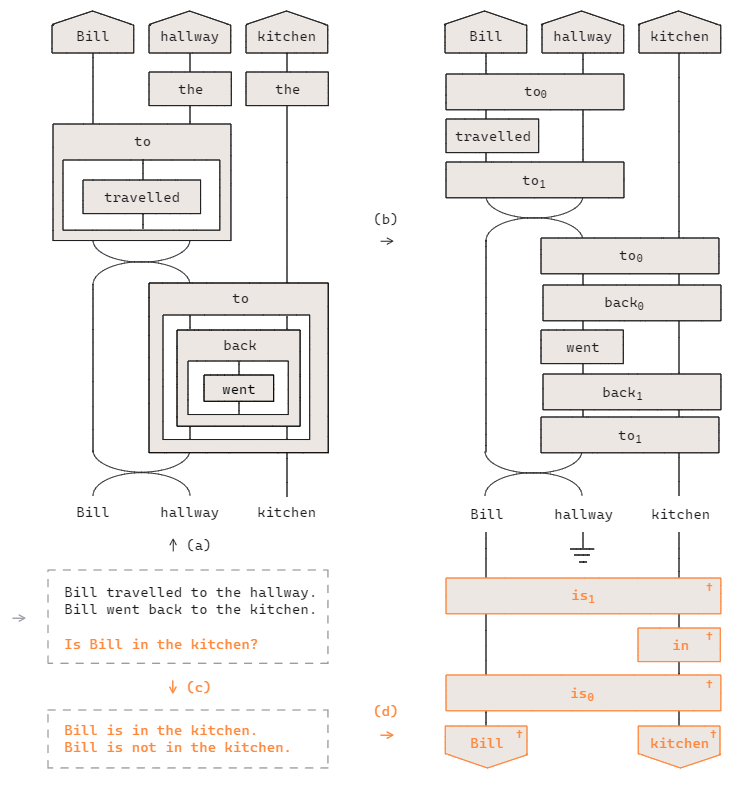}
    \caption{Converting an example data entry into a parametrized quantum circuit. Starting from the text dataset entry, we convert the text into a DisCoCirc diagram (a), then apply semantic rewrites and frame expansions (b). We find the assertions that correspond to the question (c), then parse these as with (a), (b). The resulting diagram is then daggered and connected to the appropriate wires in the text diagram, whilst discarding the other wires (d). We then apply the ansatze to the resulting diagram to obtain a quantum circuit (not shown).}
    \label{fig:training_babi6_parsing_example}
\end{figure}

These stories describe actors moving around a series of rooms. Additionally, the actors may either pick up or discard items. For task 6, the position of the items is irrelevant to the task, as the questions are always of the form \sentq{Is person \textsf{x} in place \textsf{y}?}. An example story is given in \autoref{fig:training_babi6_parsing_example}, and more details are provided in \autoref{app:datasets}.

\subsection{Pipeline}
The first step required is to convert the text-based inputs into DisCoCirc diagrams, or text circuits. In this work the structures were implemented manually based on the outputs of the \texttt{lambeq} parser \citep{kartsaklis2021lambeq, krawchuk_efficient_2025}. These base diagrams must then be converted according to the choices made by the model - this includes any built in semantic rewrites, frame expansions and the choice of ansatze. 
The result is then a collection of parametrised diagrams (quantum circuits in the quantum case, or neural networks in the classical case).
\autoref{fig:training_babi6_parsing_example} provides a step-by-step view of the processing steps applied.

The quantum circuits are converted into tensor networks, which are then evaluated exactly via tensor contraction using \texttt{Tensornetwork}~\citep{roberts2019tensornetwork} on Quantinuum's \textsl{duvel4} server. Neural networks are implemented directly as \texttt{PyTorch}~\citep{Paszke_PyTorch_An_Imperative_2019} models with no inputs.
We use \texttt{PyTorch} to track the parameters and the Adam optimizer~\citep{adam_kingma_2015} to train them. The hyper-parameters are tuned using \texttt{Ax}~\citep{bakshy2018, AxTuningRepo} where applicable.

For each training epoch, the model is evaluated on the entire training dataset in batches, with a parameter update occurring once per batch. We save the parameters obtained at the end of each epoch and record the loss and validation accuracy. Every three epochs, starting from the first, we evaluate and record the accuracy of the model on the training dataset\footnote{We do not always do this every epoch in order to keep model training times down.}. For each training run, we select the model from the epoch with the best score, tie-breaking if necessary with the closest logged training accuracy, then loss if required.

\subsection{Training Setup}
We consider two different training regimes, and conduct preliminary investigations as to their impact on the final model's capacity for compositional generalisation.

\paragraph{Extended validation}
In \citet{duneau_scalable_2024}, it was identified that a single validation set drawn from the same distribution as the \textit{Train} data can be an unreliable estimator for compositional generalisation. In that work, a secondary validation dataset was utilised to tie-break between a collection of candidate models. Here, we consider instead extending the initial validation set.
We construct a validation set that overlaps with both the \textit{Train} and \textit{Test} distributions, targeting the more compositionally difficult examples of the \textit{Train} set, and a tractable subset of the \textit{Test} examples. By evaluating the datapoints on each half of the validation set separately, we can hence obtain an estimate of the model's compositionality score, and use this value for model selection.
This scheme is known as \textit{Valid AB}, other schemes are defined and analysed in \autoref{sec:valid-schemes-results}.

\paragraph{Curriculum learning}
As well as changing the contents of the \textit{Validation} set, we might additionally alter the \textit{Train set}. In this case, the model is trained on progressively more data, where we gradually include more items requiring compositional generalisation.

\paragraph{}
We train a model each for the quantum, \textsf{Linear} and non-linear neural models, using the extended validation regime with the \textit{Valid AB} scheme, for the productivity, systematicity and substitutivity datasets, while for the overgeneralisation task, we consider only a single quantum and neural model architecture.
We conduct a more in-depth analysis of the results on the productivity dataset. For each architecture choice investigated we run 5-fold cross validation to investigate the variability of the results. 
Finally, for the quantum architecture, we conduct an initial exploration of curriculum learning, for each of the productivity, systematicity and substitutivity datasets.
For full details of the setup and hyperparameter choices consult \autoref{app:training}.

\FloatBarrier

\section{Compositional generalisation}
\label{sec:compgen_babi}

\begin{figure}[h]
    \centering
    \includegraphics[width=\textwidth]{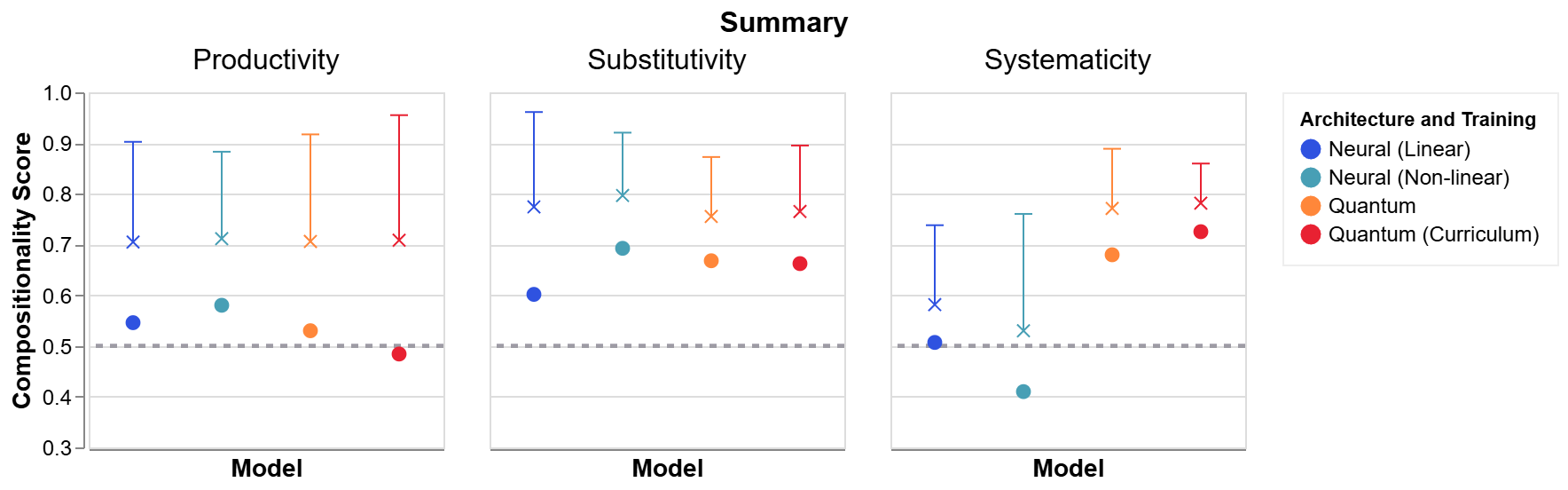}
    \\
    \caption{
        Summary compositionality scores achieved by the best model selected for each task by the quantum and neural DisCoCirc models considered. For each dataset, we plot the model's compositionality score as a point. The \textit{Train} and \textit{Test} accuracies used to compute the productivity score are plotted as a dash and cross respectively, and linked by a line. Recall that the compositionality score is a function of the difference between the training and test accuracies, multiplying the training accuracy.
    }
    \label{fig:results/babi6/summary}
\end{figure}

We test the model's capacity at compositional generalisation for each of the following aspects of compositionality: productivity, systematicity, substitutivity and overgeneralisation.
\autoref{fig:results/babi6/summary} summarises the compositionality scores (as defined in \autoref{eq:comp_score}) achieved by the models for the productivity, substitutivity and systematicity tasks. For more detailed plots split per compositional difficulty of the datapoints, consult \autoref{fig:results/babi/compogen-per-stratum} of \autoref{app:results}.
Recall that localism is not explicitly measured, as the model is local by construction for the fragment of English used in the bAbI 6 dataset.

\subsection{Productivity}
For the productivity task, all of the models scored very similarly on the \textit{Test} set, such that the  relative difference between the training accuracies was the main source of variation in the final productivity score.
We notice that the curriculum training in this case led to the most significant overfitting. In light of the analysis conducted in \autoref{sec:interp-babi}, this is likely due to a bias present in the shorter stories, which the curriculum structure amplified.

\subsubsection{Cross-validation}
\begin{figure}[H]
    \centering
    \textsf{\textbf{Productivity: cross validation}}
    \\
    \begin{subfigure}[b]{0.24\linewidth}
        \centering
        \includegraphics[width=\textwidth]{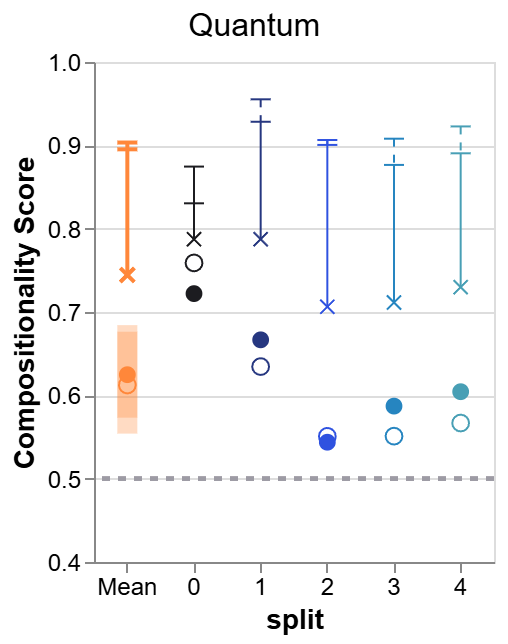}
        \caption{}
    \end{subfigure}
    \hfill
    \begin{subfigure}[b]{0.24\linewidth}
        \centering
        \includegraphics[width=\textwidth]{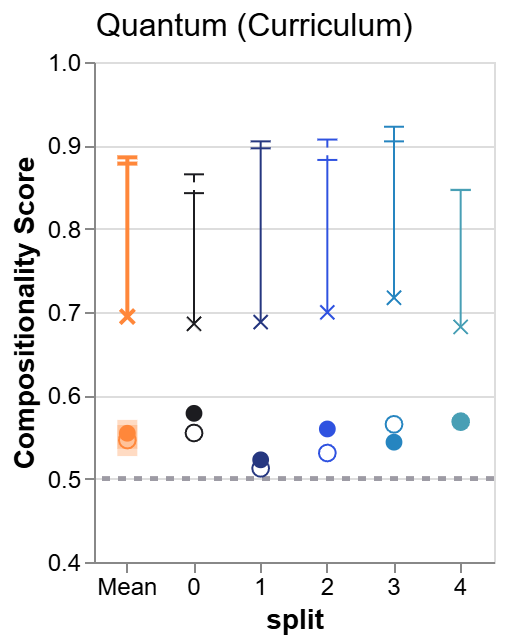}
        \caption{}
    \end{subfigure}
    \hfill
    \begin{subfigure}[b]{0.24\linewidth}
        \centering
        \includegraphics[width=\textwidth]{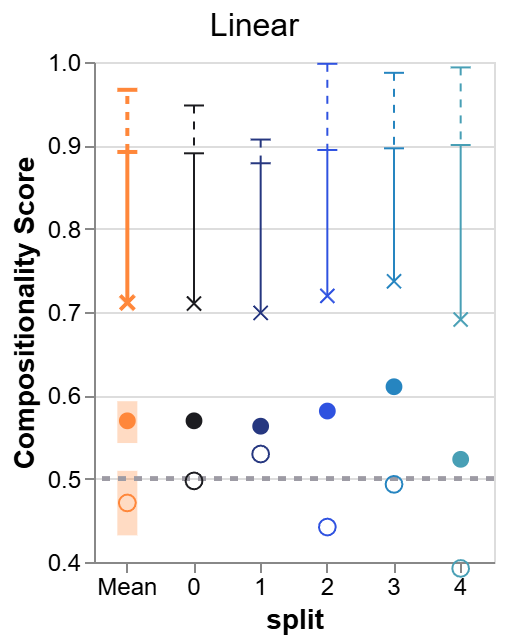}
        \caption{}
    \end{subfigure}
    \hfill
    \begin{subfigure}[b]{0.24\linewidth}
        \centering
        \includegraphics[width=\textwidth]{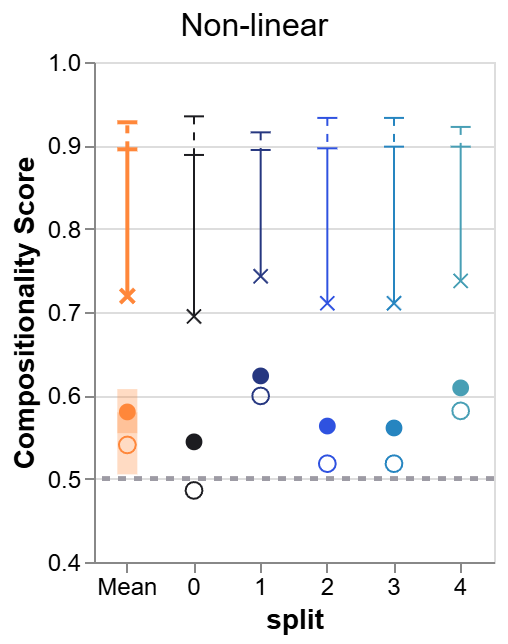}
        \caption{}
    \end{subfigure}
    
    \caption{
    Best selected models according to the \textit{Comp AB} scheme, and the mean performance across the five splits, for the various model architectures considered. (a) and (b) display the quantum architecture (\textsf{Sim4} with 3 layers and 1 qubit per noun), where (b) has been trained via the curriculum method. (c) and (d) display the neural architectures considered: \textsf{Linear} (36 dimensions per noun) and \textsf{Flat} (12 dimensions per noun and \textsf{Mish} activation). 
    The productivity score is plotted as a circle for each split. The mean value is shown in orange, with shaded confidence interval. The \textit{Train} and \textit{Test} accuracies used to compute the productivity score are plotted as a dash and cross respectively, and linked by a line. The dashed line indicates the drop in accuracy between the \textit{Train} set and the \textit{Train'} set. A filled circle indicates the compositionality score relative to the \textit{Train'} distribution, while the open circle plots the score relative to the \textit{Train} accuracy.
    }   
    \label{fig:results/productivity-validation-cv}
\end{figure}

For each of the architectures considered, we also performed 5-fold cross validation to explore how sensitive to the input data each method appeared on the productivity dataset.
\autoref{fig:results/productivity-validation-cv} displays the productivity scores of the best models selected for each fold using the \textit{Valid AB} comp selection method. The results are generally somewhat worse than plotted in \autoref{fig:results/babi6/summary}. This is likely due at least in part to the reduced size of the validation set (20\% of the data available at each stratum rather than the 40\% used when not cross-validating), leading to more a volatile compositionality score estimate. Additionally, for the quantum models, the maximum number of epochs used for each training run was reduced. Given that some of the selected models in the previous section were from epochs in the tail-end of their respective runs, it is possible that this earlier stopping caused some good candidates to be missed.
To quantify how much the models overfit the \textit{Train} data, we also evaluated them on a secondary sub-dataset of the Productivity' dataset, drawn from the same distribution as the \textit{Train} set and known here as \textit{Train'}.

The extended-validation quantum model shows the most variation between splits, which is also somewhat reflective of the available models for each split studied in \autoref{fig:results/productivity_validation-schemes-fit}. Meanwhile, the quantum model trained with curriculum learning is much more uniform, though unfortunately at the cost of selecting models with compositionality scores below $0.6$.
Looking towards the neural models, we see that both perform similarly, though the linear models show more substantial overfitting than the non-linear models: indeed, when considering the compositionality score calculated from the \textit{Train'} distribution, both neural architectures obtain very similar scores.

\subsection{Substitutivity}
The models all again perform similarly, though the \textsf{Linear} model seemed to overfit the training set more, leading to a worse compositionality score. Compared to the productivity task, the models exhibit a lower drop in accuracy from the \textit{Train} to the \textit{Test} set, however we still find that the accuracy seems to decay as the number of synonyms encountered increases.

\begin{figure}[H]
    \centering
    \includegraphics[width=0.9\textwidth]{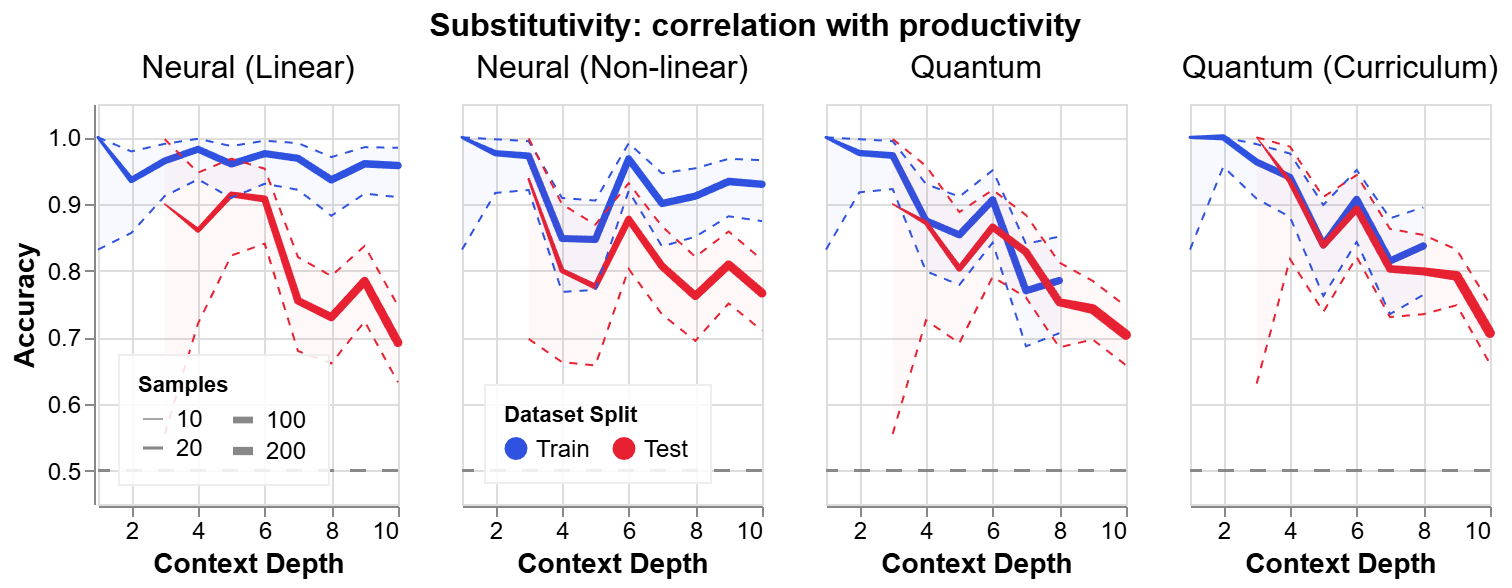}
    \caption{
    Visualising the substitutivity models' accuracy per productivity stratum (context depth).
    The shaded regions represent the 95\% Clopper-Pearson interval for the mean. The thickness of the line reflects the number of datapoints available at each stratum. We display only the scores for the \textit{Train} and \textit{Test} datasets.}   
    \label{fig:results/babi/substitutivity-per-stratum-prod}
\end{figure}

As visualised in \autoref{fig:results/babi/substitutivity-per-stratum-prod}, we see that this may be linked to the models' struggles with productivity, especially for the quantum models. The substitutivity strata are somewhat correlated with the context depth, as shown in \autoref{fig:datasets/babi/sub-balancing} of \autoref{app:datasets}.

\subsection{Systematicity}
This task showed the most variation between the architectures: while the neural models struggled with this task, the quantum model maintained similar scores to the other tasks. The non-linear neural model in particular consistently overfit the training data, with accuracy decaying to random guessing levels over the \textit{Test} set. The quantum models seem to perform best on this task, and indeed, in \autoref{fig:results/babi/systematicity-per-stratum} we see that the accuracy per stratum stays fairly flat over the \textit{Test} set.

\subsection{Overgeneralisation}
\label{sec:results/overgen}

For overgeneralisation, we use a corrupted version of the productivity dataset, including contexts of up to 8 sentences in the training set. The validation set contained only uncorrupted examples, which we sampled in the \textit{Vanilla} fashion\footnote{See \autoref{app:training}, \autoref{fig:training/dataset-split-vanilla} for a definition.} from the entire \textit{Train} distribution.
We additionally evaluated the final models on an unseen, uncorrupted dataset drawn from the same distribution as the training set, described in \autoref{sec:babi-datasets} as \textit{Productivity'} and labelled here as \textit{Test'}.

\begin{figure}[H]
    \centering
    \textbf{\textsf{Overgeneralisation}}
    \\
    \begin{subfigure}[t]{0.24\textwidth}
        \centering
        \includegraphics[width=\textwidth]{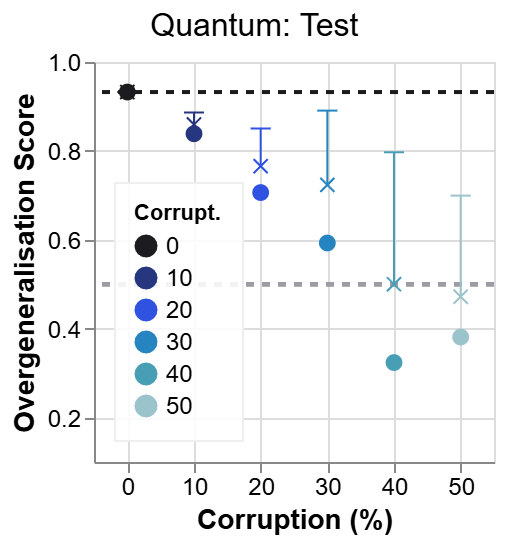}\\
        \caption{}   
        \label{fig:results/overgeneralisation-comp-score-q}
    \end{subfigure}
    \hfill
    \begin{subfigure}[t]{0.24\textwidth}
        \centering
        \includegraphics[width=\textwidth]{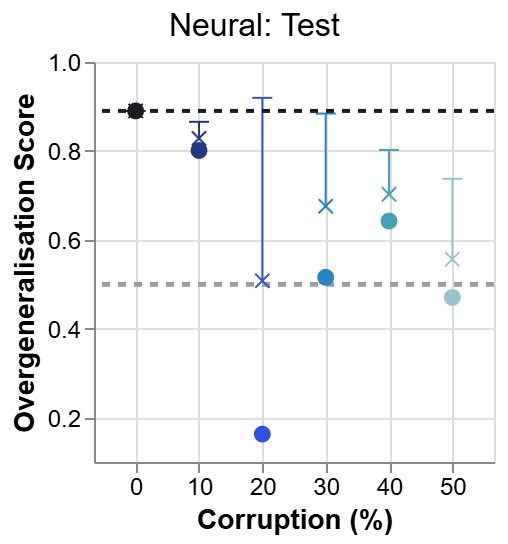}\\
        \caption{}
        \label{fig:results/overgeneralisation-comp-score-n}
    \end{subfigure}
    \hfill
    \begin{subfigure}[t]{0.24\textwidth}
        \centering
        \includegraphics[width=\textwidth]{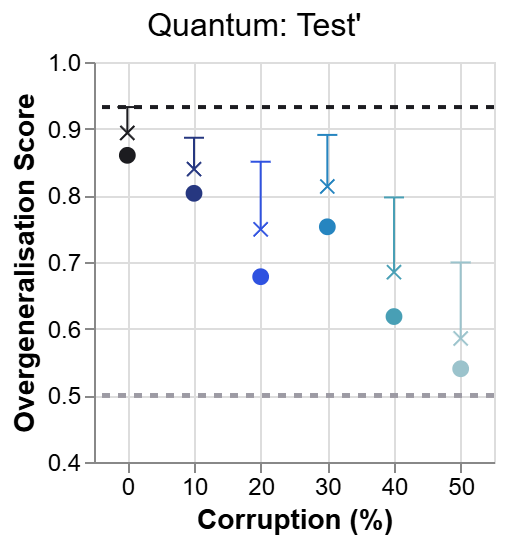}\\
        \caption{}   
        \label{fig:results/overgeneralisation-comp-score-q_unseen}
    \end{subfigure}
    \hfill
    \begin{subfigure}[t]{0.24\textwidth}
        \centering
        \includegraphics[width=\textwidth]{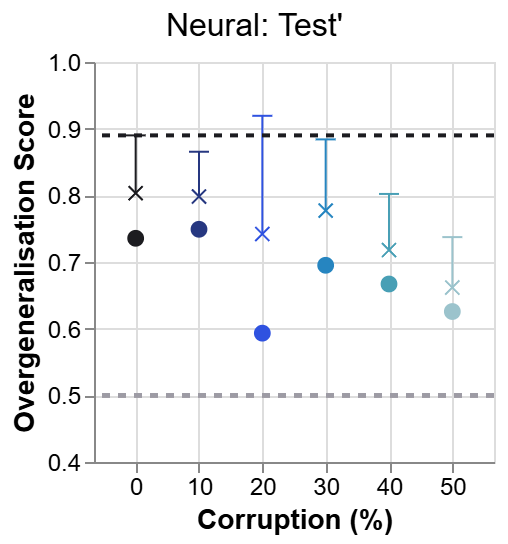}\\
        \caption{}
        \label{fig:results/overgeneralisation-comp-score-n_unseen}
    \end{subfigure}
    \caption{
    Overgeneralisation scores for (a,c) quantum and (b,d) neural models, trained on corrupted datasets. We plot the overgeneralisation score as a filled point; the uncorrupted \textit{Train} accuracy as a dash, and the \textit{Test} or \textit{Test'} accuracy as a cross, linked by a line to the \textit{Train} accuracy.
    The baselines are marked as dashed lines: the \textit{Train} accuracy of the models trained on the uncorrupted dataset (0\%) in black, while the random guessing baseline of 50\% is in grey.
    }
    \label{fig:results/overgeneralisation-comp-score}
\end{figure}

As visualised in \autoref{fig:results/overgeneralisation-comp-score-q}, the quantum model overgeneralises for corruption of up to 30\%, and only starts significantly overfitting from 40\% corruption.
In \autoref{fig:results/overgeneralisation-comp-score-q_unseen}, we see that the drop in accuracy from \textit{Train} to \textit{Test'} is largely proportional to the number of corrupted datapoints, though with a small factor. This confirms that the model is not able to overgeneralise perfectly, and may learn to overfit certain exceptions that then cause mistakes on the \textit{Test'} set.

Meanwhile, the neural model (a \textsf{Flat} model with \textsf{Mish} activation and wire dimension 12) does not seem to exhibit as clear a trend, as per \autoref{fig:results/overgeneralisation-comp-score-n}. In fact, the relatively consistent drop in accuracy from \textit{Train} to the \textit{Test'} dataset distribution in \autoref{fig:results/overgeneralisation-comp-score-n_unseen} suggests that the model is likely overfitting many of the uncorrupted samples too, rather than learning the intended compositional rules. The model at 20\% is particularly guilty of this, but even the reference model trained on an uncorrupted dataset exhibits a similar amount of decay as the model trained on a dataset that had 50\% of the instances corrupted.

\FloatBarrier

\section{Compositional interpretability}
\label{sec:interp-babi}

The built-in compositional structure of DisCoCirc allows us to interpret its behaviour via \textit{compositional interpretability}.
We begin by studying the relationships between the learnt components of the model in isolation. Next, we extend our analysis by examining certain diagram fragments, and compare how they relate to one another in order to find underlying rules the model has learnt to apply. We then use the model's inherent compositional structure to piece together how it performs on the entire dataset.

In previous work, the models studied allowed for direct visualisations on the Bloch sphere, enabling mechanistic interpretability where the behaviour of each box could be studied directly \citep{duneau_scalable_2024}, however this was only feasible due to the small-scale nature of the models. The diagram fragment based techniques used can nonetheless be extended to situations where such a representation is not possible (for example in a model that assigns multiple qubits per noun wire). We demonstrate this approach here, using it to explain how the quantum model trained on the productivity dataset works. We find that the model instead learns to answer a more general question than intended, which accounts for the decay in accuracy as the stories get longer. A detailed analysis and visualisations of the quoted results are provided in \autoref{app:compinterp}.

\paragraph{}
Note that due to the task construction, we are interested in checking whether diagram fragments are equivalent \textit{relative to the question}, as what goes on in the rest of the space is effectively irrelevant. In some cases, we then compute \emph{assertion-relative} overlaps by considering states in the format described in \autoref{fig:interpret/babi/question-relative-diag}, for each of the assertions relevant to answering the question.
An assertion state is appended to the diagram fragments as per the usual rules, with any extra wires discarded. We can then compute the overlaps between the resulting diagram fragment effects.

\subsection{Individual boxes}
When considering the individual boxes, we find that the model has learnt to broadly group the words by synonym equivalence class, though in some cases we find some classes are split as two groups. The object interaction verbs, and the object states themselves are much less clustered, which could be due to their relative unimportance for solving the task. Nevertheless, when factored through the assertion states, we find that the model has still managed to cluster synonyms of \word{grab} together.

\subsection{Diagram fragments}
Next, we compare a series of diagram fragments, and their overlaps relative to the question assertions.
Note that checking these overlaps relative to the question states, is not the same as checking the overlaps between the diagram fragments and the questions, as we would compute when actually solving the task. We consider a representative of each of the movement verb equivalence classes, though not mixed applications. A more in-depth study is provided in \autoref{app:interp/fragments}.

\begin{figure}[h]
    \centering
    {\renewcommand{\thesubfigure}{ID}
    \begin{subfigure}[b]{0.25\linewidth}
        \centering
        \includegraphics[scale=0.45]{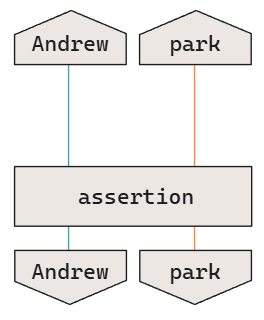}
        \caption{Identity}
        \label{fig:interpret/babi/diag-fragments-(ID)}
    \end{subfigure}}
    \hfill
    {\renewcommand{\thesubfigure}{Ap}
    \begin{subfigure}[b]{0.3\linewidth}
        \centering
        \includegraphics[scale=0.45]{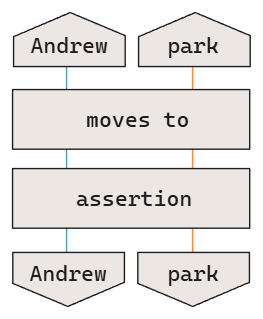}
        \caption{\sent{Andrew moves to the park.}}
        \label{fig:interpret/babi/diag-fragments-(Ap)}
    \end{subfigure}}
    \hfill
    {\renewcommand{\thesubfigure}{Ao}
    \begin{subfigure}[b]{0.3\linewidth}
        \centering
        \includegraphics[scale=0.4]{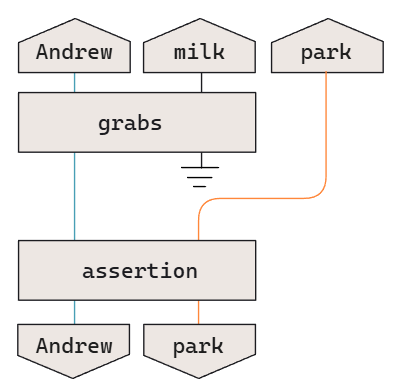}
        \caption{\sent{Andrew grabs the milk.}}
        \label{fig:interpret/babi/diag-fragments-(Ao)}
    \end{subfigure}}
    \caption{
        The assertion-relative fragments evaluated for (ID) the \emph{id}entity operation, (Ap) a single movement (depicted here between \word{\emph{A}ndrew} and \word{the \emph{p}ark}), and (Ao) a single object interaction (between \word{\emph{A}ndrew} and an \textit{\emph{o}bject}).
        The wires corresponding to \word{Andrew} and \word{park} are coloured to help distinguish them.
    }
    \label{fig:interpret/babi/diag-fragments-base}
\end{figure}

\paragraph{}
First, we consider the base cases, where the assertions are appended directly to the initial states. We find that the model has learnt to align the assertion states almost perfectly in the (ID) and (Ap) cases displayed in \autoref{fig:interpret/babi/diag-fragments-base}. For the isolated object interaction (Ao), the model captures the correct relation for the \word{grab}-type verbs, but with a lower margin for the \word{drop}-type verbs.
Though the object embeddings were each distinct, we find that this difference is no longer evident when factoring through the question states, suggesting the model has learnt to consider them equivalent. 

\paragraph{}
Secondly, we compare fragments with at most one verb. We find that although the model appears sensitive to whether a person has changed location, as well as the particular location a person has moved to, the overlap patterns are less ideal when it comes to distinguishing \textit{which} person moved. In particular, changing \textit{who} visited a location will have a much larger impact on the overlap between the story with the negative assertion than the positive one.

While intuitively we might like both overlaps to be close to zero, this does not necessarily mean that the model is \textit{insensitive} to which person visits a location. This is possible since we only care about the \textit{difference} between the assertion overlaps when computing the model's answers, such that a sufficient change in only one of the overlaps may be enough to change the model's final answer, provided the fixed overlap was not too extreme to start with.
Though we did not explicitly check this here, comparing the fragments relative to \textit{different} assertion states would be a way to measure whether the changes identified are in the correct direction. 

\paragraph{}
Thirdly, we consider cases where the target person moves to a different location.
We find that the model does not capture the required axioms very well, suggesting that it struggles to distinguish the order in which a person visits locations, such that the previous locations visited have a non-trivial impact on the final state. This result is somewhat unsurprising given the unitary nature of the model.

Next, we consider cases where another person enters the target location.
We conclude that the model may be quite sensitive to the presence or absence of other people in the target location, despite our intuition that this information is irrelevant to solving the task.

Finally, we consider adding an object interaction. For \word{grab}-type verbs, the model appears to successfully ignore the object interaction, while the \word{drop}-type verbs (especially \word{put down}) have a larger impact. This difference may be due to the sparsity of such instances in the data.

\paragraph{}
\begin{figure}[h]
    \centering
    \includegraphics[width=0.8\linewidth]{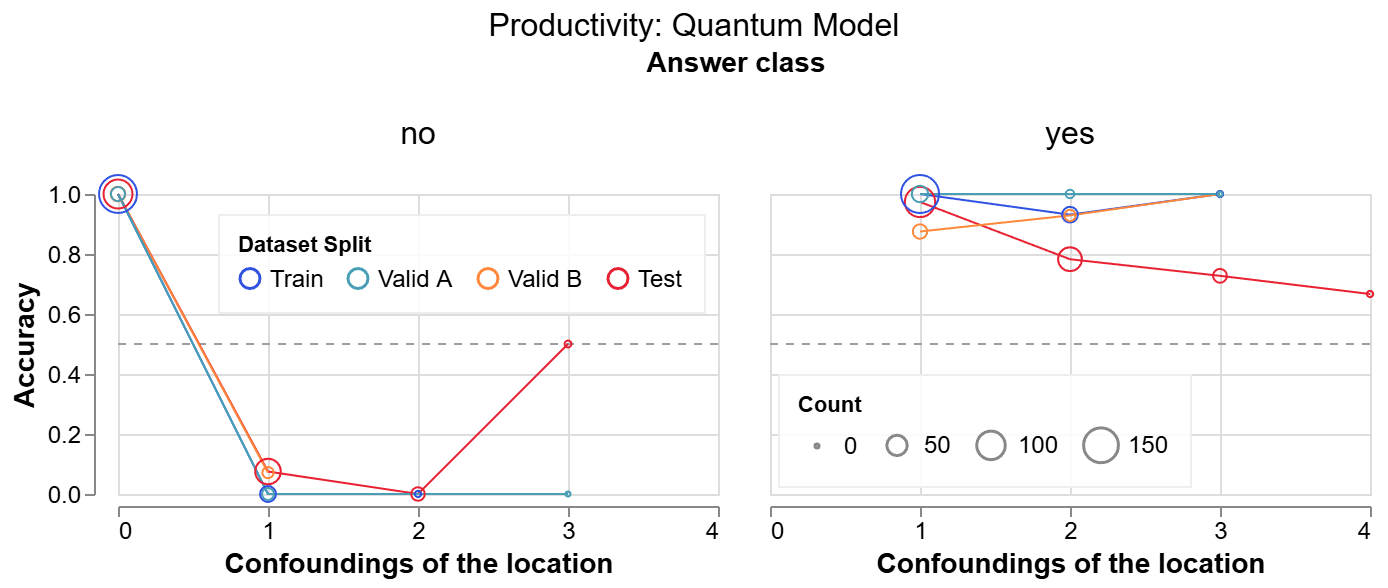}
    \caption{
        Model accuracy per number of sentences in the context that include the target location. A \emph{confounding} sentence is a sentence involving the target location, but a potentially different person. When the expected answer is \word{yes}, the supporting sentence itself is counted, so there are no 0 confoundings. The line thickness represents the number of datapoints present.
    }
    \label{fig:interpret/babi/prod-q-acc-split}
\end{figure}
To summarise these investigations, we find that the model does not distinguish between \textit{when} a person has entered a room very well, nor \textit{who} has entered the room. This suggests that the model learns to answer a question closer to \sentq{Is somebody in the park?} rather than the expected \sentq{Is Andrew in the park?}.
Indeed, we can use this metric to approximately split the model's accuracy, as shown in \autoref{fig:interpret/babi/prod-q-acc-split}. When there is another person present in the target location, the model tends to answer incorrectly, most often by incorrectly answering \word{yes}.
This can then explain why the model struggled with productive generalisation, as the longer contexts tend to have more sentences involving the target person and location.

\FloatBarrier
\section{Discussion}
\label{chap:outro}

In this work we investigate two features generic models can exhibit: \textit{compositional generalisation} - the capacity for a model to learn compositional rules in its training data and thus solve instances drawn from an unseen distribution constructed from the training distribution by applying such rules; and \textit{compositional interpretability} - the capacity for humans to investigate and come to understand how the model works and why it gives a particular answer to a given question.

\paragraph{}
For the specific case of DisCoCirc models, we conduct an initial comparison of various architectures according to their capacity for compositional generalisation over four aspects of compositionality: productivity, systematicity, substitutivity, and overgeneralisation identified in \autoref{sec:tests_for_comp}. 
We implemented both a quantum model, extending the results from previous work \citep{duneau_scalable_2024}, as well as two neural architectures.
Both versions of the model performed similarly well on the productivity and substitutivity tasks, achieving an accuracy of about $0.9 - 0.97$ on the \textit{Train} set, and $0.7 - 0.8$ on the \textit{Test} sets, leading to compositionality scores of about $0.52$ for productivity and $0.65$ for substitutivity. The models differed more significantly in performance over the systematicity task, with the neural architectures reaching a compositionality score of at most $0.5$ while both quantum models were above $0.6$. The two architectures also exhibited different trends on the overgeneralisation task, with the quantum model appearing to overgeneralise for low corruption levels, while the neural model overfit the data more significantly.

Although the overall results did not differ significantly for productivity and substitutivity, we note that the neural architectures tended to require much more significant fine-tuning to obtain these results, while the quantum hyperparameters were more robust across tasks. Quantifying the difference between the quantum and classical architectures in more depth is left for future work.

\paragraph{}
Secondly, we demonstrated how to interpret a model, without relying on understanding or visualising the action of each component box individually. This approach, although allowing for a less deep understanding than that obtained for the \textit{following} models in \citet{duneau_scalable_2024}, is much more scalable to situations where each box applies to more than just two qubits, and indeed could also be adapted to models with a neural architecture.

Though the model we studied did not do a perfect job of capturing the desirable axioms, we attribute this largely to limitations of the dataset and the relatively small scale at which the experiments ran. Despite this, we were nevertheless able to explain how the model performed as it did, and identify the kind of situations that made the model fail. This was possible due to the inherent syntactic compositionality of the DisCoCirc model, which in turn allowed us to leverage compositional interpretability to construct relevant diagram fragments with which to test the model.

\subsection{Future work}
This work acts as a preliminary investigation, paving the way towards further experiments to test the difference between the various implementations of DisCoCirc models, and indeed other classes of model within the framework of compositional generalisation defined in \autoref{chap:compositionality}. We identify two main areas for further experimentation in this direction, exploring the impact of different choices of monoidal product. Finally, we also consider how the tests of compositionality presented here could be further refined.

\paragraph{Architectures with $\otimes$}
Throughout this work, we explored a quantum model as a representative model for taking the monoidal product to be the tensor product $\otimes$. 
There are, however many more options to consider.
Quantum circuits were chosen as a trade-off between expressivity and ease of evaluation (on the assumption that a quantum computer is available at test time), however we heavily restricted the use of ancillas when implementing the models. This in turn restricted the expressibility of the models. A first direction to explore would then be the addition of more ancillas to gates, or perhaps allowing each noun to contain an extra qubit that is ultimately discarded.

Moving further from the quantum circuits studied here, we can also explore classes of models falling within more desirable complexity classes. In the easier-to-evaluate direction, a first choice might be to explore Clifford circuits. While these are easy to simulate, we expect training to be more difficult as optimisation must occur over a discrete training space. 
A more promising approach in this direction is then free-fermion based circuits \citep{surace_fermionic_2022} - a restricted class of quantum circuit that is both easy to simulate and trainable with continuous parameters. These circuits also admit relaxations that increase expressivity at the cost of simulability in a controlled manner, which can allow for a more nuanced trade-off to be reached.

Going in the other direction, we could also try training unconstrained tensor networks directly. While these may be harder to simulate, they are more expressive and would allow models to solve more complex tasks than when restricting to practical\footnote{We take this to mean quantum circuits with limited post-selections and ancilla wires, whose evaluation does not require an exponential number of shots as the size of the circuit increases.} quantum circuits.
Having generalised to generic tensor networks, we can also chose to restrict these in a different way, to retrieve a class that is easy to simulate, for example by considering classical probability distributions. See \citet{glasser_expressive_2019} for an introductory overview.

\paragraph{Architectures with $\oplus$}
In this work we considered two (\textsf{Linear} and non-linear) neural network based architectures, however both were ultimately only considered on a very small scale. One option to explore would then be to scale up the components - exploring deeper and wider networks for each box to take better advantage of the relative efficiency for running the $\oplus$ models. Another line to explore are more sophisticated, or different activation functions. A notable example that was not studied here were the \textsf{Sigmoid}-type activations.

While the quantum hyperparameters were relatively robust across tasks, the neural models were much more sensitive: further exploration could be useful to either find more robust hyperparameters, or determine better heuristics for selecting them.

\paragraph{Further compositionality testing}
In this work, we introduced an initial framework in which to test a series of aspects of compositionality by splitting the data into two parts and comparing the model's accuracy on each one, however the particular choice of the initial set $A$ from which the respective test sets were computed is arbitrary. Future work might then explore whether certain models or architectures can make do with a more minimal base set. For productivity we already have a parametrised notion of the productive closure $\mathsf{prod}_n(A)$, while the notions of \textit{weak} and \textit{quasi} systematicity introduced in \citet{hadley_systematicity_1994} provide a way to grade the `difficulty' of a particular base set relative to its systematic closure.

\FloatBarrier

\subsection*{Acknowledgements}
We thank Benjamin Rodatz, Jonathon Liu, and Razin Shaikh for providing a preliminary version of the neural implementation used in this work; Colin Krawchuk, Sean Tull, Anna Pearson and the anonymous SYCO referees for helpful feedback on drafts and earlier versions of this work, as well as the Oxford team at Quantinuum for supporting this work.

\addcontentsline{toc}{section}{References}
{\small
\bibliographystyle{unsrt}
\bibliography{refs, refsBob, refsThesis}
}

\clearpage
\appendix
\stopcontents[overview]

\startcontents[appendices]
\printcontents[appendices]{l}{1}{\section*{Appendices}\setcounter{tocdepth}{2}}
\clearpage
\section{Defining compositionality}
\label{app:compositionality}
\label{sec:comp-defs}

In this section, we take steps towards formalising the intuitive definitions provided in \autoref{chap:compositionality}.
We want to consider compositionality as a property of a process, which tells us something interesting about how its outputs are related to its inputs. We will primarily be interested in \textit{functors}. In the context of ML and solving a particular task, we will also call them \textit{models}.
\begin{definition}\label{def:function}
     A \emph{functor}, written $f: \mathcal{A} \rightarrow \mathcal{B}$, is a mapping between categories $\mathcal{A}$ and $\mathcal{B}$, that preserves identities and the composition of morphisms. $f$ sends each object in $\mathcal{A}$ to an object in $\mathcal{B}$, and each morphism $a: A \rightarrow A'$ in $A$ to a morphism in $\mathcal{B}$ such that $f(a): f(A) \rightarrow f(A')$.
\end{definition}
A function in the usual sense (as a mapping between sets) can be expressed as a functor between discrete categories, in which the objects are indexed by members of a set, and the only morphisms are identities.
In this work, the domain will usually consist of sentences from a specified language, $L$, which we package as a category where each sentence is an isolated morphism.

\subsection{Atoms and rules (languages)}
Formally, a language $L$ is a set of \emph{sentences}, $\sentences$, each composed of \emph{atoms} $a \in \mathcal{A}$. However $\sentences$ may not contain all possible combinations of the atoms.
One way of specifying a language is by an inclusion rule, for example as a set comprehension: $\{a^nb^n : n \in \mathbb{N}\}$. In this case, $\mathcal{A}$ would be the set $\{a, b\}$. Thanks to  \citet{Chomsky65}, we can also present a language via a \emph{generative grammar}.
Here, our goal is to translate such grammars into a categorical setting, to make formal the intuitive pictures described in \autoref{sec:comp-defs-ex} and used throughout this paper.

Informally, a generative grammar is a set of \emph{production rules} $\mathcal{R}$ that specify how to construct all and only sentences of $\sentences$. Each rule can be understood as a way to replace a non-empty list of symbols with a (possibly empty) list of new symbols. We can write $R: s_0 \cdot ... \cdot s_n \rightarrow t_0 \cdot ... \cdot t_n$ for the rule that allows the string of symbols $s_0 \cdot ... \cdot s_n$ to be replaced with the string $t_1 \cdot ... \cdot t_n$. Valid sentences are taken to be those obtained by iteratively applying rules to a starting symbol $t_*$, until all remaining symbols are atoms $a \in \mathcal{A}$.
More formally:
\begin{definition}\label{def:generative_grammar}
    A \emph{generative grammar} is a tuple $(\mathcal{A}, \mathcal{T}, \mathcal{R}, t_*)$ where:
    \begin{itemize}
        \item $\mathcal{A}$ is a finite collection of atoms.
        \item $\mathcal{T}$ is a finite collection of types.
        \item $\mathcal{R}$ is a finite collection of rules of the form 
            \begin{align*}
                R_{base}   & :b_0 \cdots b_n \rightarrow a\\
                R_{complex}& :s_0 \cdots s_n \rightarrow t_0 \cdots t_m
            \end{align*}
            where $a \in \mathcal{A}$, $b_i, s_i, t_j \in \mathcal{T}$ and $n \geq 1, m \geq 0$.
        \item $t_* \in \mathcal{T}$ is a designated `starting' type.
    \end{itemize}
\end{definition}
Note that there could be multiple rules that apply to a given type $t \in \mathcal{T}$. 
Without loss of generality, we can enforce the $R_{base}$/$R_{complex}$ split by introducing intermediate symbols as necessary to defer the introduction of atoms into a separate rule.
Notice that $\mathcal{A}$, $\mathcal{T}$ and $\mathcal{R}$ are \textit{finite}, while $\sentences$ may be infinite. This is important.

Typically, a generative grammar is paired with a \emph{parser}, which, given a sequence of atoms, tells us how it can be decomposed as a sequence of grammatical rules, and corresponding intermediate type sequences. This sequence of rules is known as a \emph{syntax derivation}. For many practical purposes, this is assumed to be tree-shaped for tractability (as we shall see later, context-free grammars are this shape), however it is most generically a directed graph, or indeed a diagram in a monoidal category.

Derivations whose final type is $t_*$ are called sentences, which we say the parser \emph{recognises}.
Conversely, every sequence of atoms that can be obtained from successive applications of rules to the starting type $t_*$ is a sentence that the grammar \emph{generates}.
We say that a grammar generates (and the associated parser recognises) a language $L$ when the set of sentences it generates (respectively, recognises) is exactly $\sentences$.
There may be multiple grammars that generate the same language.

\subsubsection{Syntax}
We can very naturally translate generative grammars into monoidal signatures, from which we can derive a free monoidal category.

Recall that a monoidal signature is the following data:
\begin{itemize}
    \item A finite set of base objects.
    \item A monoidal product $\otimes$.
    \item A finite set of base morphisms, with source and targets taken as iterated monoidal products of the base types.
\end{itemize}
Which can be extended to a monoidal category by allowing the objects to be defined as the free monoidal compositions of the base objects under $\otimes$, and the morphisms to be the freely (type-checking) compositions of base morphisms, subject to the usual coherence conditions for monoidal categories.
We call each particular composition of morphisms a \emph{diagram}, as it can be drawn as such using the graphical calculus of monoidal categories, as presented in e.g. \citet{SelingerSurvey}.

\begin{definition}\label{def:syntax_cat}
    A generative grammar $(\mathcal{A, T, R}, t_*)$ gives rise to a \emph{grammatical signature} $\grammarsig$:
    \begin{itemize}
        \item Base objects are indexed by the types $t \in \mathcal{T}$ along with a special unit type $\mathsf{I}$. 
        \item The monoidal product is written $\cdot$, representing concatenation.
        \item Base morphisms are indexed by the rules, with type signature given by $R: t_0 \cdots t_n \rightarrow s_0 \cdots s_n$ for each rule $R_{complex} \in \mathcal{R}$. The terminal rules $R_{base}$ are called \emph{states}, and have special signature $R: \mathsf{I} \rightarrow b_0 \cdots b_n$.
        Notice that the type signature for the morphisms is the inverse of the rules. This arises from a difference between parsing and generating - ultimately we want the diagram to represent a parsing, while the grammar is usually used for generation.
    \end{itemize}
\end{definition}
We call the associated monoidal category the \emph{syntax category} $\syntaxcat$ of the grammar.
Notice that not all of the objects of $\syntaxcat$ will represent valid sentences.
An important aspect that is left undetermined here is the kind of structures allowed or imposed on the types and morphisms - various options are explored in \autoref{app:grammar-examples}, and introduced informally as part of the presentation of DisCoCirc in \autoref{sec:comp-defs-ex}. These choices will ultimately determine the complexity of the grammar, as well as its practical use for a particular task.

\paragraph{}
For convenience, define $\sentencescat$ to be the category whose objects are $S_s$ and $S_t$ for each sentence $S \in \sentences$, with morphisms $S: S_s \rightarrow S_t$, for each sentence (and additional identity morphisms). We can then define a syntax functor:
\begin{definition}
    A \emph{syntax} for a language $L$ is a functor $e: \sentencescat \rightarrow \syntaxcat$, assigning a diagram of $\syntaxcat$ to each sentence of $L$, according to its syntax derivation. 
    We write $\validdiags$ to denote the diagrams in $\syntaxcat$ selected by $e$ - these are the diagrams corresponding to sentences of $L$. Equivalently, this is the set of diagrams in $\syntaxcat$ with codomain $t_*$.
\end{definition}

\subsubsection{Semantics}
Languages are also generally equipped with an interpretation, which allows us to assign meaning to the sentences involved.

\begin{definition}\label{def:interpretation}
    An \emph{interpretation} of a language $L$ is a functor $\interp: \sentencescat \rightarrow \meaningcat$, that sends sentences into a \textit{meaning category} $\meaningcat$. We call this the \textit{semantics} of $L$.
\end{definition}
Importantly, we impose as little structure as possible on the interpretation $\interp$ - $\meaningcat$ is not required to have any particular structure. Typically, the models we will encounter implement an interpretation over the language of a given task, where $\meaningcat$ contains the space of possible responses. Though not imposed, $\meaningcat$ will also usually sufficient extra structure to represent computations, such that each sentence is sent to a morphism that \textit{simplifies} to the desired answer, rather than being sent to the answer directly.

There are two interpretations that we take special notice of. 
We write $\labels[]: \sentencescat \rightarrow \meaningcat$ for the interpretation that assigns the correct task labels to each input in $\sentences$.
The second interprets a language for \textit{us} by computing the \textit{meaning} of $L$, written $\interp^*: \sentencescat \rightarrow \meaningcat^*$. Although a language may not have a unique interpretation $\interp^*$, or choice of meaning category $\mathcal{M}^*$, here we will simply fix one such interpretation arbitrarily and ignore the others.
Notice that the target meaning category for $\interp^*$ and $\labels[]$ will typically be different - $\meaningcat^*$ is something that captures human-understandable meanings, while $\meaningcat$ need only contain the possible answers for the task.

\subsection{Compositionality}
We can now formalise our two notions of compositionality.
The naming distinction effectively reflects which `branch' of the language data we use to analyse the model on its way to $\meaningcat$:
$$
    \begin{tikzcd}
        & \sentencescat \arrow[ld, "e"] \arrow[rd, "\interp"'] \arrow[ld] \arrow[ld, "\text{Syntax}"', dotted, bend right=49] \arrow[rd, "\text{Semantics}", dotted, bend left=49] & \\
        \syntaxcat &  & \meaningcat
    \end{tikzcd}
$$

\begin{definition}[Syntactic compositionality]\label{def:syntactic_comp}
    A model $\model: \sentencescat \rightarrow \meaningcat$ is \emph{syntactically compositional} with respect to a syntax $e: \sentencescat \rightarrow \syntaxcat$ when the following diagram commutes, for an explicitly specified functor $g: \syntaxcat \rightarrow \meaningcat$, that preserves the chosen structure of $\syntaxcat$:
    $$
    \begin{tikzcd}
    	{\sentencescat} \\
    	{\syntaxcat} & {\meaningcat}
    	\arrow["e"', from=1-1, to=2-1]
    	\arrow["{\model}", from=1-1, to=2-2]
    	\arrow["g"', from=2-1, to=2-2]
    \end{tikzcd}
    $$
\end{definition}
In other words $\model$ is syntactically compositional if there is some composite functor $e;g$ that $\model$ agrees with for all sentences in $\sentences$, such that analysing $\model$ \textit{just is} analysing $e$ and $g$. This is quite a strong property however. 
Recall that we have not restricted the way that $\model$ is implemented - and in particular it need not be implemented with any reference to $\syntaxcat$ at all. This includes painstaking human annotations, neural networks and lookup tables, as well as implementations that factor through $\syntaxcat$ explicitly. Finding a specific $g$ and proving that $\model = e;g$ may hence be very difficult to do constructively.

\paragraph{}
In some cases, however we may be able to prove a weaker property, where $g$ need not be specified explicitly:
\begin{definition}[Semantic compositionality]\label{def:semantic_comp}
    A model $\model: \sentencescat \rightarrow \meaningcat$ is \emph{semantically compositional} with respect to a syntax $e: \sentencescat \rightarrow \syntaxcat$ when there exists a (possibly not explicitly specified) functor $g: \syntaxcat \rightarrow \meaningcat$ that decomposes $\model$ via the syntax, such that $\model = e;g$.
\end{definition}

This distinction is particularly relevant when we are interested in explaining or interpreting what a functor is doing, when its implementation is obscure (as is the case with neural networks, for example). Rather than treating the function as a black box, syntactic compositionality provides a way of breaking $\model$ into a series of smaller, and hopefully easier to understand, black boxes, whilst semantic compositionality can allow us to reason about the model's behaviour directly, without needing to open the black box at all.

\paragraph{}
We will often also require syntactic compositionality relative to a \textit{specific} $g$, arising as the decomposition of some task labels $\labels = e_{\labels};g$, for example when trying to build a function that approximates labels $\labels$ that is inconvenient to compute via $g$.
It is important that $g$ preserves the chosen structure of $\syntaxcat$ (whether this is merely monoidal, or something more involved such as a rigid or braided category), as this forces it to behave nicely, and prevents the claim from becoming vacuous in some ways. 
In particular this precludes $g$ from being an arbitrary dictionary over sentences as indexed by their grammatical types, since it must also be defined in a consistent way on all intermediate sub-sentences and atoms occurring in $\syntaxcat$.
It could still be vacuous, for example if $g$ sends everything to the same object in $\meaningcat$. 
Nevertheless, we have defined syntactic compositionality relative to a choice of syntax - if the syntax itself has no interesting structure, being compositional relative to that structure is not an interesting property after all.
In \autoref{sec:tests_for_comp} we considered some properties that can help the interpretation and syntax remain interesting, and indeed take these as defining aspects of compositionality, which we further extended into tests that can be used to investigate semantic compositionality.

\paragraph{}
When we discuss whether a language is compositional, we are really discussing whether its meaning $\interp^*$ is compositional with respect to the language's syntax (and analogously for a task with labels $\labels[]$):
\begin{definition}[Compositional language]\label{def:language_comp}
    A language $L$ is called \emph{compositional} when its meaning  $\interp^*$ is semantically compositional relative to $\syntaxcat$.
\end{definition}
We do not distinguish between semantic and syntactic compositionality at the language or task level as we are not concerned with the specific implementations. It is sufficient that there exists a suitable implementation, such that a syntactically compositional model of $\interp^*$ (respectively $\labels[]$) can be found.
On the other hand, we \textit{do} care about the specific implementation when it comes to providing an actual model to solve a given NLP task, and thus must distinguish syntactic from semantic compositionality of models.

\subsection{Useful compositionality}
As we have seen so far, constructing a syntax that makes a model compositional (either syntactically or semantically) doesn't really guarantee that this compositionality is practically helpful, most notably because the syntax that makes the model compositional might not have the `right kind' of rules. In many cases, however, we may already have access to some suitable rules, which we package as syntax-interpretation pairs that are compositional relative to this syntax. Two of these in particular are of interest: the meaning $(e^*: \sentencescat \rightarrow \syntaxcat[\grammarsig^*], \interp^*: \sentencescat \rightarrow \meaningcat^*)$ and the labels $(e_{\labels[]}: \sentencescat \rightarrow \syntaxcat[\grammarsig_\nabla], \labels[]: \sentencescat \rightarrow \meaningcat)$.

We may now reproduce \autoref{def:compogen_informal} and \autoref{def:compinterp_informal} from \autoref{sec:comp-defs-ex}:

\begin{definition}[Compositional interpretabiltiy]\label{def:compinterp}
    A model $\model: \sentencescat \rightarrow \meaningcat$ is \emph{compositionally interpretable} when:
    \begin{itemize}
        \item $\model$ is syntactically compositional relative to $e^*$, and
        \item $\interp^*$ is syntactically compositional relative to $e^*$.
    \end{itemize}
\end{definition}

This is related to the \emph{compositional interpretability} discussed in \citet{tull_towards_2024}: the syntactic compositionality of the model guarantees that $\model$ is a compositional model in the sense required.
We can hence derive an \emph{abstract interpretation} of $g$ (which extends to $\model$ via the syntactic decomposition $\modelfunc = e;g$) from $\interp^*$ via $\syntaxcat[\grammarsig^*]$:

$$
\begin{tikzcd}
	&& {\meaningcat} \\
	{\sentencescat} & {\syntaxcat[\grammarsig^*]} \\
	&& {\meaningcat^*}
	\arrow["{\mathcal{I}^\mathsf{C}}", dashed, from=1-3, to=3-3]
	\arrow[""{name=0, anchor=center, inner sep=0}, "{\model}", curve={height=-12pt}, from=2-1, to=1-3]
	\arrow["{e^*}"{description}, from=2-1, to=2-2]
	\arrow[""{name=1, anchor=center, inner sep=0}, "{\interp^*}"', curve={height=12pt}, from=2-1, to=3-3]
	\arrow["{g}"', from=2-2, to=1-3]
	\arrow["{\mathcal{I}^\mathsf{A}}", from=2-2, to=3-3]
\end{tikzcd}
$$

The concrete interpretation $\mathcal{I}^\mathsf{C}$ is shown with a dashed line to indicate where it approximately fits in (if it exists),\footnote{Technically $\mathcal{I}^\mathsf{C}$ doesn't go from $\meaningcat$ but a modified version where we index the objects according to their type in $\syntaxcat$.} however we will primarily be focusing on the abstract interpretation. 
Notice that \textit{syntactic} compositionality is required in addition to merely semantic compositionality, as the abstract interpretation only allows us to assign meanings to $g$ (and the elements of $\syntaxcat$). 
Without this condition, such an interpretation might be considered a post-hoc explanation of the model, remaining causally disconnected from the actual implementation of $\model$: recall that $\model$ could be anything at all, so long as the answers look right.
Such an explanation intuitively feels lacking - after all, $\model$ could well be implemented as a sentence dictionary, in which case the compositional interpretation has no real bearing on why $\model$ behaves as it does. To avoid delving too deeply into the philosophy of explanatory theories, we shall only attempt to explain syntactically compositional models where $\model = e;g$ is what is actually implemented.

\paragraph{}
That is not to say that semantic compositionality alone is to be discounted - on the contrary, it underpins another significant property.
\begin{definition}[Compositional generalisation]\label{def:compogen}
    A model $\model: \sentencescat \rightarrow \mathcal{M}$ \emph{compositionally generalises} over a syntax $e_{\labels[]}: \sentencescat \rightarrow \syntaxcat[\grammarsig_\nabla]$ and sentences $\Sigma \subset \sentences$ when:
    \begin{itemize}
        \item $\labels[]$ is semantically compositional relative to $e_{\labels[]}$.
        \item If the restrictions $\model|_{\Sigma} = \labels[]|_{\Sigma}$, then  $\model|_{\Sigma^*} = \labels[]|_{\Sigma^*}$ for the compositional closure $\Sigma^*$ of $\Sigma$ with respect to $e_{\labels[]}$.
    \end{itemize}
\end{definition}
To see why this is helpful, we consider the restricted signature $\grammarsig|_{\Sigma}$ (with respect to a syntax $e: \sentencescat \rightarrow \syntaxcat$, which contains items of $\grammarsig$ if and only if they occur in the image of $e|_{\Sigma}$. 
\begin{definition}\label{def:comp-closure}
    The \emph{compositional closure} of a set of sentences $\Sigma \subset \sentences$ with respect to a syntax $e$ is the set of sentences $\Sigma^* \subseteq \sentences$ whose parsing according to $e$ exists in $\syntaxcat[{\grammarsig|_{\Sigma}}]$.
\end{definition}
Thus, for a judicious choice of $\Sigma$, the compositional closure just is the language: $\Sigma^* = \sentences$. Thus if $\model$ compositionally generalises over such a set $\Sigma$, $\model = \labels[]$, and $\model$ is also semantically compositional with respect to $e_{\labels}$.
Notice that the grounding clause establishing agreement over $\Sigma$ is required to ensure that the model is indeed implementing the correct functor - abiding by the same compositional rules as $\labels$ does not by itself enforce that the atoms or rules are assigned the same base values, which is ultimately what determines the values of sentences.

In \autoref{sec:tests_for_comp} of the main text, we present a series of aspects of compositionality and associated tests which can allow us to explore the extent to which a model compositionally generalises. Ultimately, compositional generalisation is the property that we look for in models so that we can remain confident in their correctness on the unseen \textit{Test} distribution.
Semantic compositionality can provide an pathway to proving this beyond an empirical test, provided that $\model$ is semantically compositional relative to the same syntax as $\labels$. In such a case, all that must be established is the grounding clause $\model|_\Sigma = \labels|_\Sigma$ over the training set.

\newpage
\section{Further grammar examples}
\label{app:grammar-examples}
In this section we explore the presentation of a series of grammars as syntax categories, in which extra constraints or structures are placed on the objects or morphisms.

\subsection{From the Chomsky hierarchy}
The Chomsky hierarchy of languages poses constraints on the type of rule that can occur in a generative grammar, which in turn results in constraints on the rule shapes that can occur in a grammatical signature, and thus the shape of the valid derivations $\validdiags$.
The complexity of the language can affect the computational complexity of the resulting diagram when interpreted in a particular semantic category. Selecting an appropriate grammar will hence have impact beyond the parsing step when taking a compositional approach. In this section, we explore the shapes taken by each level of the hierarchy.

\paragraph{Regular}
Regular languages capture the simplest level of the hierarchy, with rules constrained to be either right- or left- regular:
\begin{align*}
    r: s \rightarrow t
    && r_{right}: s \rightarrow at
    && r_{left}: s \rightarrow ta
\end{align*}
Where $a \in \mathcal{A}$ are atoms and $s,t \in \mathcal{T}$ are complex types.
Diagramatically, this results in stair-shaped derivations, going to the right or left as as per the choice of right- or left- regularity.
All morphisms will have exactly one output wire and up to two input wires.

\paragraph{Context-free}
Context-free grammars (CFGs) allow non-terminal symbols to occur on either side of the atoms. Rules may then take a more general form:
\begin{align*}
    R_{base}   :t \rightarrow a &&
    R_{complex}: t \rightarrow t_0 \cdots t_n
\end{align*}
where $t, t_i \in \mathcal{T}$ and $a \in \mathcal{A}$.
This leads to tree-shaped derivations, which can be formalised by the rules of multicategories (to be defined in \autoref{def:multicategory}). These can also be viewed as a restricted monoidal category, where boxes may only have a single wire going out.

\paragraph{Context-sensitive}
Context-sensitive grammars allow rules of the form:
\begin{align*}
    R_{base}   :t \rightarrow a &&
    R_{complex}: s_0 \cdots s_n \cdot t \cdot s'_0 \cdots s'_m \rightarrow s_0 \cdots s_n \cdot t_0 \cdots t_k \cdot  s'_0 \cdots s'_m
\end{align*}
where $t, t_i, s_j, s'_k, \in \mathcal{T}$ and $a \in \mathcal{A}$.
As boxes, these rules result in monoidal diagrams where all the base morphisms break down as a single context-free style box, surrounded by identities. We can view this as an initial example of where the rules are partially expanded out into containing just structural morphisms - the resulting diagrams can hence be simplified into diagrams of a multicategory once constructed.

\paragraph{Recursively Enumerable (RE)}
This captures the most general kind of rule presented in \autoref{def:generative_grammar}, and replicated below, resulting in a generic monoidal category without effects (that is, all boxes must have at least one output type).
\begin{align*}
    R_{base}   :b_0 \cdots b_n \rightarrow a &&
    R_{complex}:s_0 \cdots s_n \rightarrow t_0 \cdots t_m
\end{align*}
where $a \in \mathcal{A}$ and $b_i, s_j, t_k \in \mathcal{T}$.

\subsection{The DisCo- frameworks}
A key aspect of DisCo- family of grammars is that they introduce sub-structures to the types found in the generative grammar.
In the categorical picture, this means endowing the objects of the grammatical signature $\grammarsig$ with further structure - possibly as a further signature $\grammarsig[H]$.
This can then allow for a simpler depiction of the morphisms in $\grammarsig$.

For the application of computing compositional functions of a particular text, this can be particularly useful, by providing a further way to break down or simplify the compositional computation. In the case of DisCoCirc and DisCoCat the simplification is so extreme, that the rule computations reduce to structural morphisms only: the rules hence guide how the meanings of the atoms are to be combined, but they do not introduce any further computations on the atoms they have access to. 

\subsubsection{DisCoCat}
\begin{figure}[h]
    \centering
    \includegraphics[width=0.4\linewidth]{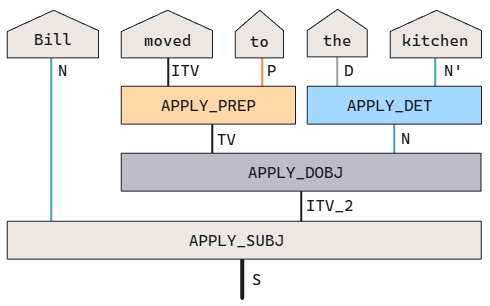}
    \hspace{3em}
    \includegraphics[width=0.4\linewidth]{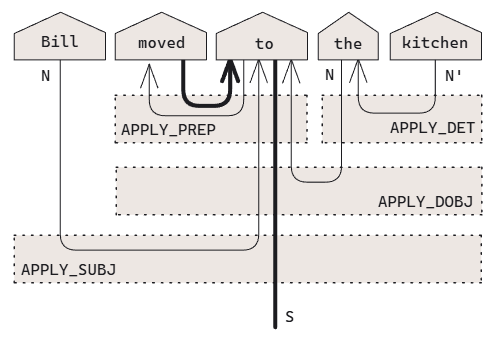}
    \caption{Generic CFG parse, and a finer-grained DisCoCat parse using a pregroup grammar.}
    \label{fig:discocat-parsing-ex}
\end{figure}
For DisCoCat \citep{coecke_mathematical_2010}, we begin with a context-free grammar $\grammarsig$, and endow the types involved with extra structure defined by a new signature $\mathcal{H}$.
In this case, $\mathcal{H}$ consists of base symbols $\{\mathsf{N}, \mathsf{N'}, \mathsf{S}\}$ which form a monoid under concatenation with operation $\otimes$ and unit $\mathsf{I}$. These are chosen to match the identity and monoidal product of $\grammarsig$.
Additionally, we assume iterated left- and right- adjoints for each base type, which we write ${}^-(\;\; )$ and $(\;\; )^-$ respectively. These come with special structural morphisms witnessing the adjoint pairings, which we call cups (and caps):
\begin{align*}
    \eta_{\mathsf{N}_l}: \mathsf{N} \otimes {}^-\mathsf{N} \rightarrow \mathsf{I}
    && \eta_{\mathsf{N}_r}: \mathsf{N}^- \otimes \mathsf{N} \rightarrow \mathsf{I}
\end{align*}
This data defines a pregroup grammar \citep{lambek_pregroups_2006}, which is context-free \citep{pentus_lambek_1993}.
As illustrated in \autoref{fig:discocat-parsing-ex}, each type in the original CFG signature $\grammarsig$ is assigned a (possibly composite) term generated by $\mathcal{H}$. The morphisms of $\grammarsig$ become composites of structural morphisms derived from $\mathcal{H}$.

\subsubsection{DisCoCirc}
\label{app:grammar-examples/discocirc}
\newcommand{\inside}[0]{\vartriangleright}
\newcommand{\insideref}[1][]{\underset{#1}{\vartriangleright}}

An informal presentation is given in the main text. Here we shall be a little more delicate about what we mean when we put (higher-order) diagrams into the wires. 
The first notion we will specify is that of a \emph{frame}. To do so, we shall pass via multicategories (see also Chapter 2 of \citet{leinster_higher_2003}). 
We will write lists of objects as $\underline{A}_i = A_0 A_1 \cdots A_n$, with the (optional) index $i$ that is being ranged over indicated next to the line. The empty list is written $[]$.

\begin{definition}\label{def:multicategory}
A \emph{multicategory} is a generalisation of a category, in which the morphisms may take as input a (possibly empty) list of objects $\underline{A}$. Identities $id_{A}: A \rightarrow A$ are defined on the singleton lists.
Given 
\begin{align*}
    f: \underline{A} \rightarrow X 
    && g: \underline{B}X\underline{C} \rightarrow Y 
    && h: \underline{D}X\underline{E}Y\underline{F} \rightarrow Z
\end{align*}
Sequential composition is written $\inside$ and pronounced `inside', and operates on lists of morphisms $\underline{f}$. Identities act as expected: $\underline{id_{A}} \inside f = f \inside id_X$.
When composing only one non-identity morphism, we can omit writing the identities, and instead specify the target of composition underneath if ambiguous: $\insideref[X]$. Thus:
\begin{align*}
    f \inside g: \underline{B} \underline{A}\underline{C} \rightarrow Y &&
    f \inside h: \underline{D}\underline{A}\underline{E}Y\underline{F} \rightarrow Z &&
    g \inside h: \underline{D}X\underline{E}\underline{B}X\underline{C}\underline{F} \rightarrow Z
\end{align*}
$\inside$ is associative:
$
    (f \inside g) \inside h = f \insideref[X_2] (g \inside h) 
    : \underline{D}X\underline{E}\underline{B}\underline{A}\underline{C}\underline{F} \rightarrow Z
$, and composing along parallel edges respects the following interchange law:
\begin{align*}
    f \insideref[X_1] (g \inside h) = g \inside (f \inside h) = (\underline{id_{D}}\ f\ \underline{id_{E}}\ g\ \underline{id_{F}})\inside h &
    : \underline{D}\underline{A}\underline{E}\underline{B}X\underline{C}\underline{F} \rightarrow Z
\end{align*}
\end{definition}
A \emph{frame} is then a morphism of a multicategory whose objects are the hom-sets of some other category.\footnote{For this to be well-defined, the categories involved must be small.} When we can also define the hom-sets of a multicategory, we can construct a frame that itself acts on frames, and iterate this construction.

\paragraph{}
We may now return to defining the sub-signature $\mathcal{H}_N$ that DisCoCirc grammars use.
In this case, $\mathcal{H}_N$ is a monoidal signature (that defines base objects and base morphisms), with an additional set of base frames defined for each $n \leq N$. Each item in the signature (other than the objects) are indexed by a particular atom.
The frames at level $n$ are defined over lists of hom-sets of level $n-1$. The base level $0$ has as objects the hom-sets of the freely generated monoidal category from the base boxes and morphisms.

Expanding this signature into a category then requires some extra items to remain coherent. The two ways of moving between levels are to either `raise' a morphism to the level above as a state, or to compute the result of a higher morphism as applied to a particular input. We allow these morphisms to be generated by the signature rather than requiring them to be specified explicitly. In the following, we shall write $(\underline{A} \rightarrow_n B)$ to denote the $n$th level hom-set from $\underline{A}$ to $B$. Note that the base hom-sets of the monoidal level are written without a level index: $(A \rightarrow B)$.
\begin{itemize}
    \item [] \textbf{Raising} Given a morphism $f: \underline{A} \rightarrow_n B$ of level $n$, we construct a state $f^+: [] \rightarrow_{n+1} (\underline{A} \rightarrow_n B)$ at level $n+1$.
    
    \item [] \textbf{Currying} Given $f: \underline{A}(\underline{X} \rightarrow_{n-1} B)\underline{C} \rightarrow_n (\underline{Y} \rightarrow_{n-1} Z)$ at level $n$ and $g:\underline{X} \rightarrow_{n-1} B$ at level $n-1$, we can curry the morphisms by composing $f$ with $g^+$ to obtain $f_{g}: \underline{A}\underline{C} \rightarrow_{n} (\underline{Y} \rightarrow_{n-1} Z) = g^+ \inside f$. 
    
    \item [] \textbf{Lowering} 
    Given $f: \underline{(\underline{A_i} \rightarrow_{n-1} B_i)}_i \rightarrow_n (\underline{Y} \rightarrow_{n-1} Z)$ at level $n$ and a series of $g_i:\underline{A_i} \rightarrow_{n-1} B_i$ at level $n-1$, we can fill all the inputs of $f$ to a state $f_{\underline{g}}: [] \rightarrow_n (\underline{Y} \rightarrow_{n-1} Z)$, which can be lowered to a morphism at level $n-1$: $f^-_{\underline{g}}: \underline{Y} \rightarrow_{n-1} Z$.
\end{itemize}
The morphisms at level $n$ are then free compositions of the base morphisms at level $n$, raised level $n-1$ morphisms, and lowered level $n+1$ morphisms, under the appropriate `inside' composition $\inside$.

\begin{figure}[h]
    \centering
    \begin{subfigure}[b]{0.3\linewidth}
        \centering
        \includegraphics[scale=0.5]{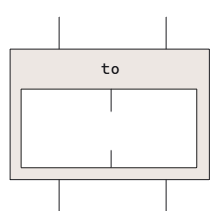}
        \caption{A (level $0$) frame.}
    \end{subfigure}\hfill
    \begin{subfigure}[b]{0.3\linewidth}
        \centering
        \includegraphics[scale=0.5]{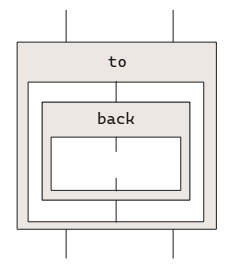}
        \caption{Composition at level $0$.}
    \end{subfigure}\hfill
    \begin{subfigure}[b]{0.3\linewidth}
        \centering
        \includegraphics[scale=0.5]{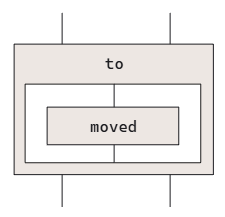}
        \caption{Lowering to the monoidal level. }
    \end{subfigure}
    \caption{
    Example frames encountered in bAbI~6. Their type signatures are as follows:
    (a) $\mathsf{to}: (\mathsf{N} \rightarrow \mathsf{N}) \rightarrow_0 (\mathsf{N} \otimes \mathsf{N} \rightarrow \mathsf{N} \otimes \mathsf{N})$;
    (b) $\mathsf{back} \inside \mathsf{to}: (\mathsf{N} \rightarrow \mathsf{N}) \rightarrow_0 (\mathsf{N} \otimes \mathsf{N} \rightarrow \mathsf{N} \otimes \mathsf{N})$;
    (c) $\mathsf{to}_{\mathsf{moved}}: \mathsf{N} \otimes \mathsf{N} \rightarrow \mathsf{N} \otimes \mathsf{N}$.
    }
    \label{fig:app/compositionality/drawing-frames-level0}
\end{figure}

\begin{figure}[h]
    \centering
    \begin{subfigure}[b]{0.3\linewidth}
        \centering
        \includegraphics[scale=0.6]{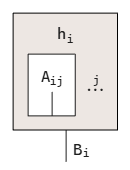}
        \caption{Level $n-1$}
    \end{subfigure}
    \hspace{1em}
    \begin{subfigure}[b]{0.3\linewidth}
        \centering
        \includegraphics[scale=0.6]{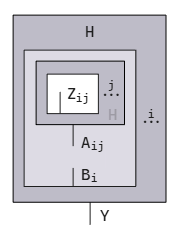}
        \caption{Level $n$}
    \end{subfigure}
    \hspace{1em}
    \begin{subfigure}[b]{0.3\linewidth}
        \centering
        \includegraphics[scale=0.6]{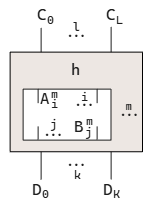}
        \caption{Level $0$}
    \end{subfigure}
    \caption{Drawing frames with varying amounts of detail.
    (a) $h_i: \underline{A_i}_j \rightarrow_{n-1} B_i$. Minimal detail; no further structure is assumed on the types $\underline{A_i}_j$ and $B_i$.
    (b) $H: \underline{(\underline{A_i}_{j_i} \rightarrow_{n-1} B_i)}_i \rightarrow_n (\underline{Z_0}_{j_0}\cdots\underline{Z_I}_{j_I} \rightarrow Y)$. the types are expanded into frames of level $n-1$.
    (c) $h: \underline{(\mathbf{A}^m_{\otimes_I} \rightarrow \mathbf{B}^m_{\otimes_J})}_m \rightarrow_0 (\mathbf{C}_{\otimes_L} \rightarrow \mathbf{D}_{\otimes_K})$. Recall that level $0$ frames are defined on the hom-sets of a monoidal category. We may hence chose to depict this structure as part of the frame. For simpler notation, we write monoidal composites as $\mathbf{A}_{\otimes_I}:= \otimes_{i=0}^I A_i$.
    As a visual aid, we shade $n-1$ level frames in light grey, and $n$ level frames in darker grey. Additionally, we distinguish the inside holes from the outer shape of the higher order frames with extra shading of the inside holes, as with the depiction of $H$ in (b). 
    Ultimately, the reason to draw them as boxes with holes is in order to visualise the source and target objects explicitly - the details in (a) are likely better represented by a box with multiple wires coming out.
    }
    \label{fig:app/compositionality/drawing-frames-detail}
\end{figure}

\begin{figure}[h]
    \centering
    \begin{subfigure}[b]{0.5\linewidth}
        \centering
        \includegraphics[scale=0.6]{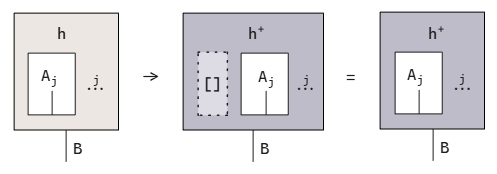}
        \caption{Raising.}
    \end{subfigure}
    \hspace{1em}
    \begin{subfigure}[b]{0.4\linewidth}
        \centering
        \includegraphics[scale=0.6]{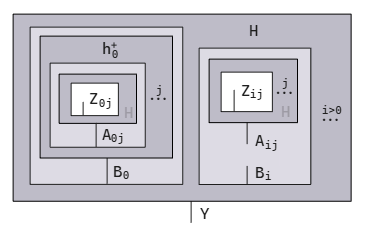}
        \caption{Currying.}
    \end{subfigure}
    \\\vspace{2em}
    \begin{subfigure}[b]{\linewidth}
        \centering
        \includegraphics[scale=0.6]{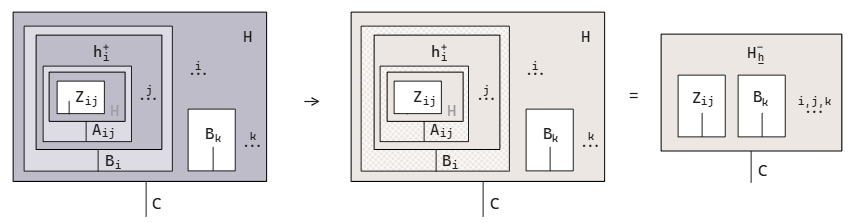}
        \caption{Lowering.}
    \end{subfigure}
    \caption{Depicting the level changing operations as diagrams. The frames displayed have the following signatures:
    (a) $h^+:[] \rightarrow (\underline{A}_j \rightarrow_{n-1} B)$;
    (b) $(h^+ \inside H): \underline{(\underline{A}_{j_i} \rightarrow_{n-1} B_i)}_{i>0} \rightarrow_n (\underline{Z}_{j_1}\cdots\underline{Z}_{j_I} \rightarrow Y)$;
    (c) $H^-_{\underline{h}}: \underline{Z_0}_{j_0}\cdots\underline{Z_I}_{j_I}\underline{B_{}}_k \rightarrow C)$.
    As a visual aid, we shade $n-1$ level frames in light grey, and $n$ level frames in darker grey. Additionally, we distinguish the inside holes from the outer shape of the higher order frames with extra shading of the inside holes, as with the depiction of $H$ in (b).}
    \label{fig:app/compositionality/drawing-frames-changing-levels}
\end{figure}

\paragraph{Drawing frames}
We draw frames as boxes with holes. \autoref{fig:app/compositionality/drawing-frames-detail} illustrates how to draw a single frame, while \autoref{fig:app/compositionality/drawing-frames-changing-levels} depicts the raising, currying and lowering operations. Finally we display some examples encountered for bAbI~6 in \autoref{fig:app/compositionality/drawing-frames-level0}. Effectively, frames are drawn as boxes with holes, such that their composition $\inside{}$ goes into the page by nesting frames inside each other. The holes inside the frame correspond to the inputs, while the outer shape reflects the target shape of the frame.
Valid compositions must place frames fully inside a single hole - this ensures composition occurs along only one type at a time.

\FloatBarrier

\paragraph{Collapsing the layers}
The categorical machinery used to describe frames may seem overly involved for a practical setup - in particular it may be difficult to find an appropriate meaning category that has sufficient higher order structures. For this reason, in practice, we typically require that the frame hierarchy can be collapsed down to the monoidal level. There are many ways to do this, each with their own benefits and drawbacks. Here we shall highlight two options. The first is to recognise that if the base category is closed, then hom-sets are themselves captured as objects, such that frames can be directly expressed as morphisms. Iterating this conversion then collapses the entire hierarchy.

\begin{figure}[h]
    \centering
    \begin{subfigure}[b]{0.3\linewidth}
        \centering
        \includegraphics[scale=0.5]{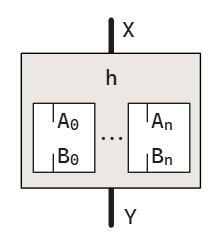}
        \caption{Frame.}
    \end{subfigure}
    \hspace{1em}
    \begin{subfigure}[b]{0.35\linewidth}
        \centering
        \includegraphics[scale=0.5]{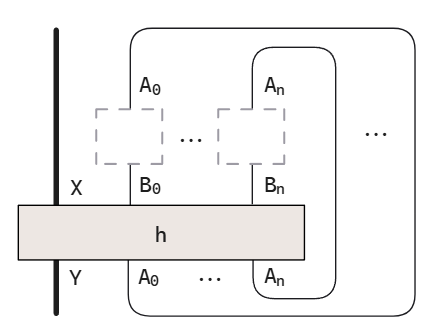}
        \caption{\textsf{Onion Rings}.}
    \end{subfigure}
    \hspace{1em}
    \begin{subfigure}[b]{0.25\linewidth}
        \centering
        \includegraphics[scale=0.5]{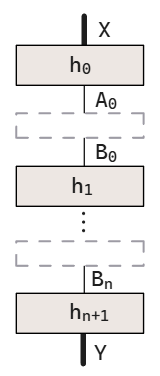}
        \caption{\textsf{Sandwich}.}
    \end{subfigure}
    \caption{Collapsing frames to the base level. (a) An example frame. (b) Using closure, morphisms can be expressed as states via the addition of cups and caps. This can be extended to frames by appropriately bending the wires. We call this the \textsf{Onion Rings} due to the shape of the added cups and caps. (c) Each frame is expanded into a series of boxes, with a layer between each hole. The frame hence becomes the `bread' in a multi-tiered \textsf{Sandwich}. The outside wires are drawn in bold as a visual aid, while the holes are marked with a dashed box. When filled, the diagrams become valid monoidal (closed) diagrams.}
    \label{fig:frame_expansion}
\end{figure}

The second is to approximate the frames in the base category by expanding them in terms of correlated boxes. This method is less general, as it does not provide a way to express a frame itself; it is instead a way to capture the action of a frame on its insides within the base category.
The approaches are summarised pictorially in \autoref{fig:frame_expansion}.

\begin{figure}[h]
    \centering
    \begin{subfigure}[b]{0.45\textwidth}
        \centering
        \includegraphics[width=\linewidth]{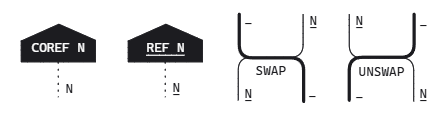}
        \caption{Co-reference rules.}
    \end{subfigure}
    \hfill
    \begin{subfigure}[b]{0.45\textwidth}
        \centering
        \includegraphics[width=\linewidth]{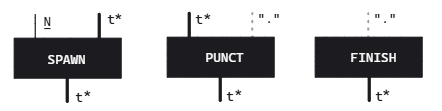}
        \caption{Noun and sentence introduction rules.}
    \end{subfigure}
    \\\vspace{1em}
    \begin{subfigure}[b]{0.45\textwidth}
        \centering
        \includegraphics[width=\linewidth]{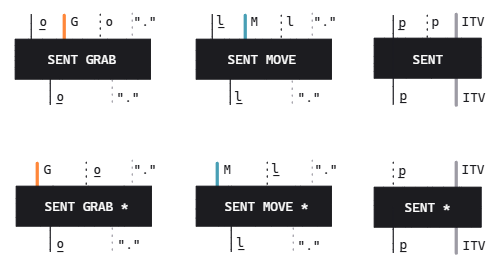}
        \caption{Sentence introduction rules.}
    \end{subfigure}
    \hfill
    \begin{subfigure}[b]{0.45\textwidth}
        \centering
        \includegraphics[width=\linewidth]{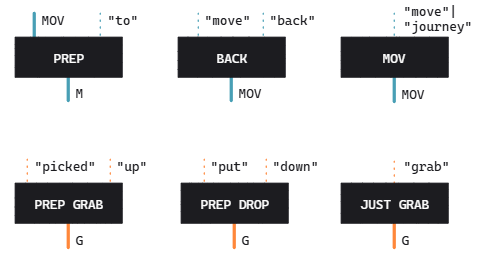}
        \caption{Intra-sentence rules.}
    \end{subfigure}
    \\\vspace{1em}
    \begin{subfigure}[b]{0.45\textwidth}
        \centering
        \includegraphics[width=0.9\linewidth]{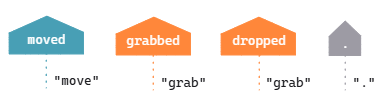}
        \caption{Base rules. Not all are shown explicitly, consult \autoref{tab:babi6_vocabulary} of \autoref{app:datasets} for a full vocabulary list by type.}
    \end{subfigure}
    \hfill
    \begin{subfigure}[b]{0.45\textwidth}
        \begin{align*}
            \mathsf{p}   & = \mathsf{Alice}|\mathsf{Bob}|... &
            \mathsf{N} & =  \mathsf{p}|\mathsf{l}|\mathsf{o}\\
            \mathsf{l}   & = \mathsf{park}|\mathsf{kitchen}|... &
            \mathsf{\_}  & = \mathsf{N}|\mathsf{t^*}\\
            \mathsf{o}   & = \mathsf{milk}|\mathsf{slippers}|... &
            \mathsf{ITV} & = \mathsf{M}|\mathsf{G}            
        \end{align*}
        \caption{Type abbreviations.}
    \end{subfigure}
    \caption{
    Generator shapes for the bAbI6 syntax category of DisCoCirc models. 
    }
    \label{fig:compositionality/babi6_grammar_rules}
\end{figure}

\begin{figure}[h]
    \centering
    \begin{subfigure}[b]{0.6\textwidth}
        \centering
        \includegraphics[scale=0.6]{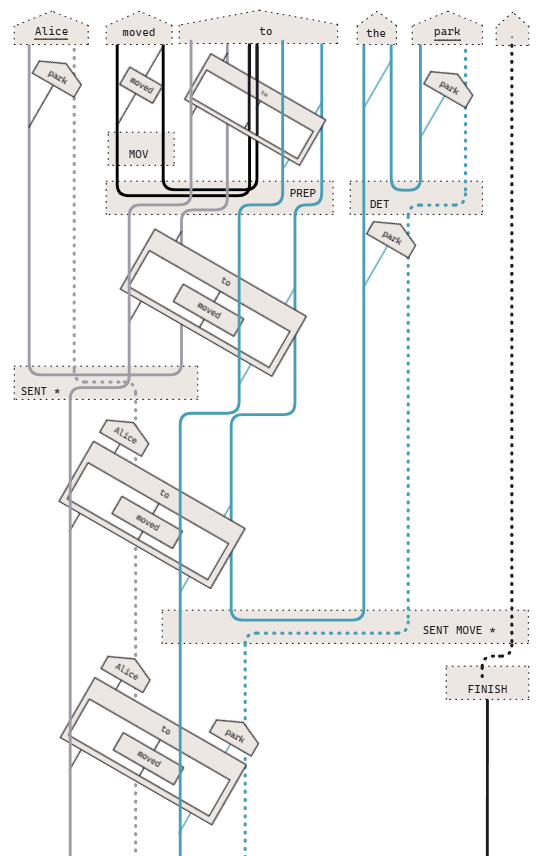}
        \caption{}
    \end{subfigure}
    \hfill
    \begin{subfigure}[b]{0.35\textwidth}
        \centering
        \includegraphics[scale=0.5]{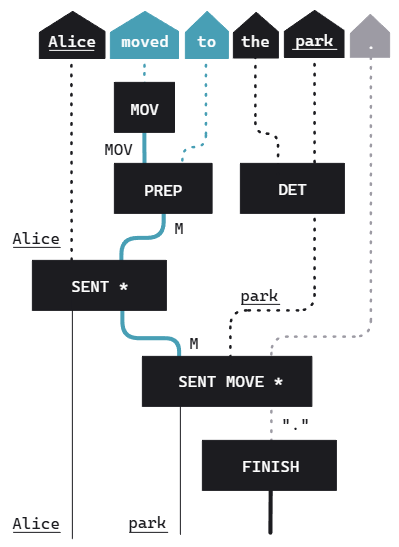}
        \\\vspace{8em}
        \caption{}
    \end{subfigure}
    \caption{(a) Informal visualisation of the rules as wiring operations for the example derivation given in \autoref{fig:compositionality/discocric-parsing-ex}, and summarised in (b). The derivation is continued in \autoref{fig:compositionality/app/discocirc_parsing_2}.
    The dashed boxes represent the rule morphisms, while rotated boxes depict the diagrams that flow through the wires of the grammar category.
    The thick wires depict the borders of the grammar category wires - the cups present in the rule boxes are to be understood as wiring the inside boxes along the given edge. This is ultimately a sequential composition of the morphisms.
    }
    \label{fig:compositionality/app/discocirc_parsing_1}
\end{figure}

\begin{figure}[h]
    \centering
    \begin{subfigure}[b]{0.6\textwidth}
        \centering
        \includegraphics[scale=0.6]{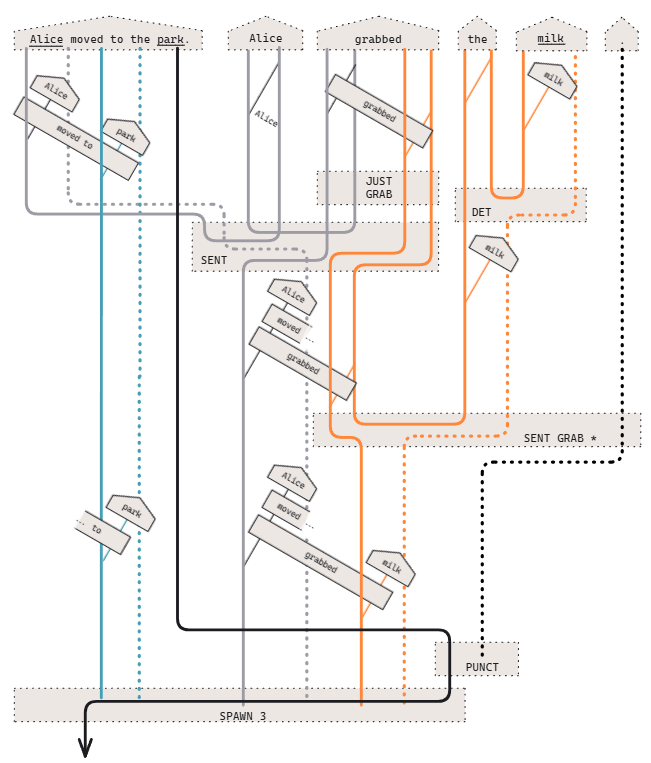}
        \caption{}
    \end{subfigure}
    \hfill
    \begin{subfigure}[b]{0.3\textwidth}
        \centering
        \includegraphics[scale=0.45]{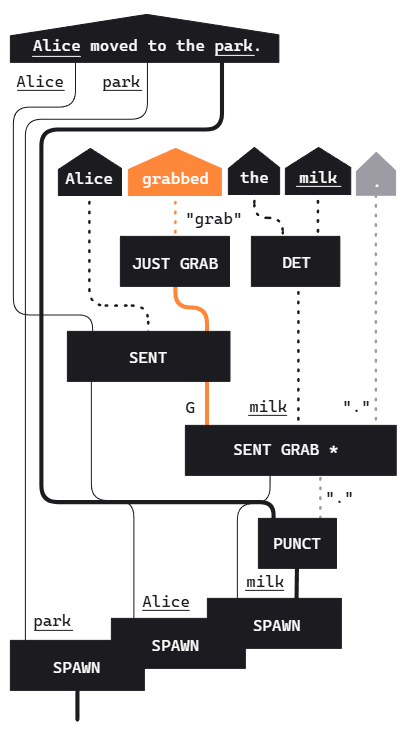}
        \\\vspace{5em}
        \caption{}
    \end{subfigure}
    \caption{(a) The second part of the derivation begun in \autoref{fig:compositionality/app/discocirc_parsing_1}, visualising the rules as wiring operations for the example derivation given in \autoref{fig:compositionality/discocric-parsing-ex} and summarised in (b).
    The dashed boxes represent the rule morphisms, while rotated boxes depict the diagrams that flow through the wires of the grammar category.
    The thick wires depict the borders of the grammar category wires - the cups present in the rule boxes are to be understood as wiring the inside boxes along the given edge. This is ultimately a sequential composition of the morphisms.
    }
    \label{fig:compositionality/app/discocirc_parsing_2}
\end{figure}

\paragraph{Back to the grammar}
Much like with DisCoCat, the grammatical morphisms of DisCoCirc setups can be entirely expressed in terms of structural morphisms of the wire category, $\syntaxcat[{\grammarsig[H]_N}]$. Thus, since any derivation in $\syntaxcat$ is properly a state, we can equivalently consider the diagram of $\syntaxcat[{\grammarsig[H]_N}]$ selected by this state as the parsed diagram of a sentence.
Practically, what this means is that the rule level of the grammar only serves to determine which diagrams are to be considered valid, such that we can simplify the grammatical signature to $\grammarsig[H]_N$ and use $\syntaxcat[{\grammarsig[H]_N}]$ as the syntax category. A full list of the rules in $\grammarsig$ is provided in \autoref{fig:compositionality/babi6_grammar_rules}.
We informally visualise how these rules become internal wirings for an example text derivation in \autoref{fig:compositionality/app/discocirc_parsing_1} and \autoref{fig:compositionality/app/discocirc_parsing_2}.


\clearpage
\section{Dataset}
\label{sec:babi-datasets}
\label{app:datasets}

\begin{figure}[h]
    \centering
    \begin{subfigure}[b]{0.3\linewidth}
        1. \word{Clara went to the bathroom.}\\
        2. \word{Bill travelled to the kitchen.}\\
        3. \word{Bill picked up the football.}\\
        4. \word{Clara moved back to the office.}\\\\
        Q. \wordq{Is Bill in the kitchen?}\\
        \caption{Example bAbI task 6 story}
        \label{tab:datasets/example-stories}
    \end{subfigure}
    \hspace{3em}
    \begin{subfigure}[b]{0.35\linewidth}
        \centering
        \includegraphics[width=\linewidth]{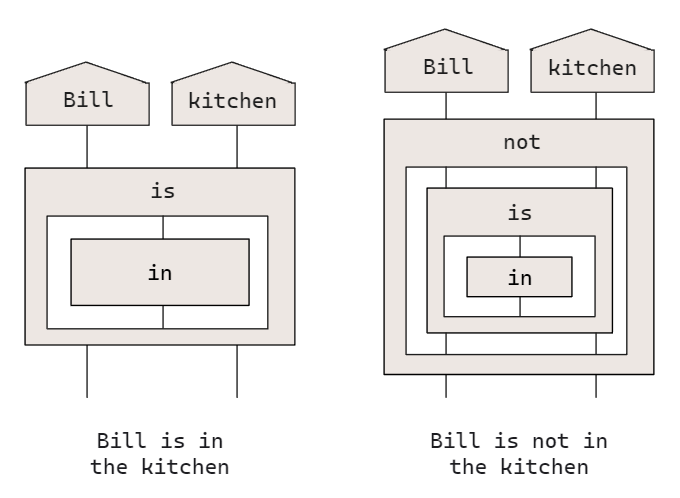}
        \caption{Assertions}
        \label{fig:babi6_assertions}
    \end{subfigure}
    \caption{An example story and pair of assertions as DisCoCirc diagrams. The binary question \sentq{Is Bill in the kitchen?} is decomposed into a pair of assertions covering the two possible answer cases.}
\end{figure}

The bAbI dataset contains higher-order boxes of various shapes. \autoref{tab:datasets/example-stories} displays an example story, \autoref{tab:babi6_vocabulary} displays the vocabulary while \autoref{fig:babi6_assertions} shows a typical question and associated assertions.
For each aspect of compositionality, we generate a dataset designed to test it specifically. 

Note that in each case, we define difficulty \emph{strata}, which are used to track the compositional difficulty of each datapoint and are used to order the data for training and test purposes.

\begin{table}[h]
    \centering
    \begin{tabular}{ll l}
        \toprule
        \rule[-0.5em]{0pt}{1.8em}
         \textbf{Shape} & \textbf{Type} & \textbf{Vocabulary} \\
        \midrule
         Noun & \textsf{P} & \word{Andrew}, \word{Bill}, \word{Clara}, \word{Denise}, \word{Eric}, \word{Fred}, \word{Gillian}, \word{Heidi} \\
              & \textsf{O} & \word{apple}, \word{football}, \word{milk}, \word{slippers} \\
              & \textsf{L} & \word{kitchen}, \word{office}, \word{hallway}, \word{bedroom}, \word{garden}, \word{bathroom}, \word{cinema}, \word{park}\\
        \rule[-0.5em]{0pt}{1.5em}
         ITV & \textsf{move} & \word{moved}, \word{went}\\
             & \textsf{journey} & \word{travelled}, \word{journeyed}\\
             & \textsf{in} & \wordq{in} \\
        \rule[-0.5em]{0pt}{1.5em}
         TV  & \textsf{grab} & \word{discarded}, \word{dropped}, \word{left}, \word{grabbed},  \word{took}, \word{got} \\
             & \textsf{put}, \textsf{picked} &  \word{put}, \word{picked}\\
        \rule[-0.5em]{0pt}{1.5em}
         HO (1, 1) & \textsf{back} & \word{back} \\
        \rule[-0.5em]{0pt}{1.5em}
         HO (2, 1) & \textsf{to}, \textsf{is} & \word{to}, \wordq{is} \\
        \rule[-0.5em]{0pt}{1.5em}
         HO (2, 2) & \textsf{up}, \textsf{down}, \textsf{not} & \word{up}, \word{down}, \wordq{not} \\
        \bottomrule
    \end{tabular}
    \caption{Vocabulary present in the bAbI 6 task, by DisCoCirc shape and syntax type. Words exclusively used in the questions are underlined.}
    \label{tab:babi6_vocabulary}
\end{table}

\subsection{Productivity}
\begin{figure}[h]
    \centering
    \begin{subfigure}[b]{0.47\linewidth}
        \centering
        \includegraphics[width=\linewidth]{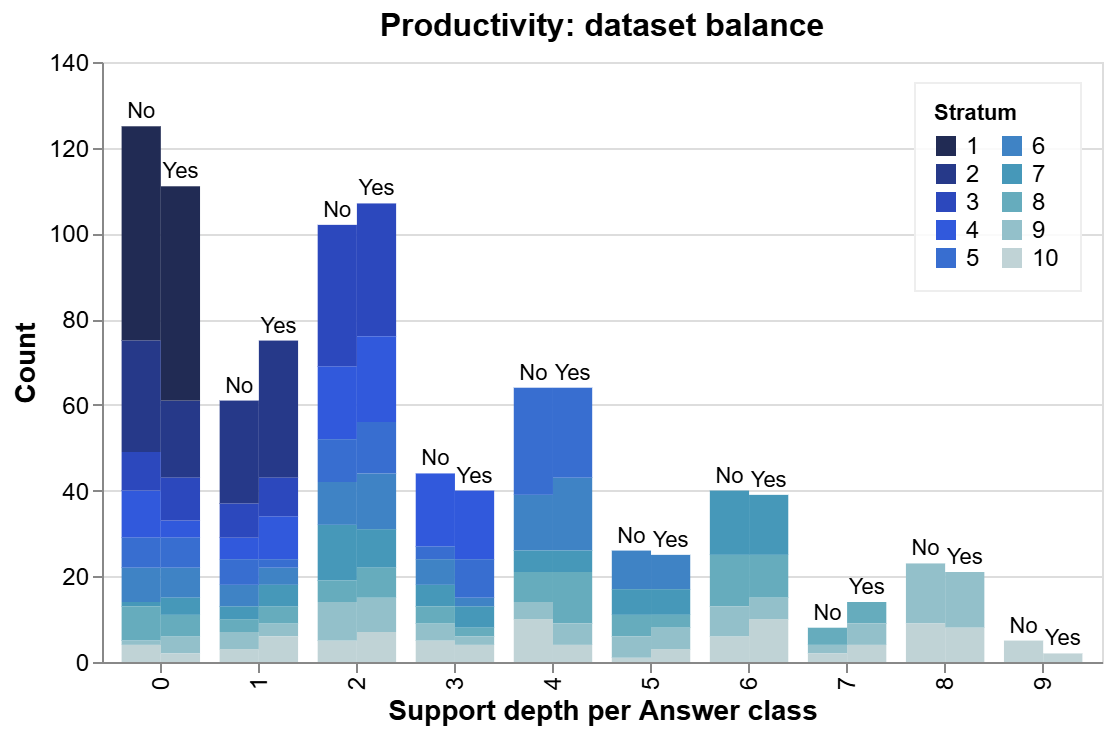}
        \caption{}
        \label{fig:datasets/babi/prod-train-strata}
    \end{subfigure}
    \hfill
    \begin{subfigure}[b]{0.47\linewidth}
        \centering
        \includegraphics[width=\linewidth]{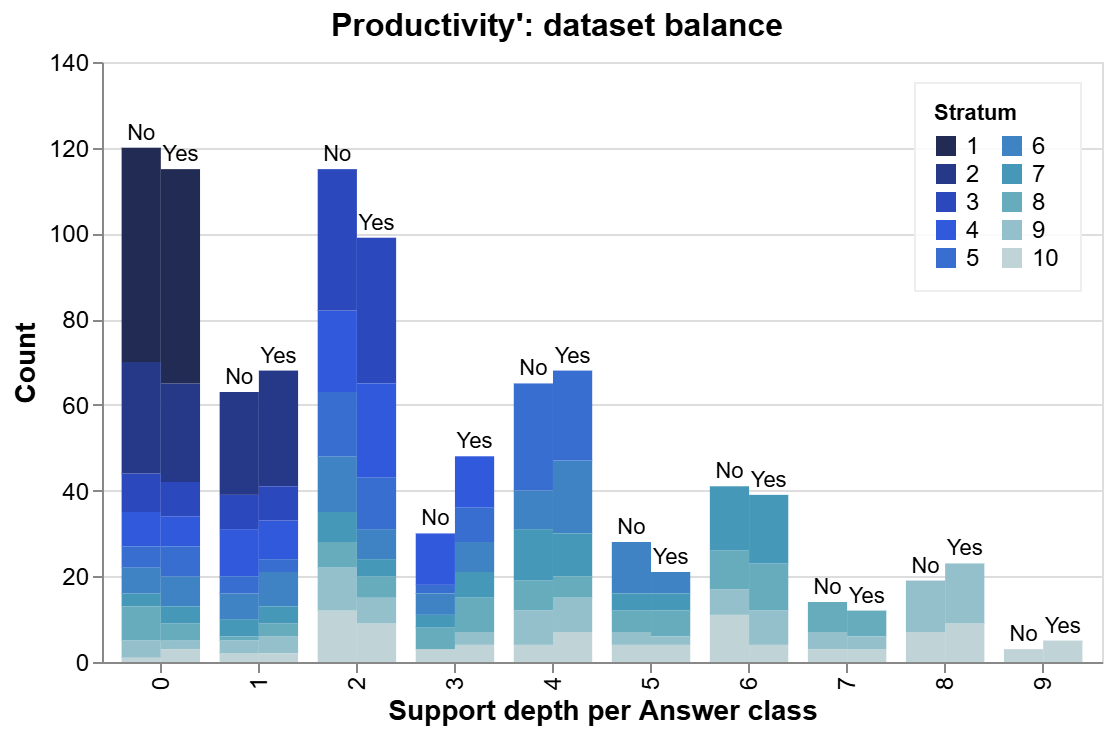}
        \caption{}
        \label{fig:datasets/babi/prod-train-strata-test}
    \end{subfigure}
    \caption{
    Visualising the balancing for the
    (a) \textit{Productivity} and (b) \textit{Productivity'} datasets. 
    The stratum in this case is the number of sentences in each story, represented by colour. 
    The distribution of the number of sentences is plotted according to the support depth and answer class. 
    }
    \label{fig:datasets/babi/prod-dataset-balance}
\end{figure}

\begin{figure}[h]
    \centering
    \begin{subfigure}[b]{0.4\linewidth}
        \centering
        \includegraphics[width=\linewidth]{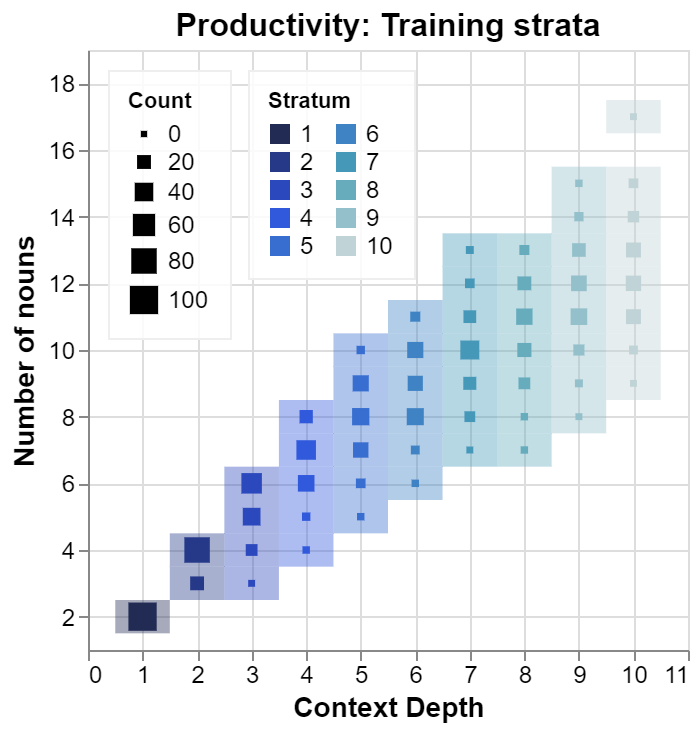}
        \caption{}
        \label{fig:datasets/babi/prod-balance}
    \end{subfigure}
    \hspace{4em}
    \begin{subfigure}[b]{0.4\linewidth}
        \centering
        \includegraphics[width=\linewidth]{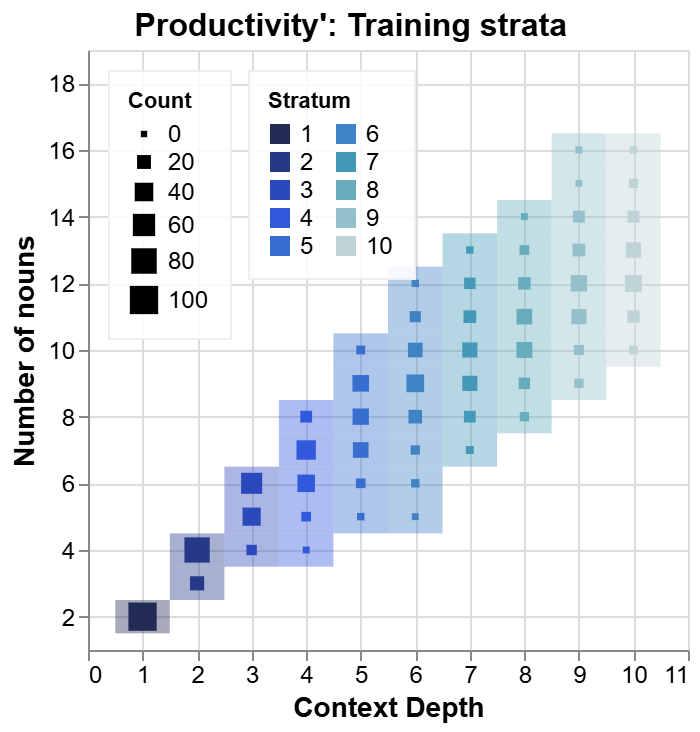}
        \caption{}
        \label{fig:datasets/babi/prod-balance-test}
    \end{subfigure}
    \caption{Visualising the distribution of story size, according to the number of sentences and number of nouns, for the (a) \textit{Productivity} and (b) \textit{Productivity'} datasets. The stratum in this case is the number of sentences for each story, represented by colour. In each case the region covered by each stratum is shaded, with the size of the points corresponding to the number of datapoints sampled for that shape combination.}
    \label{fig:datasets/babi/prod-dataset-strata}
\end{figure}

The productivity dataset matches the original bAbI 6 data most closely, so we will take this dataset as a base from which to derive others where needed. The dataset is generated procedurally by sampling sentences one at a time until a desired depth is reached. We then check that the candidate entry matches any further selection criteria, and if so add it to the dataset. 
For productivity, we enforced an even split between answer classes, and an even spread of number of sentences and nouns. We also controlled the support depth (the number of sentences between the sentence that provides the answer and the end of the story) to ensure there were sufficient datapoints present at the higher depths, however as the support depth is constrained by the total number of sentences, there are necessarily more entries with a lower support depth. 
We generate two datasets drawn from the same distribution that we split as \textit{Productivity} and \textit{Productivity'}. In most situations, we shall sample both the \textit{Train} and \textit{Test} sets from the \textit{Productivity} dataset, whilst \textit{Productivity'} is only used as an additional evaluation set to validate model performance on unseen elements drawn from a similar distribution to the training set.
We visualise the strata splits and balancing in \autoref{fig:datasets/babi/prod-dataset-strata} and  \autoref{fig:datasets/babi/prod-dataset-balance}.

\FloatBarrier

\subsection{Systematicity}
\begin{figure}[h]
    \centering
    \includegraphics[width=0.8\linewidth]{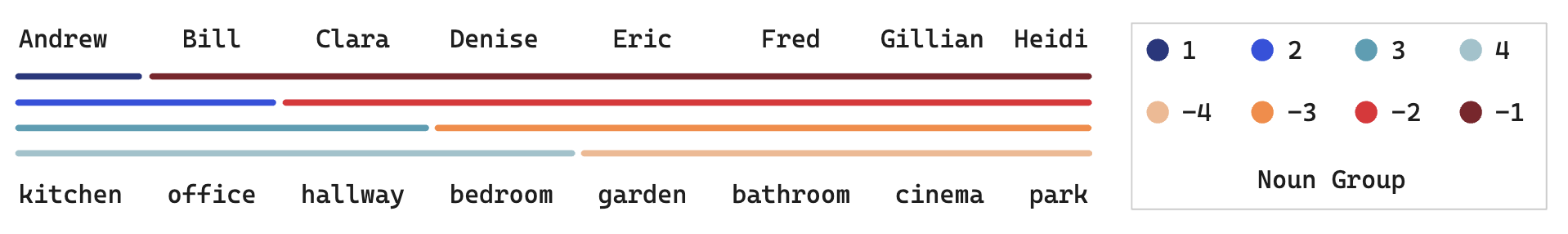}
    \caption{The noun groups used for generating the systematicity dataset. The top and bottom layers identify the people and the locations respectively. The order of the nouns is kept fixed so they can be indexed by number (shown in grey). The groups are coloured, with positive indices coloured in blue and their complements (negative indices) coloured in orange.}
    \label{fig:datasets/babi/syst-groups}
\end{figure}

\begin{figure}[h]
    \centering
    \begin{subfigure}[b]{0.4\linewidth}
        \includegraphics[width=\linewidth]{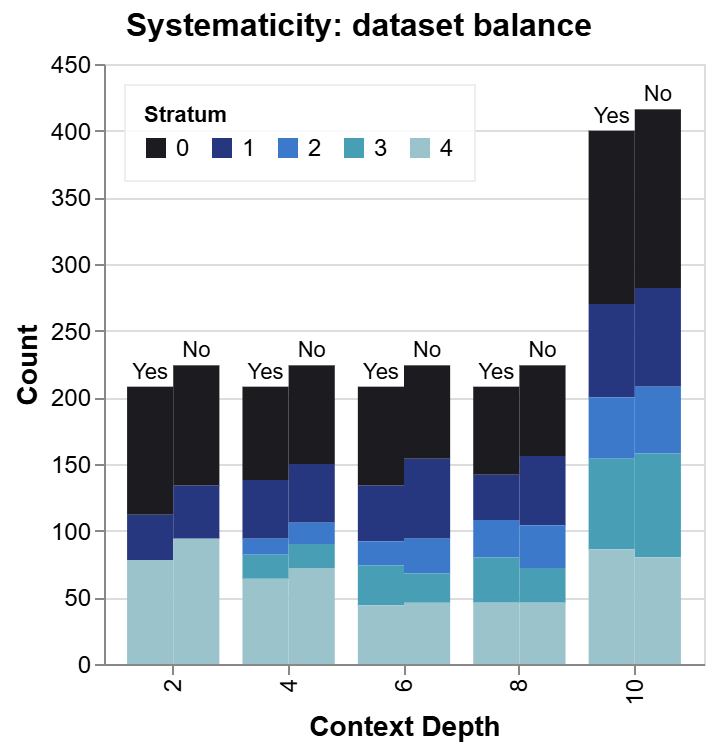}
        \caption{}
        \label{fig:datasets/babi/syst-4-group-balancing_depth-ans}
    \end{subfigure}
    \hspace{4em}
    \begin{subfigure}[b]{0.4\linewidth}
        \includegraphics[width=\linewidth]{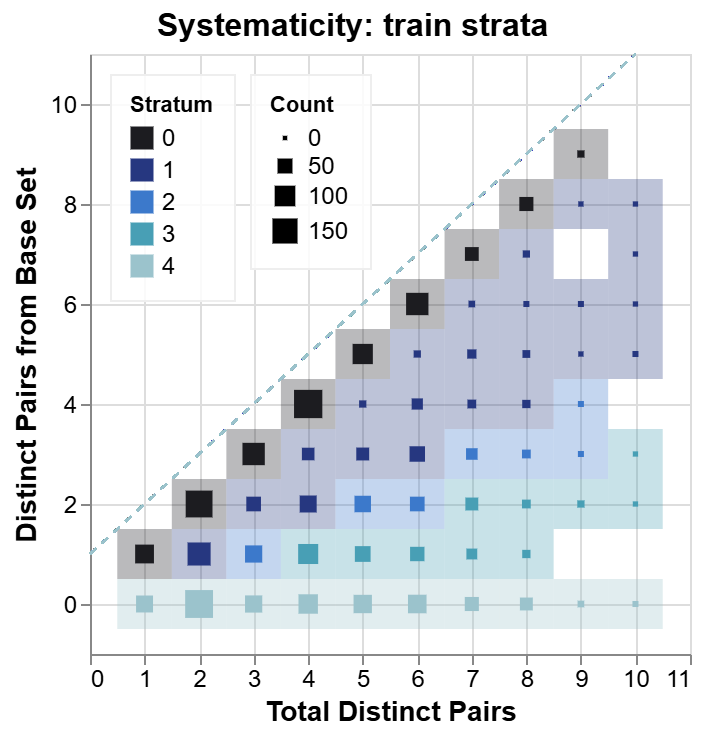}
        \caption{}
        \label{fig:datasets/babi/syst-4-group-balancing_train-strata}
    \end{subfigure}
    
    \caption{Visualising the splits for the systematicity dataset. The object interactions were not considered when counting the systematicity pairs, so some stories contain nouns from groups outside the recorded noun group.
    (a) We plot the stratum balancing across the number of sentences and answer class. Not all strata are available when there are only two sentences. The bars show the cumulative number of points for each depth.
    The stratum for each datapoint is calculated from the total distinct pairs present, and the distinct pairs that occur in the base set, as shown in (b). The regions belonging to each stratum are coloured accordingly. The size of the point plotted in each region captures the number of datapoints present. The dashed line represents the limit where all pairs seen belong to the base set whilst datapoints on the bottom row contain no pairings from the base set.
    }
    \label{fig:datasets/babi/syst4-group-balancing}
\end{figure}

\begin{figure}[h]
    \centering
    \begin{subfigure}[b]{0.47\linewidth}
        \includegraphics[width=\linewidth]{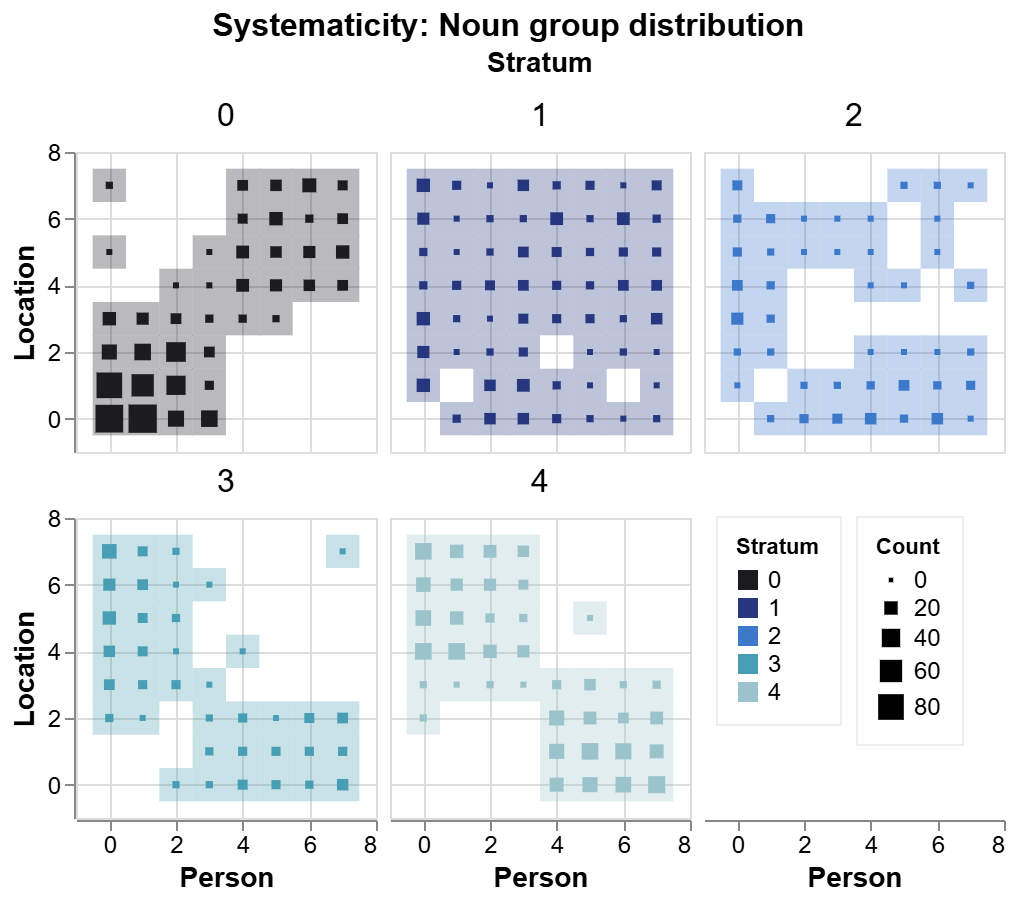}
        \caption{}
    \end{subfigure}
    \hfill
    \begin{subfigure}[b]{0.47\linewidth}
        \includegraphics[width=\linewidth]{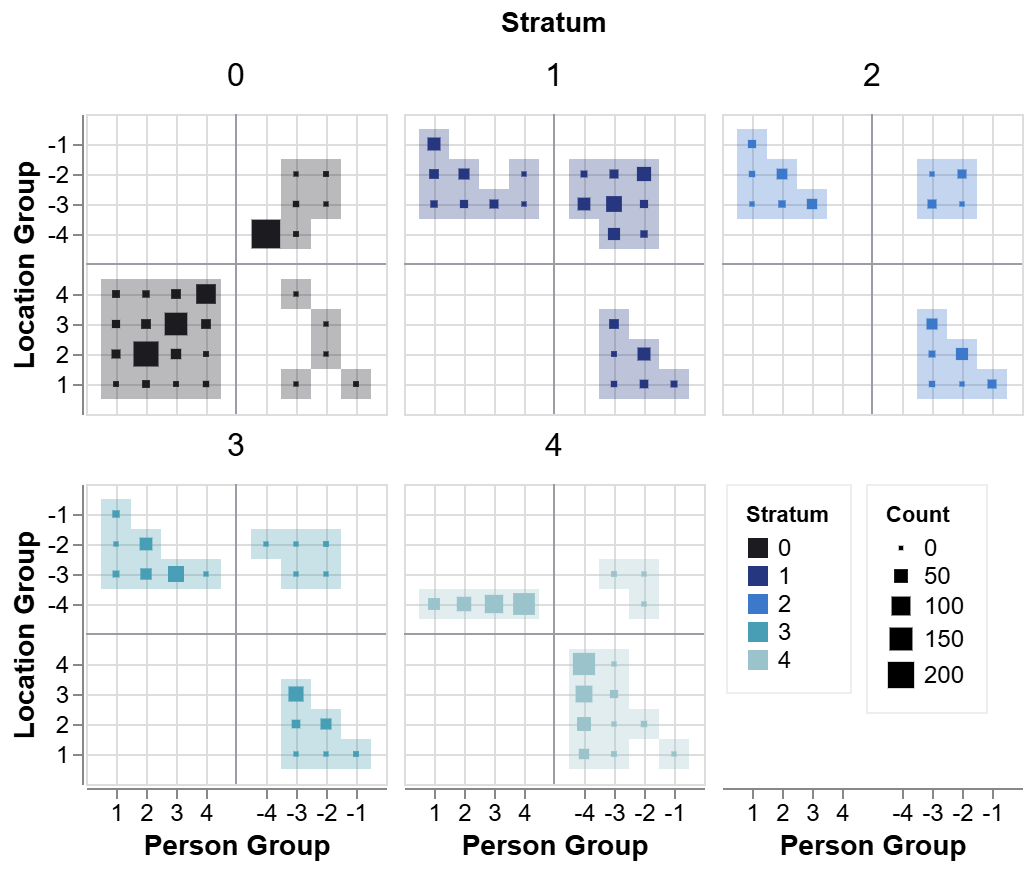}
        \caption{}
    \end{subfigure}
    \caption{
    Visualising the distribution of nouns present in each training stratum for the systematicity dataset. 
    (a) plots which nouns are seen together in the questions, whilst (b) plots which noun groups are present throughout the context, excluding the question. In each case, we shade the region covered by each stratum. The size of the points plotted for each person and location (group) combination represents the number of datapoints available.
    }
    \label{fig:datasets/babi/syst4-group-charac}
\end{figure}

When generating the systematicity dataset, we split the nouns into sub-groups (detailed in \autoref{fig:datasets/babi/syst-groups}), and generated entries limiting the nouns involved in each group. Similarly to productivity, we balanced the generated data according to the number of sentences in the text and answer class (\autoref{fig:datasets/babi/syst4-group-balancing}). We visualise the systematicity splits by inspecting which of the person and location sub-groups were paired by each entry in \autoref{fig:datasets/babi/syst4-group-charac}.
The overall dataset is slightly biased towards \word{no} as a consequence of the base case, in which the base group isolates a single person and location to be paired with the others. For a single person (respectively location), there is only one way for them to be in the location, while there are six ways for them not to be. We chose to prefer including more examples of systematicity in this case than to enforce an exact balance in answer class. The resulting baseline accuracy is $51.6\%$, rather than the expected $50\%$.
Notice that there is some overlap between the noun groups in \autoref{fig:datasets/babi/syst-groups}, such that a given datapoint may fall into multiple groups. In such cases, we assign the most restrictive group available. For example, a datapoint containing \word{Bill} and \word{Clara} will fall into groups 3, 4 and -2, but will be assigned to group 3 as this is the smallest group.

\FloatBarrier

\subsection{Substitutivity}
\begin{figure}[h]
    \centering
    \begin{subfigure}[b]{0.35\linewidth}
        \includegraphics[width=\linewidth]{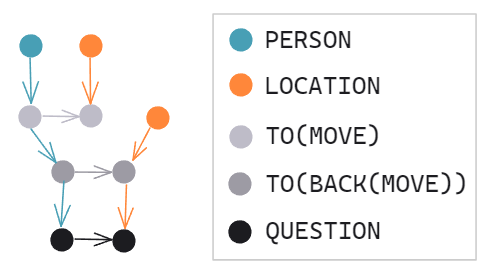}
        \caption{A structure representation}
    \end{subfigure}
    \hspace{4em}
    \begin{subfigure}[b]{0.3\linewidth}
        1. \word{Clara journeyed to the park.} \\
        2. \word{Clara went back to the garden.} \\ 
        Q. \wordq{Is Clara in the garden?}
        \\\\
        1. \word{Bill travelled to the kitchen.} \\
        2. \word{Bill moved back to the office.} \\
        Q. \wordq{Is Bill in the office?}
        \vspace{0.5em}
        \caption{Compatible stories.}
    \end{subfigure}
    \caption{
    A representation of a structure as a coloured directed acyclic graph. The nouns chosen for the question are included in the structure to fix the answer class. Boxes are represented as a sequence of nodes in order to distinguish the subject and object.
    }
    \label{fig:datasets/babi/sub-example}
\end{figure}

The substitutivity dataset was initially derived from the base productivity dataset. For each datapoint, we identified its structure such as that in \autoref{fig:datasets/babi/sub-example}. The structure of each datapoint abstracts away both the particular choice of synonym as well as some symmetries in the ordering of sentences that DisCoCirc diagrams contain by virtue of being symmetric monoidal diagrams.
We then sampled 500 unique structures, balanced across the number of sentences and answer class to form our dataset. For each sampled structure, we first construct the base example by fixing an element of the vocabulary for each verb class. We then generate four further entries from this structure, with varying synonym distances.

\begin{figure}[h]
    \centering
    \begin{subfigure}[b]{0.57\linewidth}
        \includegraphics[width=\linewidth]{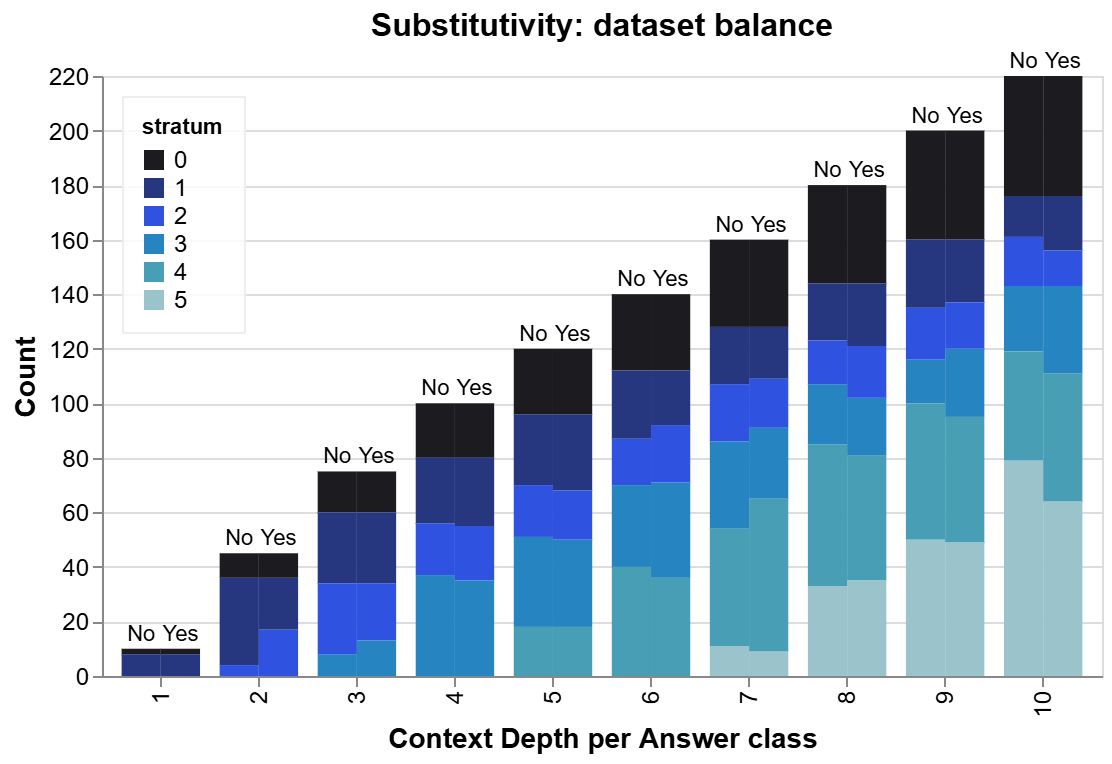}
        \caption{}
        \label{fig:datasets/babi/sub-balancing}
    \end{subfigure}
    \hfill
    \begin{subfigure}[b]{0.37\linewidth}
        \centering
        \includegraphics[width=\linewidth]{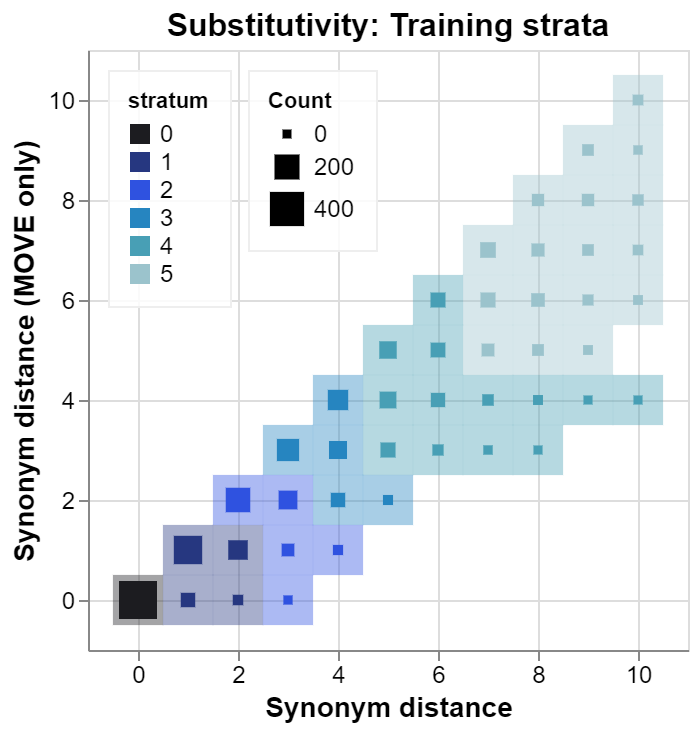}
        \caption{}
        \label{fig:datasets/babi/sub-train-strata}
    \end{subfigure}
    \caption{
    Visualising the substitutivity dataset.
    (a) Visualising the balance of splits for the substitutivity dataset structures according to the the number of sentences and answer class. The splits are not exactly balanced per stratum due to the randomness present in data generation. The stratum is necessarily somewhat correlated with the number of sentences, as the number of synonyms available for replacement depends on the number of sentences in the story.
    (b) Visualising training strata in terms of synonym distances. The region covered by each stratum is shaded, while the size of the point repesents the number of samples available.
    }
    \label{fig:datasets/babi/sub}
\end{figure}

These are then split into tiers according to the synonym distance exhibited by the datapoint. In order to maintain somewhat balanced tiers, we consider both the absolute number of synonym replacements needed, as well as the number of replacements for movement type verbs only, as these are the ones that provide relevant information for the final answer. The splits are visualised in  \autoref{fig:datasets/babi/sub}.

\FloatBarrier

\subsection{Overgeneralisation}
\begin{figure}[H]
    \centering
    \includegraphics[width=0.7\linewidth]{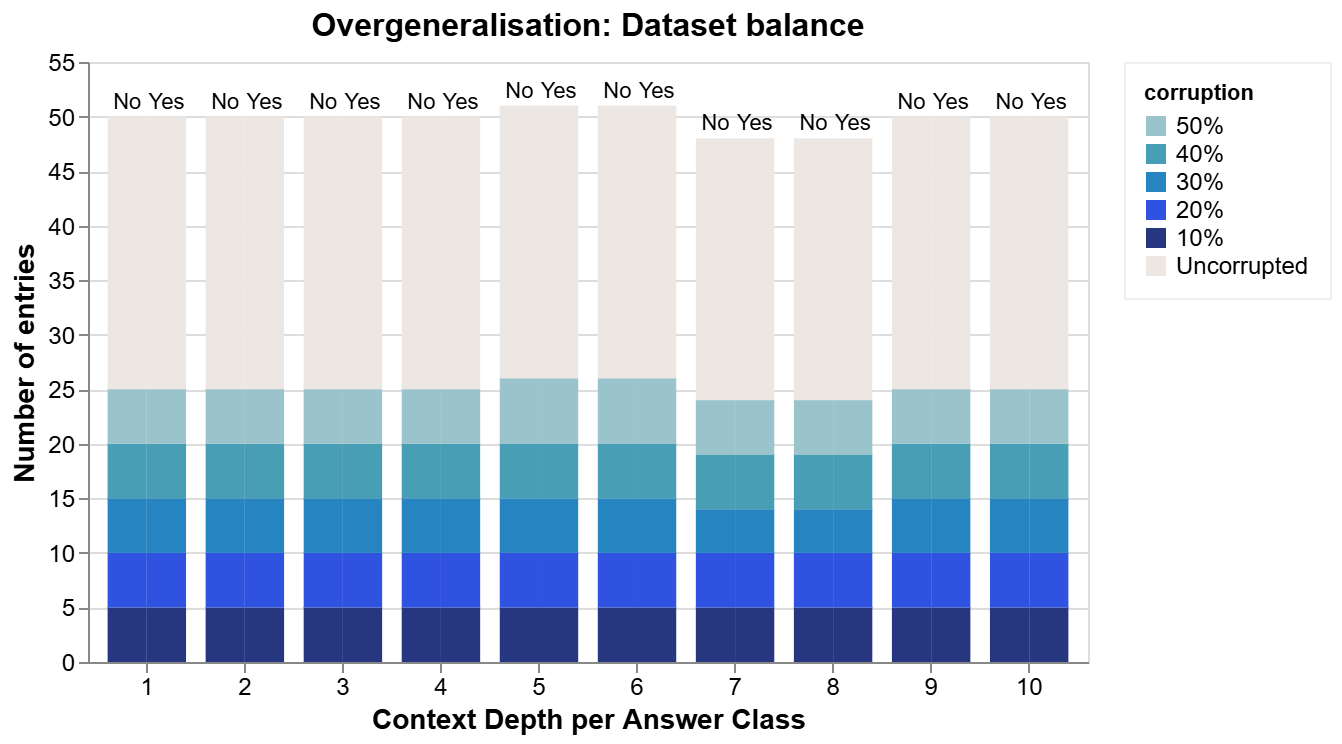}
    \caption{The splits for the Overgeneralisation datasets according to the number of sentences and answer class. The percentage of corrupted datapoints is represented by the colour of the bar, and is cumulative.}
    \label{fig:datasets/babi/overgen-balance}
\end{figure}

For overgeneralisation, we take the productivity dataset as the uncorrupted dataset. We then generate some further datasets with $10\%, 20\%, 30\%$ and $50\%$ corruption. Note that all corrupted datapoints in the $10\%$ dataset are also corrupted in the $20\%$ dataset, and so on. The datapoints to be corrupted were sampled per answer class and number of sentences to avoid biasing the dataset, as shown in \autoref{fig:datasets/babi/overgen-balance}.

\FloatBarrier

\clearpage
\section{Training}
\label{app:training}
Here we provide details of the exact hyperparameters used for each of the trained models.

\subsection{Productivity, Systematicity and Substitutivity}
We performed some initial manual tuning on the quantum models, settling on a learning rate of $0.0005$ and a single batch for the productivity dataset, which we then transferred to the other datasets without requiring further tuning. Training runs lasted between 20-75 epochs for the quantum models, and from 50 to 200 epochs for the neural models. We document the final parameters for the quantum models in \autoref{tab:training/babi/hyperparams-final-quantum}. The quantum models were run for fewer epochs largely due to resource constraints - most of the neural architectures were much faster to run, making more epochs feasible in the time allocated.

\begin{table}[h]
    \centering
    \small{
    \begin{tabular}{l rrr}
        \toprule
        \bf{Hyperparameter}  & \bf{Productivity} & \bf{Substitutivity}  & \bf{Systematicity} \rule[-0.5em]{0pt}{1.5em}\\
        \midrule
        Learning rate ($\times 10^{-3}$) & 5.00 & 5.00 & 5.00 \rule[-0.5em]{0pt}{1.5em}\\
        Batch size & 1 & 1 & 1 \rule[-0.5em]{0pt}{1.5em}\\
        Max epochs & 75 & 20 & 30 \rule[-0.5em]{0pt}{1.5em}\\
        \bottomrule
    \end{tabular}
    }
    \caption{
    Final hyperparameters selected for the quantum models.
    }
    \label{tab:training/babi/hyperparams-final-quantum}
\end{table}

For the neural models, we perform extensive hyperparameter tuning, as summarised in \autoref{tab:training/babi/hyperparams}, according to the options presented in \autoref{tab:models/neural/functor_choices}. The final values selected are summarised in \autoref{tab:training/babi/hyperparams-final}.
In cases where no hyperparameter tuning was used, we run the training with 5 different initialisations, keeping the hyperparameters constant, and select the best model out of these 5 runs, otherwise we selected the best model obtained from the tuning runs. For some of the initial tuning runs, the max epochs was set to 200, however we noticed that selected models were rarely from the last 100 epochs, so capped this further for the final training runs.

\begin{table}[h]
    \centering
    \begin{tabular}{l r r}
        \toprule
        \bf{Hyperparameter}  & \bf{Type} & \bf{Options} \rule[-0.5em]{0pt}{1.5em} \\
        \midrule
        Learning rate & range & 0.00005 - 0.005 \rule[-0.5em]{0pt}{1.5em}\\
        Batch size & choice & 1, 4, 8, 16, 32, 64, 128 \rule[-0.5em]{0pt}{1.5em}\\
        Wire dimension & choice & 2, 3, 6, 12, 24, 36, 50 \rule[-0.5em]{0pt}{1.5em}\\
        Hidden layers & choice & $\mathsf{Linear}$, $\mathsf{Flat}$, $\mathsf{Hidden}(1)$, $\mathsf{Hidden}(2)$, $\mathsf{Hidden}(1, 1)$, $\mathsf{Hidden}(2, 2)$ \rule[-0.5em]{0pt}{1.5em} \\
        Activation & choice & \textsf{ReLu}, \textsf{Mish} \rule[-0.5em]{0pt}{1.5em}\\
        \bottomrule
    \end{tabular}
    \caption{Values considered for hyperparameters when tuning neural models.}
    \label{tab:training/babi/hyperparams}
\end{table}
\begin{table}[h]
    \centering
    \small{
    \begin{tabular}{l rr rr rr}
        \toprule
        \rule[-0.5em]{0pt}{1.5em} \bf{Hyperparameter}  & \multicolumn{2}{c}{\bf{Productivity}} & \multicolumn{2}{c}{\bf{Substitutivity}}  & \multicolumn{2}{c}{\bf{Systematicity}} \\
        \midrule
        Tuning Method & \textsf{Ax} & Inherit & \textsf{Ax} & Inherit & Inherit & \textsf{Ax} \\
        \midrule
        Learning rate ($\times 10^{-3}$) & 4.97, 5.00 & 5.00 & 3.60 & 5.00 & 5.00 & 50.0 \rule[-0.5em]{0pt}{1.5em}\\
        Batch size & 16, 4 & 4 & 1 & 4 & 4 & 1 \rule[-0.5em]{0pt}{1.5em}\\
        Max epochs & 50 & 50 & 100 & 50 & 50 & 100 \rule[-0.5em]{0pt}{1.5em}\\
        Wire dimension & 36 & 12 & 24 & 12 & 36 & 2 \rule[-0.5em]{0pt}{1.5em}\\
        Hidden layers & \textsf{Linear} & \textsf{Flat} & \textsf{Linear} & \textsf{Flat} & \textsf{Linear} &  \textsf{Hidden}(1) \rule[-0.5em]{0pt}{1.5em} \\
        Activation & - & \textsf{Mish} & - & \textsf{Mish} & - & \textsf{Mish} \rule[-0.5em]{0pt}{1.5em}\\
        \bottomrule
    \end{tabular}
    }
    \caption{
    Final hyperparameters selected for the neural models. Full tuning was not possible in all cases due to time and resource constraints; for the `inherited' configurations, hyperparameters from another task were taken, and adjusted with a few rounds of manual tuning if necessary.
    }
    \label{tab:training/babi/hyperparams-final}
\end{table}

\subsection{Overgeneralisation}
We test overgeneralisation in a similar way as the other tasks, however focus on only a single quantum and neural model architecture. For the neural model, we use the same hyperparameters as the productivity task, selecting the \textsf{Flat} architecture. Likewise, we pick the same quantum architecture as for proudctivity, but capped the max epochs at 30.

\subsection{Cross-validation}
We conduct a more in-depth analysis of the results on the productivity dataset. For each architecture choice investigated we run 5-fold cross validation to investigate the variability of the results.
A different number of max epochs than for tuned models was used to keep the evaluation time more feasible. We capped the quantum models at around 2.5 hours per run, leading to 40 epochs for the basic quantum model, and 60 for the curriculum models. For the neural architectures, we allowed 100 epochs, as each run took under 30 minutes.

\subsection{Extended validation}
We investigate the effect of different extended validation schemes for the cross-validated quantum productivity models (reported on in \autoref{sec:valid-schemes-results}).
To evaluate this, we consider the training runs from the cross-validation. For each run, we compute the validation accuracy at each epoch for a series of validation sets. These sets are selected such that they never overlap with any training data. We then select the models with the best respective validation accuracies, and compare their performance on the test set.

\subsection{Curriculum}
\label{sec:training-curriculum}
Finally, for the quantum architecture, we conduct an initial exploration of curriculum learning. This introduces some new hyperparameters, concerning how new elements are selected and introduced to the training data. We used the training strata as curriculum levels, such that the model is exposed to progressively more difficult datapoints during training. We controlled this in two ways: first, by varying the number of epochs over which the data is to be introduced, and second, whether the elements in the next stratum were introduced gradually, or all at once after a number of epochs. The hyperparameters were tuned manually, with the final values summarised in \autoref{tab:training/babi/hyperparams-curriculum}.

\begin{table}[h]
    \centering
    \begin{tabular}{l r r}
        \toprule
        \rule[-0.5em]{0pt}{1.5em} \bf{Hyperparameter} & \bf{Considered} & \bf{Value} \\
        \midrule
        Learning rate & 0.0002, 0.0004, 0,0005 & 0.0005 \rule[-0.5em]{0pt}{1.5em}\\
        Batch size & 1, 4, 8 & 1 \rule[-0.5em]{0pt}{1.5em}\\
        Curriculum epochs & 6, 10, 15, 25, 30, 40 & 30 \rule[-0.5em]{0pt}{1.5em}\\
        Max epochs & 30, 50, 75 & 75 \rule[-0.5em]{0pt}{1.5em}\\
        Curriculum step &  smooth, step & smooth \rule[-0.5em]{0pt}{1.5em} \\
        \bottomrule
    \end{tabular}
    \caption{Hyperparameters considered and final values used for curriculum learning.}
    \label{tab:training/babi/hyperparams-curriculum}
\end{table}
\clearpage
\section{Training for compositionality}
\label{sec:valid-schemes-results}
In this section, we consider how the validation split impacts model selection with regards to compositionality.

\subsection{Validation schemes}
As a baseline, we consider the \textit{vanilla} approach: the data is split into three segments called \textit{Train}, \textit{Validation} and \textit{Test}. The model is trained on the training instances, with the validation set being evaluated at each epoch. The final model selected is then the one with the highest validation accuracy. We enforce that the \textit{Validation} set is generated from the same distribution as the \textit{Train} set, whilst the \textit{Test} set will in general contain items from a broader distribution to test compositional generalisation. Note that in many other ML setups, the test set is \textit{also} drawn from an identical distribution to that of the \textit{Train} (and \textit{Validation}) set(s).

\begin{figure}[h]
    \centering
    \begin{subfigure}[b]{0.45\linewidth}
        \centering
        \includegraphics[scale=0.25]{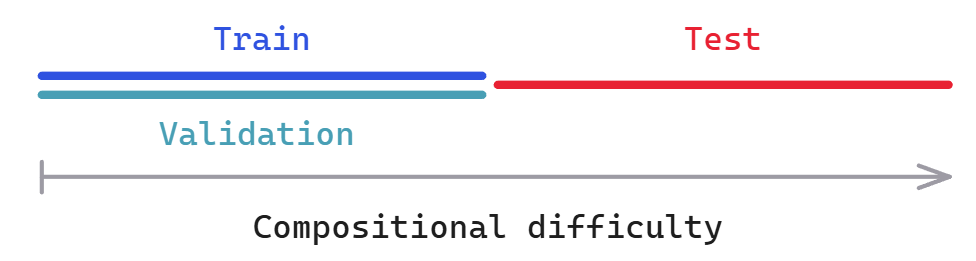}
        \caption{Vanilla}
        \label{fig:training/dataset-split-vanilla}
    \end{subfigure}
    \hspace{2em}
    \begin{subfigure}[b]{0.45\linewidth}
        \centering
        \includegraphics[scale=0.25]{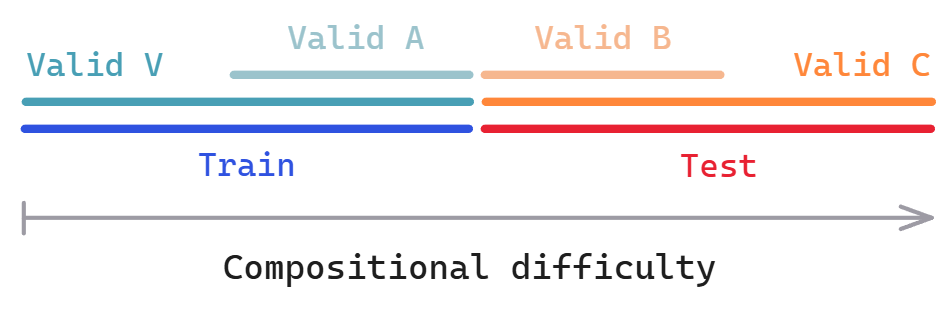}
        \caption{Extended Validation}
        \label{fig:training/dataset-split-extended}
    \end{subfigure}
    \caption{Visualising the dataset splits for the (a) vanilla and (b) extended validation training regimes.}
\end{figure}

Given the nature of compositionality, a natural extension of the vanilla approach is to introduce some datapoints from the \textit{Test} distribution to the validation set, so that the selected model is one that although still only trained on the \textit{Train} distribution, is one that shows more promise for generalisation.
We consider a selection of new validation set candidates, expressed according to the difficulty of the datapoints they contain. 
Note that \textit{Valid AB} and \textit{Valid All} can be used to estimate the compositionality score of the dataset. The splits are visualised in \autoref{fig:training/dataset-split-extended}, and summarised in \autoref{tab:validation-splits}.

\begin{table}[h]
    \begin{tabular}{p{0.1\linewidth} p{0.8\linewidth}}    
        \rule[0em]{0pt}{1.8em}%
        \textit{Valid V} & The vanilla validation set, where the distribution overlaps with the \textit{Train} set.
        \\
        \rule[0em]{0pt}{1.8em}%
        \textit{Valid A} & A subset of the \textit{Valid V} set, generally towards the `harder' end of the distribution.
        \\
        \rule[0em]{0pt}{1.8em}%
        \textit{Valid C} & An analogue to \textit{Valid V}, but where the distribution overlaps with the \textit{Test} distribution.
        \\
        \rule[0em]{0pt}{1.8em}%
        \textit{Valid B} & A (generally tractable) subset of \textit{Valid C}.
        \\
        \rule[0em]{0pt}{1.8em}%
        \textit{Valid AB} & A combination of \textit{Valid A} and \textit{Valid B}.
        \\
        \rule[0em]{0pt}{1.8em}%
        \textit{Valid All} & A combination of \textit{Valid V} and \textit{Valid C}.
    \end{tabular}
    \caption{Candidate extended validation datasets.}
    \label{tab:validation-splits}
\end{table}

\subsection{Results}
\begin{figure}[H]
    \centering
    \includegraphics[width=0.9\textwidth]{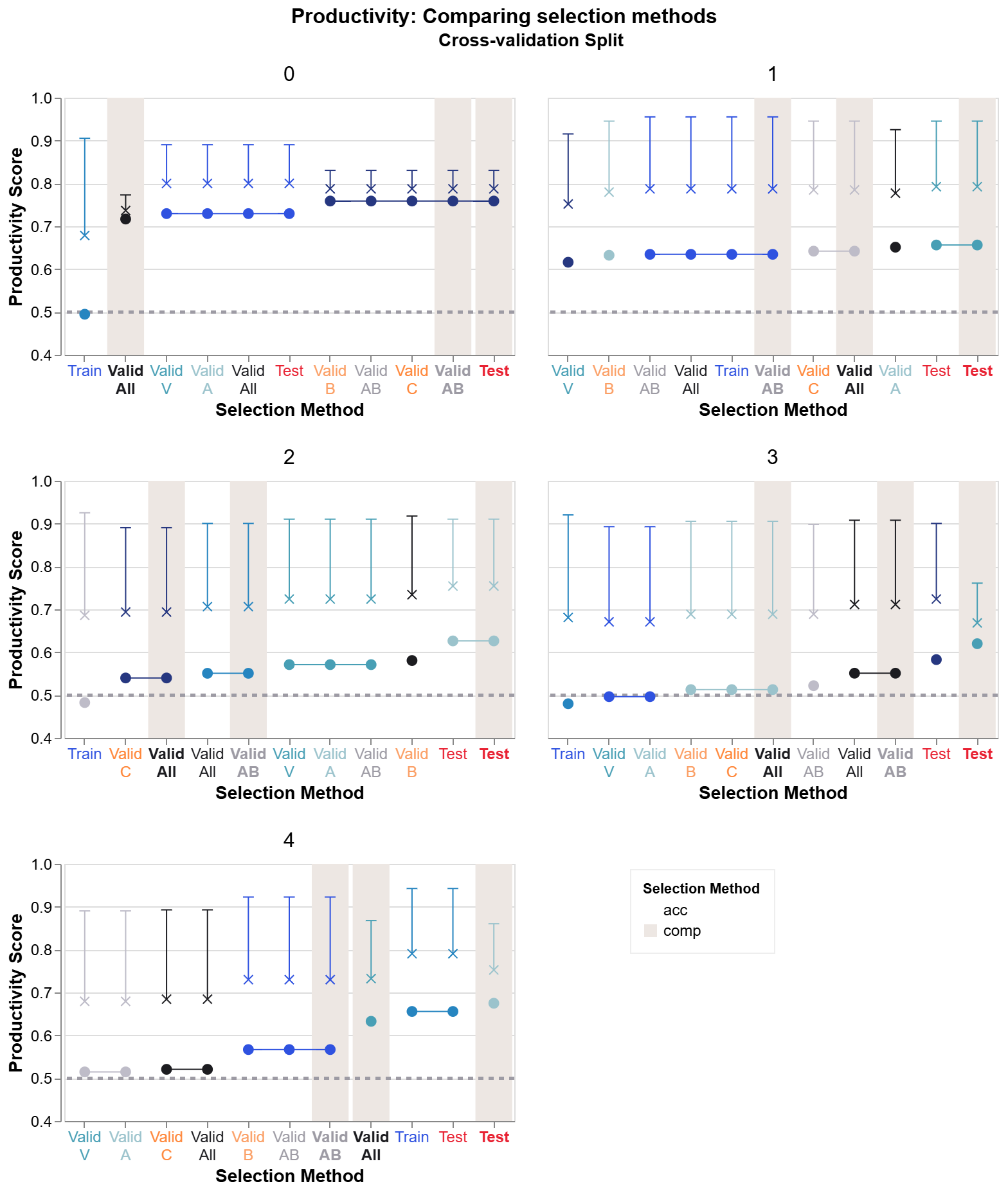}
    \caption{
    Productivity scores for the quantum models selected by each validation method, per cross-validation split. The productivity score is drawn as a point, with a bar displaying the difference in accuracy from \textit{Train} to \textit{Test}. The \textit{Test} accuracy is highlighted with a cross, while \textit{Train} is highlighted with a dash. In each case, the schemes are ranked by compositionality score. The plots are coloured according to the particular model selected per split. A line links the productivity scores obtained by the same model when selected by different schemes. The validation schemes that computed a proxy for the productivity score are highlighted with a shaded bar and bold text. For comparison, three extra schemes are added to the validation scheme selection. {\color{myBlue}Train}, depicted in dark blue, selects the model with the highest accuracy on the \textit{Train} set. We also include in red models that were selected by considering the accuracy on the \textit{Test} set: {\color{myRed}Test} selects the model with the highest \textit{Test} accuracy, while \textbf{\color{myRed}Test} (in bold) identifies the model with the highest compositional score. Note that the red \textit{Test}-based metrics are not normally available for model selection, but provide an insight as to what the best available model is for each split.
    }   
    \label{fig:results/productivity_validation-schemes}
\end{figure}

\autoref{fig:results/productivity_validation-schemes} displays the impact of the different validation choices on the selected model's final productivity score, as computed from the \textit{Test} set. 
We first notice that obtaining a high compositionality score is sometimes at the cost of \textit{Test} accuracy, as is the case with splits 0, 3 and 4.
Similarly, a high \textit{Train} score alone is insufficient to guarantee compositionality, as exhibited by splits 0, 2 and 3. This is by construction of the compositionality score - a high penalty is imposed on models that overfit the training distribution at the cost of $\varepsilon$-productivity.

This then seems to suggest one of the compositional schemes, or a scheme that encompasses both \textit{Train} and \textit{Test} distributions may be better.
Both \textit{Valid All} and \textit{Valid AB} compositional schemes are unfortunately relatively unreliable estimators of the final compositionality score (see \autoref{sec:results/comp-score-estimates} and associated commentary for a quantification).
This may be due to the low number of samples available for validation (20\% of the total dataset for each context depth sampled) leading to high volatility. Another source of error is from overfitting, where the model's high accuracy on the \textit{Train} set heavily penalises the final compositionality score, but not the estimates computed from the validation sets.

Despite this, the compositional score estimators tend to perform similarly to their accuracy based counterparts, but with a higher tendency to select models with lower overfitting between the \textit{Train} and \textit{Test} sets (as with splits 0 and 4). 
We hence suggest the compositionality score estimators (\textit{Valid All} and \textit{Valid AB}) as the best validation scheme candidates, with \textit{Valid AB} being more optimised for efficiency, being a subset of the \textit{Valid All} set and thus requiring fewer datapoint evaluations.
The results reported on in the rest of this work used the \textit{Comp AB} scheme.

\subsection{Compositionality Score Estimates}
\label{sec:results/comp-score-estimates}
\begin{figure}[h]
    \centering
    \includegraphics[width=0.9\textwidth]{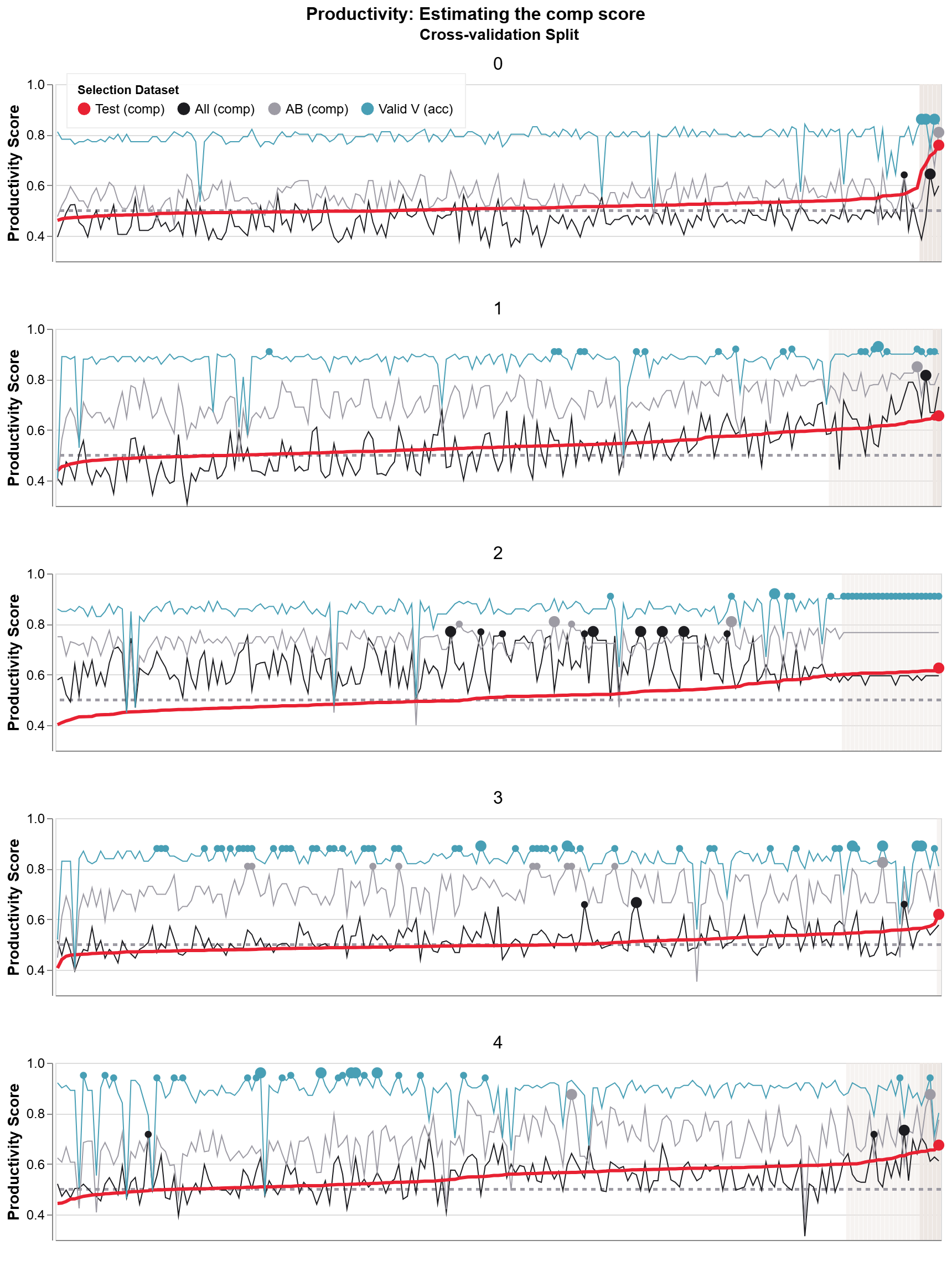}
    \\
    \textsf{Model}
    \caption{
    Comparing the validation and true compositionality scores for the models in each cross-validation split, for the quantum models. The true compositionality score is marked as a thick red line with the models arranged in increasing order. We plot the estimated compositionality score for each model as thin greyscale lines and the vanilla-style \textit{Valid V} accuracy as a blue line. The top 4\% of selected models (by score) are highlighted with a point, with the best models receiving a larger point. The areas where the true compositional score reaches 0.6 and 0.65 are shaded.
    }   
    \label{fig:results/productivity_validation-schemes-fit}
\end{figure}

In \autoref{fig:results/productivity_validation-schemes-fit}, we compare how closely each of the validation schemes track the actual productivity score of the model.
As alluded to above, the \textit{Comp All} and \textit{Comp AB} schemes are very volatile, and exhibit quite a lot of noise when compared to the true compositionality score.

In about half of the cases (splits 0, 1 and 4) the best model selected by the schemes correspond to good models, if not the best available.
These cases also correspond to instances where there was were fewer candidates with similar estimated scores. Indeed, the selection methods seem to work better when the actual score is above $0.60$, as there tends to be a model nearby with a high estimated score that rises above the noisy scores of the middling-to-bad models.

Finally, compared to the \textit{Valid V} accuracy, the compositionality score estimates select fewer candidate models. This is desirable as deciding between the candidates likely requires an expensive evaluation of a \textit{Valid Comp} style dataset: though the best model, or one very close to it, is reliably within the top 4\% of \textit{Valid V} scores, there are so many other models with similar scores that picking it out from the noise would be inefficient.
This suggests that only extending the validation set may not be enough to guarantee that a good model is selected. However, using an estimator for the compositionality score as an initial selection method, followed by a secondary validation set (\textit{Valid Comp}) to further refine the selection improves upon the naive compositional validation scheme used in \citet{duneau_scalable_2024}. 

\FloatBarrier

\section{Compositional generalisation}
\label{app:results}

\begin{figure}[H]
    \centering
    \begin{subfigure}[b]{0.8\linewidth}
        \centering
        \includegraphics[width=\textwidth]{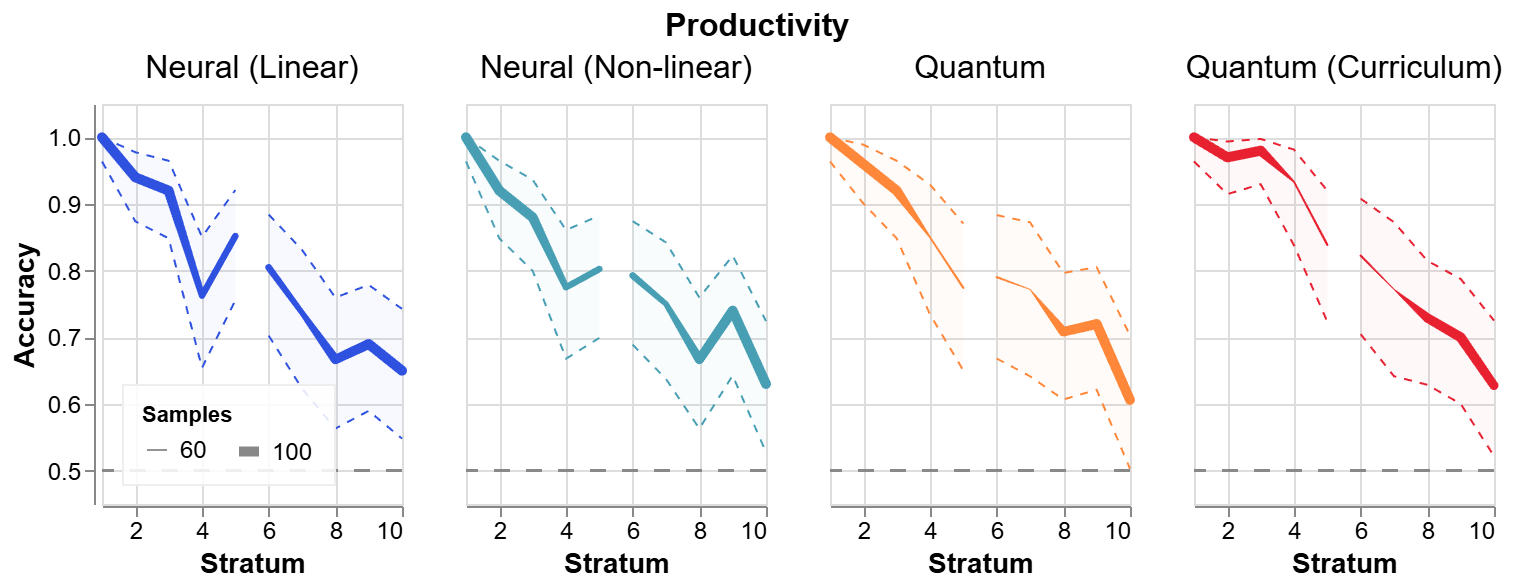}
        \caption{}
        \label{fig:results/babi/productivity-per-stratum}
    \end{subfigure}
    \\\vspace{0.5em}
    \begin{subfigure}[b]{0.8\linewidth}
        \centering
        \includegraphics[width=\textwidth]{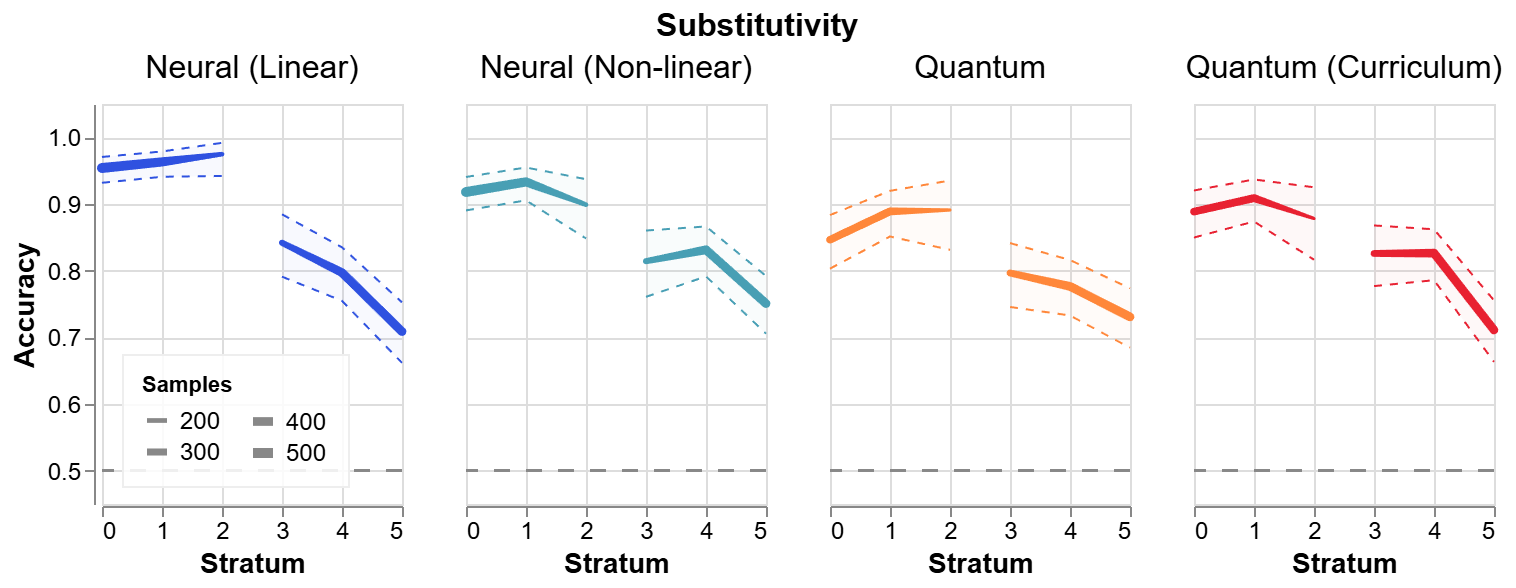}
        \caption{}
        \label{fig:results/babi/substitutivity-per-stratum}
    \end{subfigure}
    \\\vspace{0.5em}
    \begin{subfigure}[b]{0.8\linewidth}
        \centering
        \includegraphics[width=\textwidth]{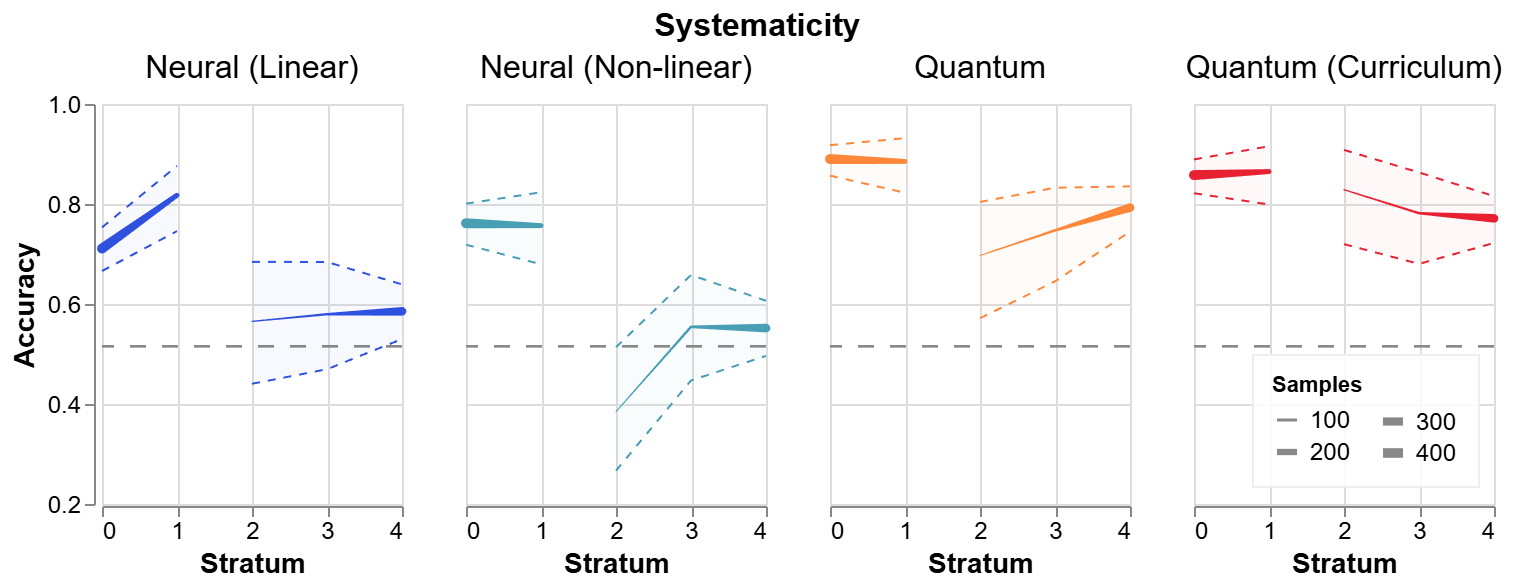}
        \caption{}
        \label{fig:results/babi/systematicity-per-stratum}
    \end{subfigure}
    \caption{
    Visualising the accuracy per stratum for the models trained on the (a) productivity, (b) substitutivity, and (c) systematicity datasets.
    The shaded regions represent the 95\% Clopper-Pearson interval for the mean. The thickness of the line reflects the number of datapoints available at each stratum. We display only the scores for the \textit{Train} and \textit{Test} datasets. These are disjoint, with the \textit{Train} samples on the left of the plot and \textit{Test} samples on the right, reflecting the divide in difficulty strata of the datapoints.
    }
    \label{fig:results/babi/compogen-per-stratum}
\end{figure}

\autoref{fig:results/babi/compogen-per-stratum} plots the model accuracies per training stratum, arranged according to the compositional difficulty for each of the productivity, substitutivity and systematicity datasets.

\FloatBarrier
\clearpage
\FloatBarrier

\section{Compositional interpretability}
\label{app:compinterp}
\begin{figure}[h]
    \centering
    \begin{subfigure}[b]{0.3\linewidth}
        \centering
        \includegraphics[scale=0.5]{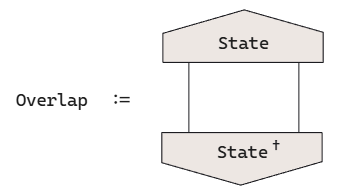}
        \caption{}
        \label{fig:interpret/babi/overlap}
    \end{subfigure}
    \hfill
    \begin{subfigure}[b]{0.65\linewidth}
        \centering
        \includegraphics[scale=0.5]{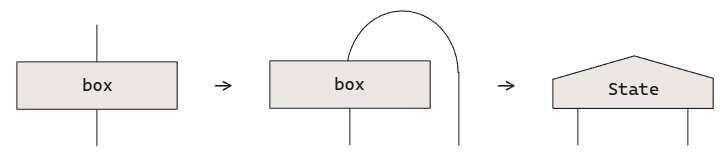}
        \caption{}
        \label{fig:interpret/babi/map-state}
    \end{subfigure}
    
    \caption{
    (a) Computing the overlap between states. Requires a notion of adjoints, or compact closure. The adjoint of a state is an effect, labelled with a dagger ($\dagger$).
    (b) Converting a box into a state. This requires a rigid category.
    }
\end{figure}

In this section we report in detail the methods used to interpret the productivity model.
We make use of the notion of overlap for quantum states, and the map-state duality (depicted in \autoref{fig:interpret/babi/overlap} and \autoref{fig:interpret/babi/map-state} respectively) in order to directly compare learnt boxes and diagram fragments with one another.

\begin{figure}[h]
    \centering
    \begin{subfigure}[b]{0.5\linewidth}
        \centering
        \includegraphics[scale=0.5]{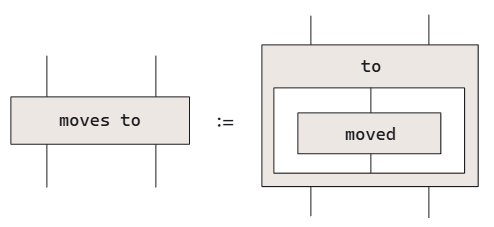}
        \caption{}
        \label{fig:interpret/babi/moves-simp}
    \end{subfigure}
    \hfill
    \begin{subfigure}[b]{0.4\linewidth}
        \centering
        \includegraphics[scale=0.5]{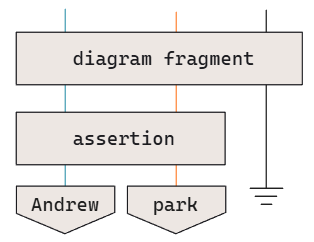}
        \caption{}
        \label{fig:interpret/babi/question-relative-diag}
    \end{subfigure}
    
    \caption{
    (a) Simplification of the depiction of \word{\wbox{moved} to}, where \wbox{moved} stands for any of the intransitive movement verbs.
    (b) An assertion-relative diagram fragment. For the bAbI~6 dataset the \wbox{assertion} box can be either \word{is in} or \word{is not in}.
    }
\end{figure}

We consider the two and three noun diagram fragments shown in \autoref{fig:interpret/babi/diag-fragments-base}, \autoref{fig:interpret/babi/diag-fragments-move-loc}, \autoref{fig:interpret/babi/diag-fragments-move-pers} and \autoref{fig:interpret/babi/diag-fragments-move-obj}. For a simpler depiction, we simplify the movement frame into a single box as per \autoref{fig:interpret/babi/moves-simp}. Similarly, we condense the object interaction verbs into a single box \wbox{grabs}.

\subsection{Box overlaps}
\begin{figure}[h]
    \centering
    \includegraphics[width=0.6\linewidth]{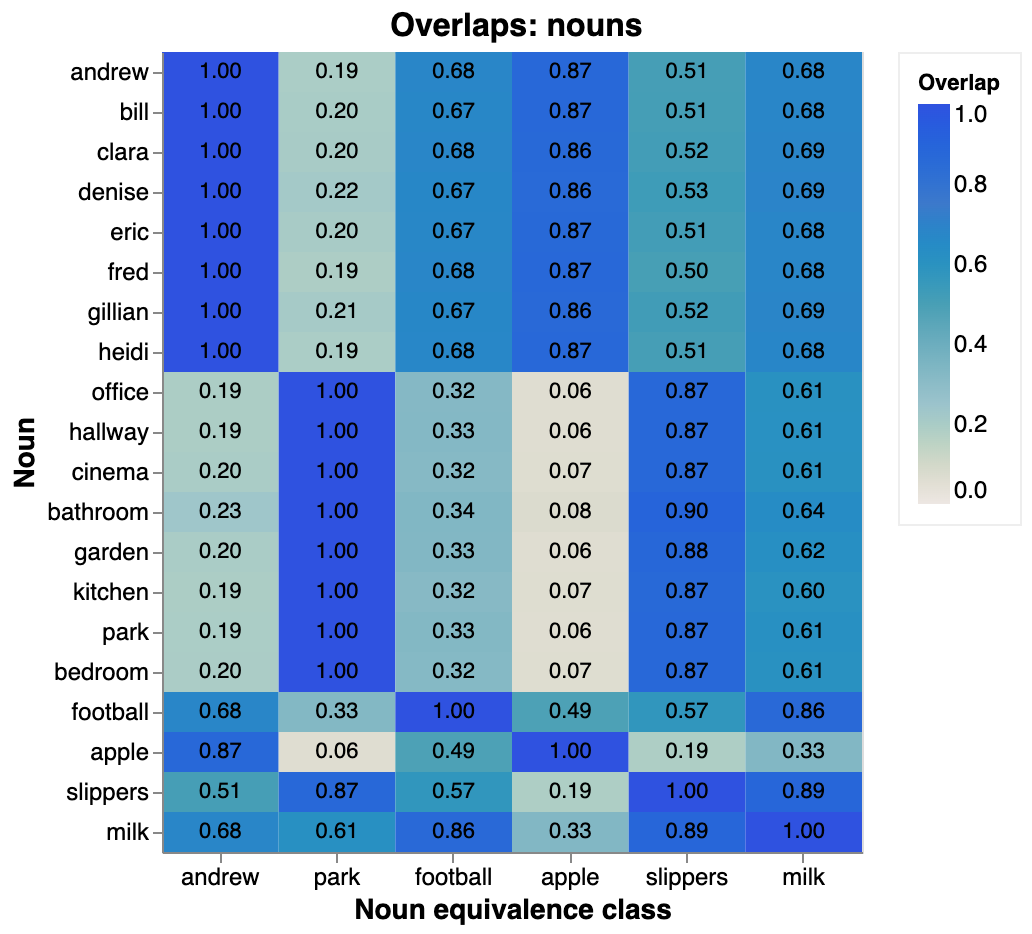}
    \caption{
        Visualising the relative overlaps between nouns. On the x-axis, we group the nouns by equivalence class, while the y-axis displays each of the nouns considered.
    }
    \label{fig:interpret/babi/prod-q-nouns}
\end{figure}

\begin{figure}[h]
    \centering
    \includegraphics[width=0.4\linewidth]{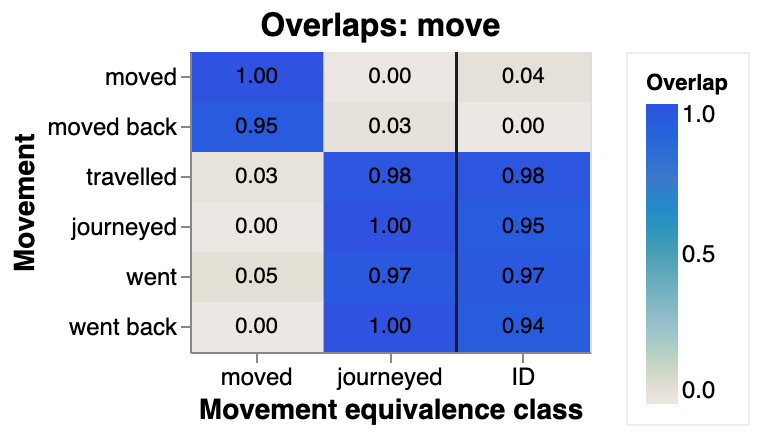}
    \caption{
        Visualising the relative overlaps between the intransitive \word{move}-type verbs. On the x-axis, we group the nouns by equivalence class, and include the identity for reference. The two classes are approximately orthogonal.
    }
    \label{fig:interpret/babi/prod-q-itv}
\end{figure}

\begin{figure}[h]
    \centering
    \includegraphics[width=0.7\linewidth]{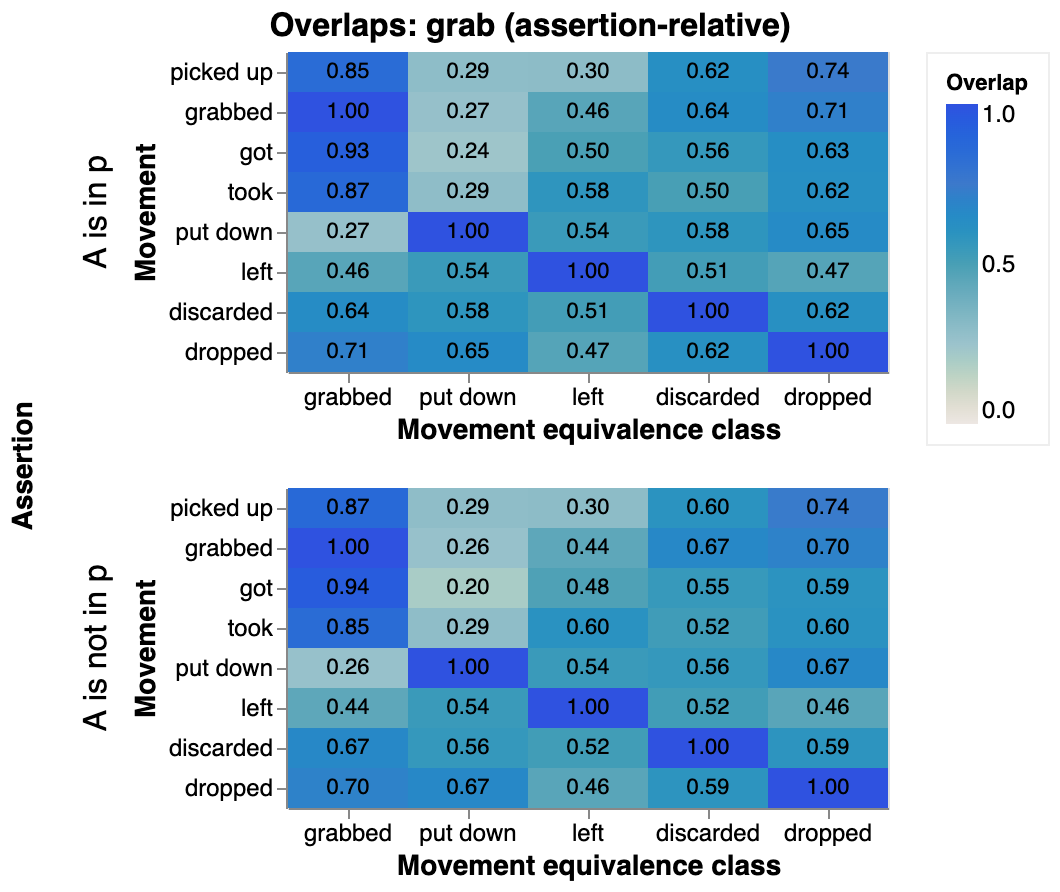}
    \caption{
        Visualising the assertion-relative overlaps between the transitive \word{grab}- and \word{drop}-type verbs. On the x-axis, we group the nouns by equivalence class.
    }
    \label{fig:interpret/babi/prod-q-grab}
\end{figure}

We visualise the distances between the nouns in \autoref{fig:interpret/babi/prod-q-nouns}, the movement verbs in \autoref{fig:interpret/babi/prod-q-itv}, and the objects interactions in \autoref{fig:interpret/babi/prod-q-grab}.

\paragraph{Nouns}
We find the the model has learnt to group the people and places together, though it remains largely agnostic when it comes to the objects. This is somewhat to be expected, as the task structure provides very little information to constrain the embedding of the objects, since they are not relevant to the task solution.

\paragraph{Movement verbs}
We first consider the inner movement verbs that apply to only a single actor, then the composite \word{\textsf{moves} to}. We find that the intransitive movement verbs can be classed into two equivalent groups - this comes some way to realising the synonyms present in the actual task (in which all movement verbs are equivalent).
Inspecting the symmetries, we also note that one group - the larger one - acts as approximately the identity.

\paragraph{Object interaction verbs}
Like the object noun states, these verbs are also less constrained by the dataset, as they do not contribute to the final answer but rather act as confounding information. Plotting their overlaps with respect to each other directly, we find that they are almost all small. A more instructive comparison is then to take their overlaps relative to the assertion states. We find that relative to these states, all \word{grab} type words are very similar, while there is more variation for the \word{drop}-type words.
This suggests that the model has come some way to identifying these verbs as synonyms, despite the relatively low contribution these verbs have to the overall task.

\paragraph{Questions}
As a binary task, there are again two possible question assertions. Since we find that there is approximately one person state, and one location state, we can express the assertions directly as states.
We find these to be approximately orthogonal, with an overlap of $0.00053$.

\FloatBarrier

\subsection{Comparing diagram fragments}
\label{app:interp/fragments}
Here we provide an analysis for each fragment considered. For ease of reference, the fragments are abbreviated with a pair of letters corresponding to each verb. We use capitals to refer to people (\word{\emph{A}ndrew} or \word{\emph{C}lara}), and lowercase letters to refer to locations (\word{\emph{p}ark} and \word{\emph{k}itchen}) and \emph{o}bjects. We also include a special \emph{id}entity (ID) fragment, where no operations are applied to the initial noun states. When multiple sentences follow each other, we indicate this with a dash, for example the abbreviation (Ap-Ck-Ao) would stand for the fragment capturing \sent{\emph{A}ndrew moves to the \emph{p}ark. \emph{C}lara moves to the \emph{k}itchen. \emph{A}ndrew picks up the milk.}.

\begin{figure}[h]
    \centering
    \begin{subfigure}[b]{0.33\linewidth}
        \centering
        \includegraphics[width=\linewidth]{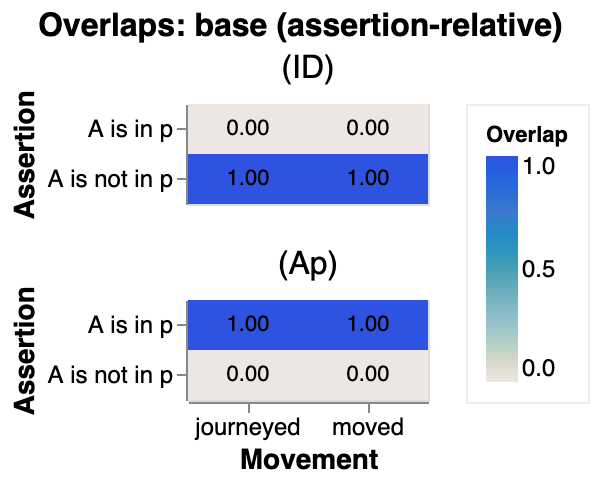}
        \caption{}
        \label{fig:interpret/babi/prod-q-axioms-base}
    \end{subfigure}
    \hfill
    \begin{subfigure}[b]{0.6\linewidth}
        \centering    
        \includegraphics[width=\linewidth]{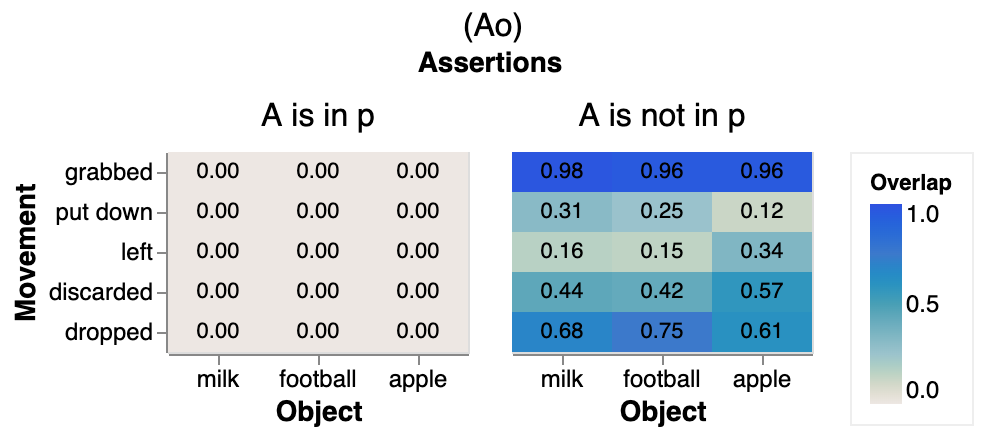}
        \caption{}
        \label{fig:interpret/babi/prod-q-axioms-Ao}
    \end{subfigure}
    \caption{
    (a) Visualising the magnitude of the overlaps between the diagram fragments (ID), (Ap) (visualised in \autoref{fig:interpret/babi/diag-fragments-base}), relative to each question state.
    (b) Visualising the magnitude of the overlaps between the diagram fragment (Ao) and the assertion states, for each choice of action and object. 
    The values have been renormalised such that the maximum overlap remains 1.
    }
\end{figure}

\paragraph{Base cases}
First, we consider the base cases, simulating the simplest texts. The diagram fragments and associated results are given in \autoref{fig:interpret/babi/diag-fragments-base} and \autoref{fig:interpret/babi/prod-q-axioms-Ao}.

(Ap) \emph{Base case: \sentq{Andrew is in the park.}}
The model is able to correctly identify the positive base case. Indeed, it seems that the question states are directly aligned with this state, as they achieve the maximum/minumum possible scores.

(ID) \emph{Base case: \sentq{Andrew is not in the park.} I}
As with the positive case, the model correctly identifies this instance, with the states again lining up approximately perfectly.
Interestingly, such a situation never occurs in the dataset, as it would require a character to be introduced without doing anything, so this is a new rule that the model has successfully extrapolated.

(Ao) \emph{Base case: \sentq{Andrew is not in the park.} II}
One situation that does occur in the dataset is for a person to be introduced via an object interaction and nothing else. We hence check how the model handles this situation as a secondary base case for the negative assertion, and find that the model captures this well for the \word{grab}-type verbs, but with a lower margin for the \word{drop}-type verbs. 
Though the object embeddings were each distinct, we also find that this difference is no longer as evident when factoring through the question states, suggesting the model has learnt to consider them equivalent.

\FloatBarrier

\begin{figure}[H]
    \centering
    \begin{subfigure}[b]{0.32\linewidth}
        \raggedleft
        \includegraphics[scale=0.55]{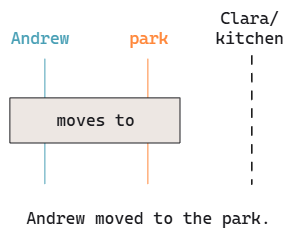}
        \caption{Ap}
        \label{fig:interpret/babi/diag-fragments-Ap}
    \end{subfigure}
    \hfill
    \begin{subfigure}[b]{0.32\linewidth}
        \centering
        \includegraphics[scale=0.55]{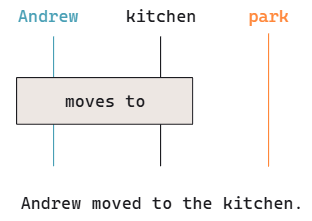}
        \caption{Ak}
        \label{fig:interpret/babi/diag-fragments-Ak}
    \end{subfigure}
    \hfill
    \begin{subfigure}[b]{0.32\linewidth}
        \raggedright
        \includegraphics[scale=0.55]{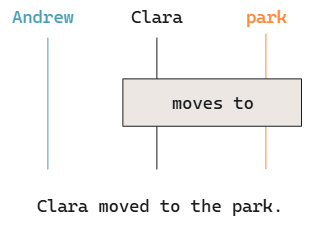}
        \caption{Cp}
        \label{fig:interpret/babi/diag-fragments-Cp}
    \end{subfigure}
    \\\vspace{1em}
    \begin{subfigure}[b]{0.7\linewidth}
        \centering
        \includegraphics[width=\linewidth]{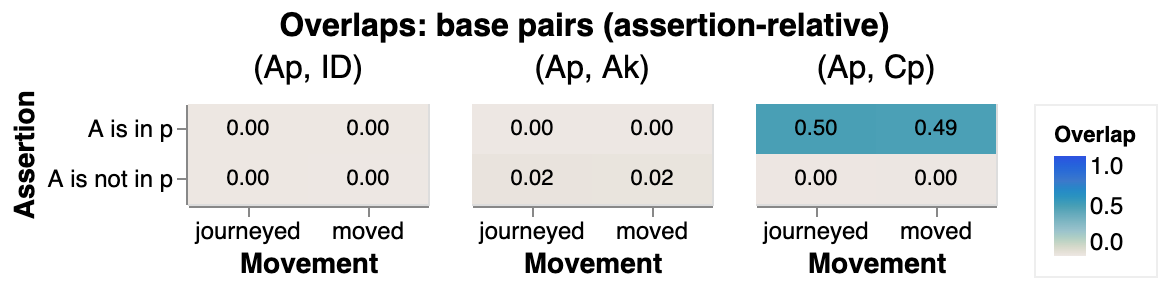}
        \caption{}
        \label{fig:interpret/babi/prod-q-axioms-move-base}
    \end{subfigure}
    \caption{
    The selection of single-action diagram fragments considered and their overlaps.
    (a)-(c) The diagram fragments and their shorthand identifiers. The \word{Andrew} and \word{park} wires are coloured for emphasis.
    (d) Visualising the magnitude of the overlaps between the assertion-relative diagram fragments. For mixed states, the overlaps have been renormalised so that the maximum remains 1.
    }
    \label{fig:interpret/babi/diag-fragments-move-singles}
\end{figure}

\paragraph{Singletons}
Next, we compare fragments with at most one verb, shown in \autoref{fig:interpret/babi/diag-fragments-move-singles}.

(Ap $\neq$ ID) \emph{Moving matters.}
Here, we check whether the act of moving has consequences, by comparing the fragment \sent{\emph{A}ndrew moves to the \emph{p}ark} with the initial states for \word{Andrew} and \word{park}. We see that the overlap between the states is very small for both assertion states, which suggests that we cannot simply delete a direct movement without affecting the final answer.
 
(Ap $\neq$ Ak) \emph{Distinguishable locations.}
This axiom captures the idea that it matters which place Andrew visits.
In this case, all overlaps are close to zero, suggesting that the model is indeed sensitive to which location the target person visits.

(Ap $\neq$ Cp) \emph{Distinguishable people.}
For this pair, we check whether which person visits the target location has an impact. 
The results in this case have a similar structure for the different movement verbs, but the overlaps differ depending on the assertion state. 
In both cases, we find the overlap is zero relative to the negative assertion, and large relative to the positive one, suggesting that changing \textit{who} visited a location will have a much larger impact on the overlap between the story with the negative assertion than the positive one.

Though we might intuitively like both overlaps to be close to zero, this does not necessarily mean that the model is \textit{insensitive} to which person visits a location, since we only care about the \textit{difference} between the assertion overlaps. A sufficient change in only one of the overlaps may be enough to change the model's final answer, provided the fixed overlap was not too extreme to start with.

\FloatBarrier

\begin{figure}[h]
    \centering
    \begin{subfigure}[b]{0.29\linewidth}
        \centering
        \includegraphics[scale=0.5]{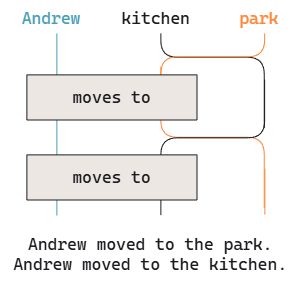}
        \caption{Ap-Ak}
        \label{fig:interpret/babi/diag-fragments-Ap-Ak}
    \end{subfigure}
    \hfill
    \begin{subfigure}[b]{0.29\linewidth}
        \centering
        \includegraphics[scale=0.5]{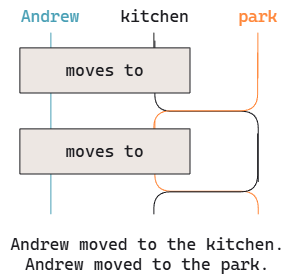}
        \caption{Ak-Ap}
        \label{fig:interpret/babi/diag-fragments-Ak-Ap}
    \end{subfigure}
    \hfill
    \begin{subfigure}[b]{0.35\linewidth}
        \centering
        \includegraphics[width=\linewidth]{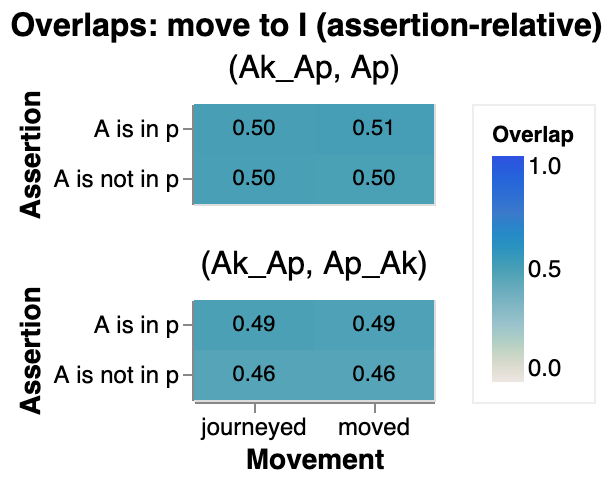}
        \caption{}
        \label{fig:interpret/babi/prod-q-axioms-move-loc}
    \end{subfigure}
    
    \caption{
    The selection of three-noun diagram fragments considered for movement verbs involving a second location, and the paired assertion-relative overlaps.
    (a), (b) The diagram fragments and their shorthand identifiers. The wires corresponding to \word{Andrew} and \word{park} are coloured to help distinguish them.
    (c) Visualising the magnitude of the overlaps between the states relative to each assertion state. For mixed states, the overlaps have been renormalised so that the maximum remains 1.
    }
    \label{fig:interpret/babi/diag-fragments-move-loc}
\end{figure}

\paragraph{Extra location}
Thirdly, we consider cases where the target person moves to a different location, as per \autoref{fig:interpret/babi/diag-fragments-move-loc}.

(Ap-Ak $\neq$ Ak-Ap) \emph{Visit ordering.}
We expect this pair to have low overlap for a satisfied axiom, reflecting that the order in which places are visited by a person matters. Instead, we find a fairly high overlap between the states, suggesting that the model is perhaps not distinguishing this as as well as it should.

(Ak-Ap $=$ Ap) \emph{Forgetfulness (other locations).}
For the purposes of this task, we do not need to remember where a person has been, only where they are now. This axiom captures the idea that we can forget where they were previously to entering the target location. The overlaps are all roughly 0.5 here, suggesting that the model does not manage to forget locations very well. This result is somewhat unsurprising given the unitary nature of the model.

\FloatBarrier

\begin{figure}[h]
    \centering
    \begin{subfigure}[b]{0.29\linewidth}
        \centering
        \includegraphics[scale=0.5]{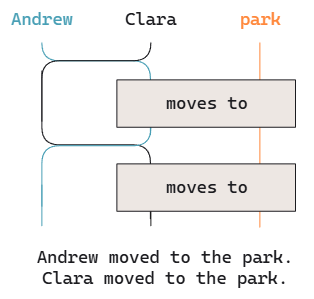}
        \caption{Ap-Cp}
        \label{fig:interpret/babi/diag-fragments-Ap-Cp}
    \end{subfigure}
    \hfill
    \begin{subfigure}[b]{0.29\linewidth}
        \centering
        \includegraphics[scale=0.5]{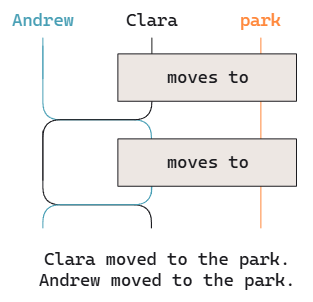}
        \caption{Cp-Ap}
        \label{fig:interpret/babi/diag-fragments-Cp-Ap}
    \end{subfigure}
    \hfill
    \begin{subfigure}[b]{0.35\linewidth}
        \centering
        \includegraphics[width=\linewidth]{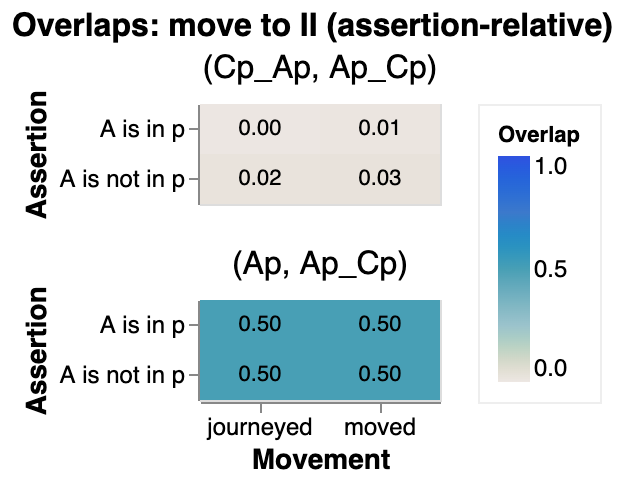}
        \caption{}
        \label{fig:interpret/babi/prod-q-axioms-move-pers}
    \end{subfigure}
    
    \caption{
    Diagram fragments and overlaps for movement combinations involving a second person.
    (a), (b) The selection of diagram fragments considered, and their shorthand identifiers. The wires corresponding to \word{Andrew} and \word{park} are coloured for emphasis.
    (c) Visualising the assertion-relative overlaps between the states. The overlaps have been renormalised so that the maximum remains 1.
    }
    \label{fig:interpret/babi/diag-fragments-move-pers}
\end{figure}

\paragraph{Extra person}
Next, we consider cases where another person enters the target location, as depicted in \autoref{fig:interpret/babi/diag-fragments-move-pers}.

(Cp-Ap $=$ Ap-Cp) \emph{Arrival order.}
Here we expect a high overlap, as the order in which people arrive at the target location should have no influence on the target person's presence in that location. The model displays low overlaps with both assertions, however, suggesting that other people visiting the target location are likely to confuse it.

(Ap-Cp $=$ Ap) \emph{Forgetfulness (other people).} 
Next, we consider whether the model finds the presence of other people in a location to be important. As with the previous forgetfulness axiom, the values are around 0.5, suggesting that the model struggles to ignore the presence of other people in the target location completely.
Combined with the behaviour of the (Cp-Ap, Ap-Cp) pair, we can conclude that the model may be quite sensitive to the presence or absence of other people in the target location.

\FloatBarrier

\begin{figure}[h]
    \centering
    \begin{subfigure}[b]{0.32\linewidth}
        \centering
        \includegraphics[scale=0.56]{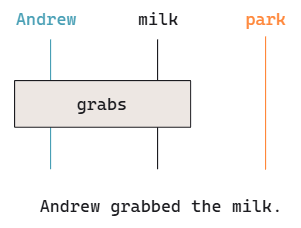}
        \caption{Ao}
        \label{fig:interpret/babi/diag-fragments-Ao}
    \end{subfigure}
    \hfill
    \begin{subfigure}[b]{0.32\linewidth}
        \centering
        \includegraphics[scale=0.56]{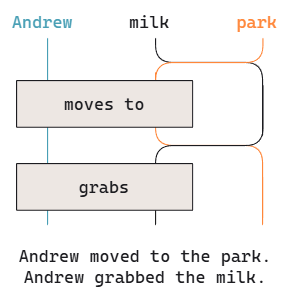}
        \caption{Ap-Ao}
        \label{fig:interpret/babi/diag-fragments-Ap-Ao}
    \end{subfigure}
    \hfill
    \begin{subfigure}[b]{0.32\linewidth}
        \centering
        \includegraphics[scale=0.6]{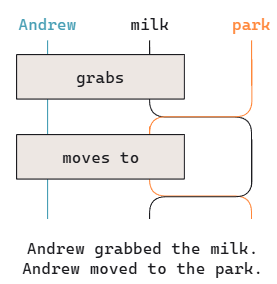}
        \caption{Ao-Ap}
        \label{fig:interpret/babi/diag-fragments-Ao-Ap}
    \end{subfigure}
    
    \caption{
    The selection of three noun diagram fragments considered for movement verbs and object interactions, with their shorthand identifiers. The wires corresponding to \word{Andrew} and \word{park} are coloured to help distinguish them.
    }
    \label{fig:interpret/babi/diag-fragments-move-obj}
\end{figure}

\begin{figure}[h]
    \centering
    \includegraphics[width=0.6\linewidth]{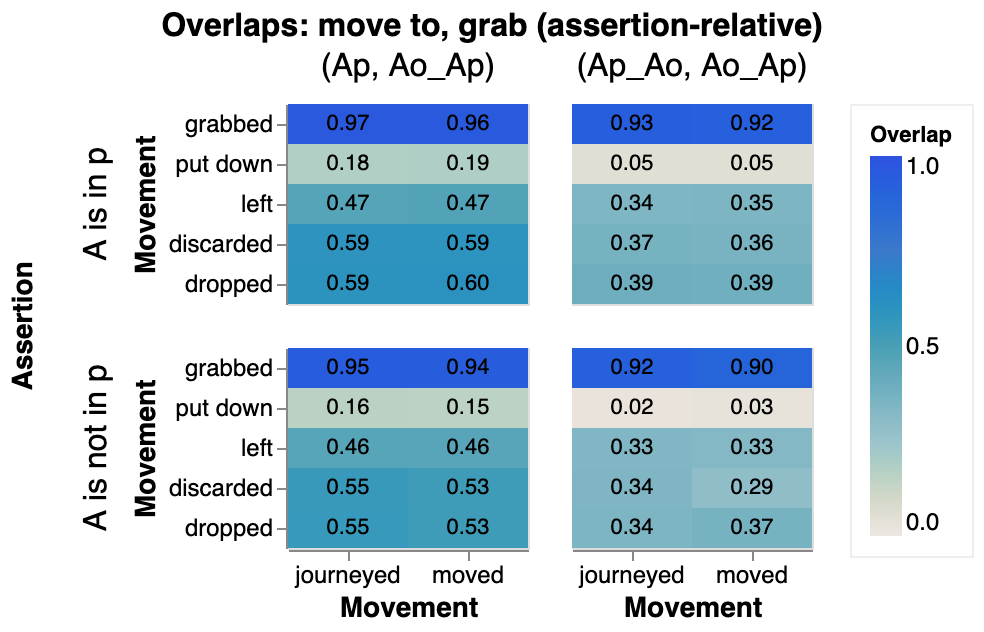}
    \caption{
        Visualising the magnitude of the assertion-relative overlaps between diagram fragments involving object interactions, from \autoref{fig:interpret/babi/diag-fragments-move-obj}. The values have been renormalised such that the maximum overlap is 1.
    }
    \label{fig:interpret/babi/prod-q-axioms-grab}
\end{figure}

\paragraph{Extra object}
Finally, we consider adding an object interaction. We display the diagram fragments in \autoref{fig:interpret/babi/diag-fragments-move-obj}, and the results in  \autoref{fig:interpret/babi/prod-q-axioms-grab}.

(Ao-Ap $=$ Ap) \emph{Forgetfulness (object interactions).} 
The third kind of interaction to ignore are those with objects. \autoref{fig:interpret/babi/prod-q-axioms-grab} visualises the overlaps for the object interactions. We find that the model behaves similarly across the different \word{move}-type verbs, but quite differently over the the \word{grab}- and \word{drop}-type verbs. For  \word{grab}-type verbs, the model appears to successfully ignore the object interaction, while the \word{drop}-type verbs have lower overlaps, especially \word{put down}. This difference may be due to the sparsity of such instances in the data.

(Ap-Ao $=$ Ao-Ap) \emph{Teleporting objects.} 
For the purposes of this task, it does not matter where a person is when they interact with an object (indeed the object interactions do not matter at all) - the objects can hence teleport between locations with no impact on the final answer\footnote{We note here that it is possible to construct a more difficult version of the task in which a person's location can be inferred via their object interactions, given that people can only interact with objects in their present location. Thus from \sent{Clara moved to the park. Clara put down the milk. Andrew picked up the milk.} we could infer that Andrew is in the park. We did not make use of this rule when generating the datasets used here.}.
The model appears to have largely captured this invariance for the \word{grab}-type words, but again struggles for the \word{drop}-type verbs, especially \word{put down}.

\stopcontents[appendices]

\end{document}